%% file: ragu_preprint.tex
\documentclass[10pt]{article} % For LaTeX2e
\usepackage[accepted]{tmlr}
\input{math_commands.tex}

\usepackage{hyperref}
\usepackage{url}

\usepackage[T1]{fontenc}
\usepackage[utf8]{inputenc}

\usepackage{microtype}

\usepackage{inconsolata}

\usepackage{graphicx}

\usepackage{caption}

\usepackage{mathtools} 
\usepackage{amsmath} 
\usepackage{hhline}
\usepackage{enumitem}
\usepackage{makecell}
\usepackage{multirow,booktabs}
\usepackage{xcolor}
\usepackage[most]{tcolorbox}

\definecolor{myblue}{HTML}{5B7C99}  % Define a custom blue color
\usepackage{tikz,subfig}
\usepackage{pgfplots}
\usepackage{filecontents}
\usetikzlibrary{patterns,calc}
\usepackage{amsmath, amssymb}
\usepackage{textcomp}
\definecolor{bluishGreen}{RGB}{0, 158, 115}
\definecolor{vermilion}{RGB}{213, 94, 0}
\definecolor{reddishPurple}{RGB}{204, 121, 167}
\definecolor{skybleu}{RGB}{86, 180, 233}
\pgfplotsset{compat=1.9}

\usepackage{pifont}
\newcommand{\cmark}{\ding{51}}%

\newtcolorbox{promptbox}[2][]{colback=white,
colframe=black!10!white, colbacktitle=white,enhanced,coltitle=black,
attach boxed title to top center={yshift=-2mm},
title={#2},#1}

\makeatletter
\newcommand{\thickhline}{%
    \noalign {\ifnum 0=`}\fi \hrule height 1pt
    \futurelet \reserved@a \@xhline
}
\makeatother

\title{Uncertainty Quantification in Retrieval Augmented Question Answering}

\author{\name Laura Perez-Beltrachini \email lperez@ed.ac.uk \\
      \addr University of Edinburgh
      \AND
      \name Mirella Lapata \email mlap@ed.ac.uk \\
      \addr University of Edinburgh}

\def\month{09}  % Insert correct month for camera-ready version
\def\year{2025} % Insert correct year for camera-ready version
\def\openreview{\url{https://openreview.net/forum?id=JLkgI0h7wy}} % Insert correct link to OpenReview for camera-ready version

\begin{document}

\maketitle

\begin{abstract}
Retrieval augmented Question Answering (QA) helps QA models overcome
knowledge gaps by incorporating retrieved evidence, typically a set of
passages, alongside the question at test time.  Previous studies show
that this approach improves QA performance and reduces hallucinations,
without, however, assessing whether the retrieved passages are indeed
useful at answering correctly. In this work, we propose to quantify
the uncertainty of a QA model via estimating the utility of the
passages it is provided with. We train a lightweight neural model to
predict passage utility for a target QA model and show that while
simple information theoretic metrics can predict answer correctness up
to a certain extent, our approach efficiently approximates or
outperforms more expensive sampling-based methods.\footnote{Code and data are available at 
%$\mathbb{XXXX}$}.
\href{https://github.com/lauhaide/ragu}{https://github.com/lauhaide/ragu}}
\end{abstract}

\section{Introduction}

Retrieval augmented Question Answering (QA) allows QA models to
overcome knowledge gaps at test time through access to evidence in the
form of retrieved passages \citep{rag-lewis,guu20a-realm,atlas}.
Recent work leverages external retrievers
\citep{chen-etal-2017-drQA,izacard-grave-2021-leveraging} and the
language understanding and generation capabilities of Large Language
Models (LLMs; \citealt{fewshot-learners,instructGPT}) to predict
answers based on questions \emph{and} retrieved passages which are
provided as input context. In Figure~\ref{fig:ragu:data:ex}, we show
an example of a question (\textsl{Who sings Does He Love Me with
  Reba?}),  retrieved passages, and predicted answers.

Retrieval augmented QA architectures have proven beneficial in
increasing LLM performance on tail knowledge
\citep{atlas,mallen-etal-2023-popQA}, reducing hallucinations in the
generated answers \citep{shuster-etal-2021-retrieval-augmentation}, 
and even improving model calibration
\citep{10.1162/tacl_a_00407}.  However, there are various ways in
which retrieval augmented QA can go wrong.
%at inference time. 
The set of retrieved passages is far from perfect
\citep{sciavolino-etal-2021-simple,yoran2024making,realtimeQA:2024}
containing irrelevant, incomplete, or misleading evidence, the model might be under-trained to read certain passages and reason over these and the question \citep{atlas,liu-etal-2024-lost,sun2025redeep}, and the question can be ambiguous or unanswerable \citep{realtimeQA:2024}.  
Ultimately, QA models may not follow the provided passages 
\citep{xie2024adaptive,joren2025sufficientcontextnewlens}.
When faced with uncertainty, QA models should ideally acknowledge it
(e.g., by communicating it)
rather than risk an incorrect response.

A good deal of previous work has focused on quantifying \emph{answer
uncertainty} in the context of \emph{closed-book} QA tasks, where the answer
is predicted based on a question and the model's encoded
knowledge. Sampling-based methods rely on output discrepancies among
multiple predictors of the same input
\citep{mcdropout-gal16,NIPS2017_deep:ensembles}.
They measure diversity on a set of answers
\citep{kuhn2023semantic,chen-mueller-2024-quantifying} sampled via
temperature scaling \citep{pmlr-v70-guo17a}, with larger variance
indicating higher uncertainty. LLM-based methods rely on the QA
model's own judgment of uncertainty
\citep{kadavath2022-ptrue,lin2022teaching,tian-etal-2023-just}. Through prompting, the model is encouraged to express its uncertainty
(e.g.,~0.5 or `\textsl{almost certain}'), either alongside the
predicted answer \citep{lin2022teaching,tian-etal-2023-just} or after
generating it \citep{kadavath2022-ptrue,tian-etal-2023-just}.
None of these approaches has been applied in the context of 
\emph{retrieval augmented} QA \citep{10.1145/3744238}.

\begin{figure}[t]
    \centering
  \tcbset{colback=myblue!20!white,colframe=myblue!75!white}
\begin{tcolorbox}[arc=4mm,outer arc=1mm]
\begin{center}\footnotesize{\texttt{Who sings does he love me with Reba?}}\end{center}
\end{tcolorbox}
%\vspace{-.2cm}
\begin{tcolorbox}[colback=yellow!10!white,colframe=green!45!black,enhanced,title=\texttt{\scriptsize Linda Davis},
%{yshift=-3mm,yshifttext=-1mm},
%attach boxed title to bottom text right={yshift=6mm},
 boxed title style={colback=green!40!black}]
\scriptsize{\texttt{Does He Love You. Does He Love You "Does He Love You" is a song
written by Sandy Knox and Billy Stritch, and recorded as a duet by
American country music artists Reba McEntire and Linda Davis. It was
released in August 1993 as the first single from Reba's album "Greatest
Hits Volume Two". It is one of country music's several songs [cont.]
\hfill \textcolor{blue}{\texttt 2.77}}}
\vspace*{-.1cm}
\end{tcolorbox}
%\vspace{-.2cm}
\begin{tcolorbox}[colback=yellow!10!white,colframe=red,enhanced,title=\texttt{\scriptsize{Patti LaBelle}},
boxed title style={colback=red}]
\scriptsize{\texttt{Does He Love You. on Patti LaBelle's album, \"Flame\". The song features a vocal battle between two female narrators who are in love with the same man. Both women know that the man is being unfaithful to them and are wondering who he truly loves. The big-budget, Jon Small-directed video was filmed over 3 days in mid-1993. It begins with Reba in her dressing room wearing a lilac feather gown, where she sees a picture of her lover, which she glances at. [cont.]\hfill \textcolor{blue}{\texttt -0.10}}}
\vspace*{-.1cm}
\end{tcolorbox}
    \vspace{-0.5em}
    \caption{Example question from Natural Questions dataset
      \citep{kwiatkowski-etal-2019-natural} with two top-retrieved
      passages using Contriever-MSMARCO
      \citep{izacard2022unsupervised}.  On top of each passage, we
      show the answer generated by \textsc{Gemma2-9B} when prompted
      with that passage and the question.  The QA model answers
      correctly (green) only when prompted with the first passage.
      Numbers at the bottom right of each passage are
      \textcolor{blue}{utility scores} predicted by our model (higher
      values indicate more useful passages).}
    \label{fig:ragu:data:ex}
\end{figure}

In this paper, we focus on answer uncertainty estimation in the context of retrieval augmented QA. We hypothesize that
a passage is \textit{useful}, if a model can correctly answer
questions based on it.  If passages are informative and prime the QA
model towards appropriate knowledge \cite{geva-etal-2021-transformer}, 
we expect it to produce a correct
answer.  On the contrary, if passages are irrelevant or misleading and 
the question falls outside the QA model's knowledge, it is likely to 
produce an incorrect answer, either factually inaccurate or entirely
fabricated. Importantly, this notion of \emph{utility} is based on how
the target QA model will answer with the provided passages and not on 
what an external judge (e.g., entailment model) thinks about them.
We quantify the {utility} of a retrieved passage with a small neural model trained on utility judgments obtained by observing the target QA model's answering behavior. We borrow ideas from direct uncertainty
quantification approaches
\citep{pmlr-v119-van-amersfoort20a,lahlou2023deup} but do not
decompose uncertainty or outline shifts in the input distribution.  We
make utility predictions for each retrieved passage which we then use
to estimate the uncertainty of the QA model when prompted with a set thereof.

We evaluate our approach on short-form information-seeking QA tasks
\citep{rodriguez-boyd-graber-2021-evaluation} (see Figure~\ref{fig:ragu:data:ex} for an example).  Results on six datasets 
show that our uncertainty estimator is comparable or outperforms existing
sampling-based methods while being more test-time efficient. Sampling-based
solutions are expensive for in-production QA systems, in terms of latency 
and cost (e.g.,~QA engines built on top of proprietary language models would need to process relatively long prompts). Moreover, our experiments
reveal that variation is less prominent in model answers in the context of retrieval augmented QA (e.g.,~the QA model is more confident on incorrect answers supported by retrieved passages in the prompt).
Our contributions can be summarized as follows:

\begin{itemize}[noitemsep,nolistsep]
\item We  quantify QA model uncertainty via estimating
  the utility of the passages it is provided with.

 \item We (contrastively) train a small neural model on utility scores
   obtained through combining accuracy (is the generated answer
   correct?)  and entailment (is the generated answer supported by the passage?) metrics.
   
  \item Our approach is lightweight, improves upon more expensive
    sampling-based methods, and is not tied to the retriever (and
    passages) used to prompt the QA model.

  \end{itemize}

\section{Related Work}

%\textcolor{purple}{
\paragraph{Uncertainty Quantification for Question Answering}
 Several methods have been proposed to predict answer uncertainty in the context of closed-book QA; however, none of them has analysed uncertainty in retrieval augmented QA models.  
 Many of these methods rely on the assumption
 that variation in the model's output is an expression of its uncertainty
 \citep{kuhn2023semantic,farquhar2024semantic-hallu,chen-mueller-2024-quantifying}. For example, some approaches \citep{kuhn2023semantic,farquhar2024semantic-hallu} first cluster answers with similar meaning (in a sample) via Natural Language Inference (NLI) before computing entropy while other work \citep{chen-mueller-2024-quantifying} focuses on \emph{black-box} models; they also compute similarities in the set of answers but associate them with the model's self-judgement of confidence. 
In a recent study, \citet{soudani-etal-2025-uncertainty} show that, in the context of retrieval augmented QA, these methods tend to exhibit overconfidence and sensitivity to the input context. Indeed, our experiments corroborate their analysis. We find that sampling-based methods exhibit less variation in a retrieval setting, and do not offer a significant advantage over methods such as perplexity or LLM prompting for self-assessment \citep{kadavath2022-ptrue}, unlike in closed-book settings.

Sequence entropy methods \citep{kuhn2023semantic,farquhar2024semantic-hallu} focus on detecting incorrect answers stemming from arbitrary fluctuations in model outputs (referred to as confabulations). Our approach extends to additional error sources, including incorrect training data or misleading evidence. 
 \citet{hou2024decomposing} focus on quantifying aleatoric uncertainty (i.e.,~uncertainty in the data) caused by ambiguous questions, an approach that could be combined with ours. 
 Sampling-based methods are expensive to run at inference time for a production QA system; they require several inference steps in addition to performing similarity computations, which can become more complex with longer answers \citep{zhang-etal-2024-fine}.
 In contrast, our approach is light-weight at inference time, but requires training data with observations pertaining to the accuracy of the target model. Our approach optionally uses an NLI model at training time, depending on the chosen objective.
 In Section~\ref{sec:discussion}, we further summarise key differences between our approach and existing uncertainty estimation methods in the context of retrieval augmented QA.
 %}

\paragraph{Judging the Quality of Retrieved Passages}

Previous work has analysed the quality of retrieved passages
\citep{yu2023CoN,asai2024selfrag,wang2024rear,xu2024recomp,yoran2024making}
as they can be irrelevant or misleading.  \citet{asai2024selfrag} make
use of an external critic model to create training data exemplifying cases where a question requires retrieval (or not) and, in the case that retrieval is needed, whether retrieved passages contain the information (or not) 
to formulate the answer.
Note that the usefulness judgment is made by an external critic.
Then, a QA model (i.e., the Self-RAG model) is trained on this data to
learn to reflect by itself whether passages are supportive and
relevant and to predict special tokens indicating this. While it is
possible to derive uncertainty from those special tokens'
probabilities, they only reflect Self-RAG's uncertainty state. Our
proposal is more general and aims at predicting answer uncertainty in
zero-shot QA models (e.g., instruction fine-tuned LLMs). Thus, using
Self-RAG special tokens would be like using an off-the-shelf
sophisticated classifier (i.e., a specialized textual entailment) to
predict the uncertainty of a zero-shot QA model \citep{yoran2024making}. 
In contrast, we predict uncertainty by observing the 
errors of the target QA model (in a zero-shot setting). 
Other work creates auxiliary tasks around retrieved passages enforcing the
QA model to reason on them; e.g., by taking notes about each passage
\citep{yu2023CoN} or generating passage summaries
\citep{xu2024recomp}. These methods also use extreme-scale LLMs to
generate training data for \emph{fine-tuning} a retrieval augmented QA
model. \citet{park-etal-2024-enhancing} select 
 low and high quality in-context examples
in order to instruct the LLM to reason on input passages.  Concurrent
work by \citet{joren2025sufficientcontextnewlens} defines the concept
of context sufficiency, i.e., whether the content in the set of
retrieved passages is sufficient to answer the question and uses an
external judge to assess this. However, the external judgment might not agree with the answering behavior of the target QA model (i.e.,~the judge may indicate that the context is sufficient but the model
may still incorrectly answer).  All these approaches aim at improving
QA performance while our primary goal is modeling QA uncertainty.

\paragraph{Using a Separate Model to Predict Confidence} 
Our passage utility predictor is related to methods aiming to estimate
error \emph{directly} \citep{lahlou2023deup}, e.g., by training a
secondary model to estimate target model loss; instead, our predictor is trained with sequence-level metrics, i.e.,~accuracy and entailment, which
measure error \emph{indirectly}.
Some work \citep{kamath-etal-2020-selective,zhang-etal-2021-knowing} predicts
answer correctness in the context of Reading Comprehension (the
task of generating an answer based on a single supportive
passage). However, as there is no retrieval involved, the input
passage is by default useful, and the main goal is to detect answer
uncertainty due to the QA model being under-trained. In our setting,
the number and quality of passages varies leading to additional sources of
uncertainty (e.g., misleading information).
Some approaches train a specific model to predict answer confidence scores
\citep{dong-etal-2018-confidence,kamath-etal-2020-selective,zhang-etal-2021-knowing,mielke-etal-2022-reducing}
by incorporating various features from the input and model output.
Our approach predicts answer uncertainty directly from individual
passage utilities and its predictions could also be combined with 
other features (e.g., output sequence probability).

\section{Modeling Answer Uncertainty}

We formally define retrieval augmented QA as follows. Given
question~$x$ and a set of retrieved passages $R=\{p_1, p_2, \cdots,
p_{|R|}\}$ obtained with retriever~$\mathcal{R}$, an LLM-based QA
model~$\mathcal{M}$ is prompted to generate answer~$y$ to question~$x$
token-by-token according to its predictive distribution:
\begin{equation}
\hspace*{-.5ex}P(y | x, R; \mathcal{M}) = \prod_{t=1}^{|y|} P(y_t|y_{1..t-1}, x, R; \mathcal{M}).
\label{eq:most:likely:answer}
\end{equation}
We wish to estimate ~$\mathcal{M}$'s  uncertainty (i.e.,~chance of error)
of generating~$y$ given~$x$ and~$R$.

When a retrieved passage is useful to answer a given question (such as
the first passage in Figure~\ref{fig:ragu:data:ex} for the question
\textsl{Who sings Does He Love Me with Reba?}), the QA model is likely
to be confident when generating the answer (\textsl{Linda
  Davis}). When the passage is not useful (such as the second
passage in Figure~\ref{fig:ragu:data:ex}), the QA model is likely to be
uncertain and provide an incorrect answer (\textsl{Patti LaBelle}).
Our hypothesis is that the utility of each passage~$p$ in~$R$ is
indicative of the QA model's uncertainty in generating~$y$, when
prompted with~$R$. If there are passages in $R$ with high utility
(e.g., in Figure~\ref{fig:ragu:data:ex}, the first passage is useful
to answer the question), it is likely that the QA model will be
confident when generating answer~$y$. In contrast, if all passages in
$R$ have low utility, it is likely that the QA model will be uncertain
when generating the answer.

The core of our approach is estimating the
utility~$\upsilon_\mathcal{M}$ of individual passages for a target QA
model~$\mathcal{M}$. Specifically, we develop an estimator $\{x, p\}
\mapsto \upsilon_\mathcal{M}(\{x, p\})$ for each passage~$p \in R$
(Section~\ref{sec:utility:ranker}). We then leverage the predicted
passage utility~$\upsilon_\mathcal{M}$ to estimate $\mathcal{M}$'s
uncertainty when generating answer~$y$ to question~$x$ based on
evidence~$R$, $\{x, R\} \mapsto \text{\textbf{u}}_{\mathcal{M}}(\{x,
R\})$ (Section~\ref{sec:ragunc}).

\subsection{Passage Utility Prediction}
\label{sec:utility:ranker}

%\paragraph{Passage Utility}
Intuitively, a retrieved passage~$p$ is useful for a QA
model~$\mathcal{M}$ if~$\mathcal{M}$ can correctly answer question~$x$
when prompted with~$p$. However, the model's dependence on~$p$ may
vary. In some cases, $\mathcal{M}$~may generate the correct answer
even if $p$~does not explicitly contain it, instead it positively
primes the model to draw upon its memorized knowledge.  In
Figure~\ref{fig:ragu:data:ex}, the first passage has high utility
because the QA model generates a correct answer when prompted with it,
and  explicitly states that ``Linda Davis sings alongside Reba
McEntire''. In contrast, the second passage, while related
to the question's topic, is not useful. The QA model struggles to
answer correctly when prompted with it, suggesting
uncertainty. Since this passage does not provide helpful information
and leads to incorrect answers, its utility is low.

We estimate the utility of passage~$p$ in answering question~$x$ for
QA model~$\mathcal{M}$ by combining two key measures, accuracy and
entailment:
\begin{equation}
  \upsilon_{\mathcal{M}} = \frac{(a(y) + e(y))}{2}
  \label{eq:utility:score}
\end{equation}

Accuracy, denoted as~$a(y)$, indicates whether the
generated answer~$y$ is correct, while entailment, denoted as~$e(y)$,
measures the degree to which~$p$ supports~$y$. Accuracy is determined
by an evaluator~$A$, and entailment is assessed using an NLI model~$E$. The combined
passage utility~$\upsilon_{\mathcal{M}}$
ranges between~0 and~1, given that $a$ takes values in $\{0, 1\}$ and
$e$ falls within the~$[0, 1]$ interval. The intuition behind this composite term $v_{\mathcal{M}}$ is to encourage a ranking of passages that spans a broad spectrum: from highly useful ones (producing correct answers with strong entailment), through more ambiguous cases (e.g., correct answers with weak entailment or incorrect answers with strong entailment), to clearly unhelpful passages (producing incorrect answers with weak entailment).\footnote{Note that for the pairwise ranking loss, using the sum $\upsilon_{\mathcal{M}} = a(y) + e(y)$ would be equivalent. In Appendix~\ref{sec:app:objective:abl}, we include ablation experiments with different implementations of the Passage Utility score $\upsilon_{\mathcal{M}}$.}

We train a lightweight neural model on dataset $D_{\mathcal{M}} =
\{(x, p, \upsilon_{\mathcal{M}})\}$ to predict passage utility scores,
$\{x, p\} \mapsto \upsilon_\mathcal{M}(\{x, p\})$, We construct~$D$ by
pairing input questions~$x$ and passages~$p$ with utility
scores~$\upsilon_{\mathcal{M}}$ which we obtain after
running~$\mathcal{M}$ on examples ${(x, p)}$ and computing observed
answer accuracy and entailment scores from the QA
model~$\mathcal{M}$. We retrieve $|R|>1$ passages per question to
ensure a diverse range of usefulness and create training instances
$\{(x, p_i, \upsilon_{i}) \, | \, p_i \in R\}$ under model
$\mathcal{M}$.  We leverage this data to train the passage utility
predictor with a contrastive learning scheme.  Specifically, if two
passages~$p_i$ and $p_j$ belong to~$R$ and $p_i$ is more useful
than~$p_j$ for answering question $x$, the predicted utility
score~$\upsilon_{i}$ should be higher than~$\upsilon_{j}$ by
margin~$m$, ensuring that~$p_i$ is ranked above~$p_j$.  We train the
utility predictor with the following ranking objective:
%
%{\footnotesize
\begin{equation}
  \mathcal{L}_{rank} = 
  \hspace*{-1cm}\sum\limits_{((x,p_i), (x, p_j)) \in R \times R, i \neq j} \hspace*{-1cm}\max(0,
  -z(\upsilon_{i} - \upsilon_{j}) + m)),
\label{eq:rank}
\end{equation}
%}

\noindent
where $z$ controls the gold order between~$p_i$ and~$p_j$ (e.g., if
$z=1$, $p_i$ has higher utility, and conversely $z=-1$ indicates the
opposite ordering) and $m$~is a hyper-parameter.

We train the passage utility predictor with a Siamese neural network
consisting of a BERT-based encoder \citep{devlin-etal-2019-bert}
followed by  pooling and two MLP layers stacked on top of BERT
outputs \citep{fang-etal-2024-efficiently}. The output layer computes
the utility score as $\upsilon_{i} = W_o h^L + b_o$ where $h^L$ is the
vector representation for $(x, p_i)$ from the last hidden layer (the
$L$-th layer) of the network. At inference time, we compute a single
utility score for each passage. We provide implementation and training
details in Section~\ref{sec:experimental}.

To strengthen the role of accuracy prediction as a training signal and
regularize the range of utility values learned by the ranking scheme,
we combine the ranking objective in Equation~(\ref{eq:rank}) with the
following Binary Cross Entropy (BCE; \citealt{combined:sculley})
objective:
\begin{equation}
\begin{aligned}
\mathcal{L}_{BCE} = \sum\limits_{((x,p_i), (x, p_j)) \in R \times R, i \neq j} [ a\times \log(p(x, p_i)) +  (1 - a) \times \log(1-p(x, p_i))] \hspace{1cm} \\ + [ a\times \log(p(x, p_j)) +  (1 - a) \times \log(1-p(x, p_j))], 
\label{eq:bce}
\end{aligned}
\end{equation}
where $p(x, p_i) = \operatorname{sigmoid}(\upsilon_i)$ and $a$ is the target accuracy label $a(y)$ under model $\mathcal{M}$ taking values in the set $\{0, 1\}$. We train the utility predictor with the following combined objective where $\lambda$ is a hyper-parameter to adjust the $\mathcal{L}_{BCE}$ penalty:
\begin{equation}
    \mathcal{L} = \mathcal{L}_{rank} + \lambda \, \mathcal{L}_{BCE}.
\label{eq:combined}    
\end{equation}

Both ranking and BCE objectives are compatible with gold annotations
that could be provided in   active learning or interactive settings
\citep{text:ranking:bo,fang-etal-2024-efficiently}. For example,
moderators of the QA system would provide judgments on the accuracy of
the answers it predicts (e.g., \textsl{correct/incorrect}) and the
extent to which these are supported by the retrieved passages (e.g.,
\textsl{not} supported to \textit{fully} supported).

\subsection{Answer Uncertainty Estimation}
\label{sec:ragunc}

For our QA task, we want to define an estimator $\{x, R\} \mapsto
\text{\textbf{u}}_\mathcal{M}(\{x, R\})$ which quantifies the
uncertainty of model~$\mathcal{M}$ when generating  answer~$y$ for
question~$x$ based on a prompt with passages~$R$.  We propose
estimating~$\text{\textbf{u}}_\mathcal{M}$ directly from the utility
scores of individual passages in $R$.  The key intuition is that the
higher the utility of one (or more) passages, the less uncertain
$\mathcal{M}$ will be when generating answer~$y$.  
Conversely, if all passage utilities in $R$ are low, it is more likely that $\mathcal{M}$ will be uncertain about the answer.
Specifically, we estimate  answer uncertainty~$\textbf{u}_\mathcal{M}$ by taking the
maximum utility score among the passages in $R$:
\begin{equation}
\textbf{u}_\mathcal{M}(\{x, R\}) = \max({\upsilon_\mathcal{M}(\{x, p\}) \, | \, p \in R}).  
\label{eq:unc:aggregation}
\end{equation}

Our approach to aggregating passage utilities is  intuitive and simple. 
However, QA models might be sensitive to  factors relating to how they are prompted, such as the number or order of passages  \citep{liu-etal-2024-lost,xie2024adaptive}. In Appendix~\ref{sec:app:unc:aggregation}, we examine such confounds more closely, comparing question answering accuracy when 
 models are prompted with individual passages in~$R$ (our aggregation approach)
versus the entire set~$R$. The study shows that there is little disagreement between the two methods and that it is possible to approximate answer uncertainty when prompting with $|R|$ passages while avoiding the combinatorial complexity of estimating uncertainty over all possible combinations of input passages.

\section{Experimental Setup}
\label{sec:experimental}

\subsection{QA Tasks and Models}

%\paragraph{QA Tasks} 

We evaluate our approach to predicting answer uncertainty on
short-form question answering tasks. Specifically, on the following six datasets: Natural Questions \citep{kwiatkowski-etal-2019-natural}, TriviaQA
\citep{joshi-etal-2017-triviaqa}, WebQuestions
\citep{berant-etal-2013-semantic}, SQuAD
\citep{rajpurkar-etal-2016-squad}, and PopQA
\citep{mallen-etal-2023-popQA}.  We also evaluate on RefuNQ
\citep{refunq}, a dataset with unanswerable questions about
non-existing entities. In Appendix~\ref{sec:app:datasets}, we describe
each dataset, provide example questions, and make available details
about the splits used in our experiments which follow \cite{lee-etal-2019-latent}.

%\paragraph{QA Models} 

We consider backbone instruction fine-tuned LLMs from different
families of similar size. These are Llama-3.1-8B-Instruct
\citep{llama3modelcard}, Mistral-7B-Instruct-v0.3
\citep{jiang2023mistral7b}, and Gemma2-9B-it \citep{Riviere2024Gemma2I}.
We also assess answer uncertainty quantification performance on QA
models of the same family but with different sizes. To this end, we
additionally evaluate on Gemma2-27B-it.  For all QA
models, we use a simple prompt including the retrieved passages and
the question in the context; the prompt is shown in
Table~\ref{tab:ragu_prompts} in the Appendix.  The QA models' answer
is the most likely answer generated with greedy sampling at
temperature equal to~0.  Following previous work on retrieval
augmented QA, we use Contriever-MSMARCO \citep{izacard2022unsupervised}
as our external retriever \citep{asai2024selfrag} and the target QA
models are prompted with $|R|=5$ passages
\citep{yu2023CoN,asai2024selfrag,xu2024recomp}.  In
Appendix~\ref{sec:app:implem:det}, we provide more details about
inference and passage retrieval.

\subsection{Evaluation}

\paragraph{QA Accuracy}

A precise metric for measuring accuracy is key when evaluating the quality of uncertainty estimation.
Token overlap metrics are imprecise and can over- or under-estimate accuracy (e.g.,~\textsl{5} will not match \textsl{five}).
Thus, we rely on a LLM-based  accuracy evaluator to create training data for the Passage Utility predictor (i.e., $A$ in Section~\ref{sec:utility:ranker}) and to evaluate retrieval augmented QA performance. 
We use Qwen2-72B-Instruct \citep{qwen2} and the prompt  proposed by \citet{sun-etal-2024-head} to obtain accuracy judgments. Details about the LLM evaluator can be found in Appendix~\ref{sec:app:implem:det}.

\paragraph{Uncertainty Estimation}

To assess the quality of answer uncertainty prediction, we follow
\citet{farquhar2024semantic-hallu} and report the Area Under the
Receiver Operator Curve (\textbf{AUROC}) on detecting incorrect
answers (i.e.,~answers with high uncertainty).  In
Appendix~\ref{sec:app:additionalres:unc}, we also report the area
under the ‘rejection accuracy’ curve (\textbf{AURAC}) which captures
the accuracy a model would have if it refused to answer questions with
highest uncertainty. Rejection accuracy is essentially the model's
accuracy on the remaining questions. 
In the main results section, we focus on selective answering performance 
when models answer 80\% of the least uncertain questions versus when 
always answering. We provide implementation details in
Appendix~\ref{sec:app:implem:det}.

\subsection{Methods}

\paragraph{Passage Utility Predictor}

We train a Passage Utility predictor per QA model and QA task. For each task, we curate dataset $D_{\mathcal{M}}  = \{(x, p, \upsilon_{\mathcal{M}})\}$ to train and evaluate a Passage Utility predictor for QA model $\mathcal{M}$.
We use the training (and development) questions available for each QA task, considering the top five retrieved passages for each question (i.e., $|R|=5$). Note that~$|R|$ is a hyper-parameter and other values would be also possible. Larger sizes of $|R|$ would yield more training data, since the Utility predictor takes individual passages (together with the question) as input. 
The target QA model~$\mathcal{M}$  is first prompted with passage~$p
\in R$ and question~$x$ to generate  answer~$y$. Then, we annotate
passages~$p$ with a utility score computed with the accuracy evaluator~$A$ and  entailment judge~$E$ on generated answer~$y$
(Section~\ref{sec:utility:ranker}). We use an ALBERT-xlarge model \citep{Lan2020ALBERT}  optimized on MNLI \citep{williams-etal-2018-broad} and the VitaminC dataset \citep{schuster-etal-2021-get}.
We provide more details about the curated datasets and training of the Passage Utility predictor training in Appendix~\ref{sec:app:implem:det}.

\paragraph{Comparison Approaches and Baselines}

There exist several uncertainty estimation methods which we group in
two categories based on whether they require access to logits or
simply model outputs (see \citealt{polygraph} for additional
methods). We choose the highest scoring ones to compare with here and
include additional results in Appendix~\ref{sec:app:additionalres:unc}
for completeness.

\emph{Information-Theoretic Measures.} We compare against uncertainty estimation methods that are based on the predictive probabilities of the target QA model. Let~$y$ denote an answer generated with probability~$p(y | x, R; \mathcal{M})$ which is computed as: 
\begin{equation}
\begin{aligned}
p(y | x, R; \mathcal{M}) = \prod_{t=1}^{|y|} p(y_t|y_{1..t-1}, x, R; \mathcal{M})
\end{aligned}
\end{equation}
The \textbf{Perplexity} (PPL) of model~$\mathcal{M}$ boils down to calculating \emph{token-level} entropy as it is based on the average negative log-likelihood of the generated tokens:
\begin{equation}
\begin{aligned}
\text{PPL}(x, R, \mathcal{M}) =  \exp{\{- \, \frac{1}{|y|} \, \sum^{|y|}_{t=1} \, \log \, p(y_t | y_{1..t-1}, x, R; \mathcal{M}) }\},    
\end{aligned}
\end{equation}

%Perplexity essentially calculates \emph{token-level} entropy as it is based on the average negative log-likelihood of the generated tokens. 

\textbf{Regular entropy}, on the other hand, is computed over \emph{sequences},  quantifying the entropy of the answers. It is defined as $\mathbb{E}[-\text{log} \, P(Y | x, R; \mathcal{M})]$ where the expected value,  $\mathbb{E}$, is computed on sequences $y$ sampled from the conditional distribution $P(Y | x, R; \mathcal{M})$, where random variable $Y$ denotes the answer sequences, and $x$ and $R$ are the input question and retrieved passages, respectively. 
In practice, regular entropy is approximated via  Monte-Carlo integration, i.e.,~sampling $N$ random answers from $P(Y | x, R; \mathcal{M})$: 
\begin{equation}
\begin{aligned}
\text{RE}(x, R, \mathcal{M}) = -\frac{1}{N} \, \sum_{n=1}^N \text{log} \, \tilde{P}(y^{n} \, | \, x, R; \mathcal{M}), 
\end{aligned}
\end{equation}
where $\tilde{P}(y^{n} \, | \, x, R; \mathcal{M})$ is the length normalised version of $P(y^{n} | x, R; \mathcal{M})$.

\citet{kuhn2023semantic} propose \textbf{Semantic Entropy}, a variant of
regular entropy that disregards uncertainty related to the surface
form of the generated answers.  The method works by sampling several
possible answers to each question and grouping the set of~$N$ samples
into~$M$ clusters (with $M \leq N$) that have similar meanings (which
are determined on the basis of whether answers in the same cluster
entail each other bidirectionally). The average answer probability
within each cluster is:
\begin{equation}
\begin{aligned}
\text{SE}(x, R, \mathcal{M}) = - \sum^M_{m=1} \hat{P}_m(x, \mathcal{M}) \, \text{log} \, \hat{P}_m(x, \mathcal{M}),
\end{aligned}
\end{equation}
where
$\hat{P}_m(x, \mathcal{M})$ is estimated as follows:
\begin{equation}
 \hat{P}_m(x, \mathcal{M})   = \frac{ \sum_{y \in C_m} \, \tilde{P}(y | x, R; \mathcal{M})}{\sum^M_{m=1} \,  \sum_{y \in C_m} \, \tilde{P}(y | x, R; \mathcal{M})}
 \end{equation}

\emph{LLM-based Measures.} We compare with \textbf{p(true)}  which  uses the
same LLM-based target QA model  to assess whether the answers it
produces are correct \citep{kadavath2022-ptrue}. We follow the p(true)
variant used in previous work \citep{kuhn2023semantic}. The QA model
is prompted with the question and a set of candidate answers (consisting of 
the most likely answer and  a sample of~$N$ answers) and is instructed to
respond whether the most likely answer is true or false (i.e.,
correct/incorrect). The score produced by this approach is the
probability of  model~$\mathcal{M}$ generating the token True.  
This method needs several in-context examples to work well;
following \citet{kuhn2023semantic}, we use 20 in-context examples. Note
that since our backbone LLMs are recent models with a large context
(unlike \citealt{kuhn2023semantic}), all~20 examples are fed in the
context making p(true) an expensive but very strong approach. 
In the context of retrieval augmented QA, we modify p(true) to include in the prompt the set of retrieved passages for the question of interest.
We provide the prompt used by p(true) in
Appendix~\ref{sec:app:prompts}. Note that p(true) can be considered as
a specialized powerful entailment model and thus we do not include
entailment based methods relying on off-the-shelf NLI models which have been
shown to perform poorly \citep{yoran2024making}.

For approaches that require sampling, we follow previous work
\citep{farquhar2024semantic-hallu} and take \mbox{$N=10$}~samples,
which we generate with multinomial sampling. We set the sampling
temperature to~1, with nucleus sampling ($P=0.9$; \citealt{nucleus})
and top$-K$ sampling ($K=50$; \citealt{fan-etal-2018-hierarchical}), and
use a different random seed to draw each sample.  We provide further
details about inference in Appendix~\ref{sec:app:implem:det} and
report inference costs for each approach in
Appendix~\ref{sec:app:app:cost}.

\section{Results and Analysis}
\label{sec:results}

\paragraph{Passage Utility is effective across
  model families, sizes, and QA tasks.} 
  Table~\ref{tab:main:unc:test} summarizes our uncertainty estimation results (test set) with four  QA models (\textsc{Gemma2-9B/27B},
\textsc{Llama-3.1-8B}, and \textsc{Mistral-7B-v0.3}) across  six
QA datasets (results on the
development set are included in Appendix~\ref{sec:app:additionalres:unc}).
We boldface the highest AUROC value for each QA model and dataset pair and
mark with~$^*$ the next best value that is significantly different from it at \mbox{$p < 0.05$}.
We use the paired De Long test \citep{delong-test} to compute whether 
 two AUROC values are significantly different.\footnote{We use the library in \url{https://github.com/Brritany/MLstatkit}
to compute significance scores.}

In general, answer perplexity (PPL) performs rather poorly, especially with
\textsc{Gemma2-9B/27B}. Perplexity is likely to underperform with less calibrated models, such as those which have undergone instruction tuning \citep{tian-etal-2023-just}.
Regular Entropy shows little improvement upon PPL but by ignoring 
surface form choices and focusing on meaning,  Semantic Entropy improves AUROC scores.
p(true) performs well at detecting answer uncertainty matching or
surpassing Semantic Entropy. 
Overall, we observe that the gap among these methods' 
performance is lower than in the context of closed-book QA studied 
in previous work \citep{farquhar2024semantic-hallu}. 
We hypothesize that, on one hand, our recent QA models admitting 
more in-context examples benefiting p(true) and, on the other hand,
that retrieved passages in the prompt make QA models' outputs
less varied.
Our Passage Utility approach performs on par or outperforms
all other methods with a \emph{single small-model inference step} 
on each input passage. 

Passage Utility performs particularly well on challenging question
answering tasks represented by datasets like PopQA and
RefuNQ. In these cases, our light-weight uncertainty estimation model
works better than p(true) which requires the same QA model (i.e., the
same backbone LLM) to judge the correctness of its own generated
answers. We speculate that for questions with high uncertainty,
i.e.,~where the model does not have the knowledge to answer
(e.g.,~questions about non-existing concepts in RefuNQ), it
confidently generates a response and also fails at assessing it.  
We attribute the Passage Utility's success to the fact that it has been
specifically trained to detect situations where the target QA model is
prone to answer incorrectly (i.e., when provided with retrieved
passages of lower relevance).
The six QA tasks pose different retrieval challenges. On TQA, retrieval results are often of good quality: for 73\% of the questions, the top-5 retrieved passages contain the gold answer string.  In contrast, on PopQA the percentage reduces to 63\% and on  RefuNQ the quality of retrieval is deliberately low (as it consists of unanswerable questions). 
Across models (Table~\ref{tab:main:unc:test}), our approach is comparable to p(true) when retrieved passages  contain the answer and excels in cases of low quality retrieval.

\begin{table}[t]
%\begin{minipage}[b]{.53\linewidth}
\centering
{\footnotesize
  \begin{tabular}{@{}l@{\hspace{3pt}}|c@{\hspace{2pt}}c@{\hspace{2pt}}c@{\hspace{3pt}}c@{\hspace{3pt}}c@{\hspace{3pt}}c@{\hspace{3pt}} 
  c@{\hspace{3pt}}|c@{\hspace{2pt}}c@{\hspace{2pt}}c@{\hspace{3pt}}
  c@{\hspace{3pt}}c@{\hspace{3pt}}c@{\hspace{3pt}}c@{}}
    \toprule %\ \\[-1ex]
    \multicolumn{1}{l}{} & \multicolumn{7}{c@{}}{\textsc{Gemma2-9B}} & \multicolumn{7}{c@{}}{\textsc{Gemma2-27B}} \\
         & {\scriptsize NQ} & {\scriptsize TQA} & {\scriptsize WebQ}  
         & {\scriptsize SQuAD} & {\scriptsize PopQA}  & {\scriptsize RefuNQ}
         & {\scriptsize AVG} 
         & {\scriptsize NQ} & {\scriptsize TQA} & {\scriptsize WebQ}  
         & {\scriptsize SQuAD} & {\scriptsize PopQA} & {\scriptsize RefuNQ} 
         & {\scriptsize AVG} \\

\midrule %\ \\[-1.9ex]
PPL & 0.64 & 0.68 & 0.52 & 0.53 & 0.59 & 0.51 & 0.58 %G9B
 & 0.64 & 0.50 & 0.53 & 0.59 & 0.58 & 0.51 & 0.56 \\ %G27B
p(true) & 0.73 & \hspace{0.5em}0.75$^*$ & 0.67 & 0.63 & \hspace{0.5em}0.81$^*$ & 0.62 & 0.70  %G9B
 & \textbf{0.77} & \textbf{0.83} & 0.67 & \hspace{0.5em}0.68$^*$ & \hspace{0.5em}0.78$^*$ & \hspace{0.5em}0.60$^*$ & 0.72 \\ %G27B
Regular Entropy & 0.66 & 0.69 & 0.54 & 0.56 & 0.61 & 0.51 & 0.60   %G9B
 & 0.67 & 0.54 & 0.55 & 0.59 & 0.62 & 0.51 & 0.58 \\ %G27B
Semantic Entropy & 0.70 & 0.73 & \hspace{0.5em}0.57$^*$ & \hspace{0.5em}0.64$^*$ & 0.73 & \hspace{0.5em}0.59$^*$ & 0.66 %G9B
 & 0.69 & \hspace{0.5em}0.62$^*$ & \hspace{0.5em}0.59$^*$ & 0.63 & 0.66 & 0.58 & 0.63 \\%G27B
Passage Utility & \textbf{0.76} & \textbf{0.85} & \textbf{0.69} & \textbf{0.78} & \textbf{0.86} & \textbf{0.79} & 0.79 %G9B
 & \hspace{0.5em}0.73$^*$ & 0.82 & \textbf{0.69} & \textbf{0.80} & \textbf{0.85} & \textbf{0.78} & 0.78 \\%G27B

\bottomrule %\ \\[-1.9ex]
\multicolumn{15}{c}{} \\\toprule
\multicolumn{1}{l}{} &\multicolumn{7}{c@{}}{\textsc{Llama-3.1-8B}} 
 &\multicolumn{7}{c@{}}{\textsc{Mistral-7B-v0.3}} \\
& {\scriptsize NQ} & {\scriptsize TQA} & {\scriptsize WebQ}  
         & {\scriptsize SQuAD} & {\scriptsize PopQA}  & {\scriptsize RefuNQ}
         & {\scriptsize AVG} 
         & {\scriptsize NQ} & {\scriptsize TQA} & {\scriptsize WebQ}  
         & {\scriptsize SQuAD} & {\scriptsize PopQA} & {\scriptsize RefuNQ} 
         & {\scriptsize AVG} \\
\midrule
PPL & 0.75 & 0.80 & 0.68 & 0.74 & 0.83 & 0.60 & 0.73  %L8B
 & 0.63 & 0.71 & 0.57 & 0.65 & 0.64 & 0.62 & 0.64 \\ %M7B
p(true) & \textbf{0.79} & \textbf{0.88} & \textbf{0.74} & 0.77 & 0.85 & \hspace{0.5em}0.67$^*$ & 0.78 %L8B
 & 0.73 & 0.82 & \textbf{0.68} & \hspace{0.5em}0.74$^*$ & \hspace{0.5em} 0.75$^*$ & \hspace{0.5em}0.68$^*$ & 0.73 \\ %M7B
Regular Entropy & \hspace{0.5em}0.76$^*$ & 0.81 & \hspace{0.5em}0.71$^*$ & 0.78 & \hspace{0.5em}0.83$^*$ & 0.65 & 0.76  %L8B
 & 0.64 & 0.75 & \hspace{0.5em}0.62$^*$ & 0.65 & 0.66 & 0.60 & 0.65 \\ %M7B
Semantic Entropy & 0.72 & \hspace{0.5em}0.82$^*$ & 0.66 & \hspace{0.5em}0.78$^*$ & 0.81 & 0.59 & 0.73   %L8B
 & \hspace{0.5em}0.66$^*$ & \hspace{0.5em}0.78$^*$ & 0.66 & 0.73 & 0.74 & 0.61 & 0.70 \\ %M7B
Passage Utility & 0.77 & 0.82 & 0.72 & \textbf{0.83} & \textbf{0.87} & \textbf{0.81} & 0.80   %L8B
& \textbf{0.74} & \textbf{0.83} & \textbf{0.68} & \textbf{0.82} & \textbf{0.85} & \textbf{0.80} & 0.79 \\ %M7B

\bottomrule
\end{tabular}
}
%\vspace*{-.2cm}
\captionof{table}{AUROC values for QA models based on \textsc{Gemma2-9B/27B},
  \textsc{Llama-3.1-8B}, and \textsc{Mistral-7B-v0.3} on Natural
  Questions (NQ), TriviaQA (TQA), WebQuestions (WebQ), SQuAD, PopQA,
  and RefuNQ test sets. The best values (per model and dataset) are highlighted in \textbf{bold}; we also mark with $ ^*$ next best values which are significantly different using the paired De Long test ($p < 0.05$). For example, on TQA with \textsc{Gemma2-9B}, p(true), the second best performing is significantly different from the Passage Utility which performs best and by extension models with lesser values than p(true) are also significantly different. }
\label{tab:main:unc:test}
\end{table}

 Passage Utility also performs well with different QA model sizes (within the same family), i.e., \textsc{Gemma2-9B} and \textsc{27B} (Table~\ref{tab:main:unc:test}). 
We observe a noticeable decrease in performance for most information-theoretic models when using the biggest \textsc{Gemma} model (27B). We attribute this to the fact that the 27B model more confidently makes less errors and its calibration may be affected more by the fine-tuning step \citep{tian-etal-2023-just}. p(true), on the other hand, benefits from the largest model's context understanding and memorized knowledge.

It is important to note that our approach is lite-weight at inference time. In Figure~\ref{fig:unc:tokens:size} we report average AUROC per method with respect to the number of input tokens at inference time and the number of parameters involved in uncertainty estimation. We report scores for
\textsc{Gemma2-9B} and its bigger version \textsc{Gemma2-27B}. As can be
seen, our approach which is  based on a BERT encoder \citep{devlin-etal-2019-bert} 
with 110M parameters and a small number of input tokens achieves on aggregate
better performance than more expensive approaches. 
For the \textsc{27B} QA model, p(true) edges  closer to passage utility, however, at the expense of thousands of input tokens and $\sim$26 billions more parameters. This indicates that p(true) will be less efficient in QA settings  where latency and cost are critical. In Appendix~\ref{sec:app:app:cost} we provide a cost analysis in terms of model calls.

\begin{figure}[t]
\centering
\begin{minipage}[b]{.49\textwidth}
  \centering
    \include{CostApproachGraph3}
    \vspace{-2em}
  \captionof{figure}{Average AUROC  across our six development sets (y-axis) with respect to number of input tokens at inference time (x-axis) and number of parameters (size of the circles). We present results for perplexity (PPL), Semantic Entropy (SE), p(true), and Passage Utility (PU).
  We exclude Entropy (which is close to SE) for readability.
  We compare the smaller \textsc{Gemma2-9B} and its bigger version \textsc{27B}. Thinner circles positioned in the left corner are better.}
    \label{fig:unc:tokens:size}
\end{minipage}
\hfill
\begin{minipage}[b]{.46\textwidth}
  \centering
    \include{ModelAcc80Graph}
    \vspace{-2.5em}
  \captionof{figure}{Average accuracy  (across our six test sets) with 
  \textsc{Gemma2} sizes 9B and 27B (G9B, G27B), \textsc{Llama-3.1-8B} (L8B), and \textsc{Mistral-7B-v0.3} (M7). Black dots: QA models always answer; colour dots: QA models answer 80\% of the cases they are most confident about.  }
  \label{fig:unc:model:acc80}
\end{minipage}
\end{figure}

\paragraph{Passage Utility leads to selective answering.}

Model uncertainty can be used to decide whether to provide an answer to a question or not. 
Figure~\ref{fig:unc:model:acc80} shows average accuracy when
 the target QA models choose to answer~80\% of the cases they are most
 confident about. For comparison, we also show QA accuracy when always
 answering, i.e.,~black bold dots.  
 All uncertainty quantification approaches improve model accuracy.
 Across different LLM QA backbones, Passage Utility performs on par with or 
 better than more expensive uncertainty estimation approaches such as p(true).
  For instance, when looking at selective performance according to Passage
 Utility, the biggest \textsc{Gemma2-27B} model improves by~+9 points
 (0.74~vs~0.65). In Appendix~\ref{sec:app:additionalres:unc} we report the 
 full set of selective accuracies at different thresholds.

\paragraph{Passage Utility shows potential for retrieval reranking.}

To further assess the quality of the Passage Utility scores and to
highlight their potential for retrieval reranking we carry out the
following ablation study. Since most previous work \citep{asai2024selfrag,xie2024adaptive,yoran2024making} on retrieval augmented QA prompts the QA model with 
the top-$5$ (or less) input passages,  
we hypothesize that our passage utility score could be an effective
reranking method after retrieval \citep{Nogueira2019MultiStageDR,rankLlama,seakr}.
We test this hypothesis by computing the accuracy of the \textsc{Gemma2-9B} QA 
model when prompted with the top-$k$ passages (with $k$~in the range of 
\{5, 3, 1\}) out of a sample of  $|R|=10$ passages provided by an external
retriever. 

In Figure~\ref{fig:unc:rerank}, we compare performance under two passage-ranking strategies: one based on relevance scores from an external retriever (gray), and the other based on the QA model’s self-assessed utility of individual passages, following \cite{seakr}. 
We report two self-assessment variants, one using the predictions
of the Passage Utility model (red) and another one based on the perplexity 
of the QA model when answering with individual retrieved passages (blue).
Passages that yield answers with lower perplexity should be ranked first.
Figure~\ref{fig:unc:rerank} shows average accuracy  values across 
five development sets (NQ, TQA, WebQ, SQuAD, and PopQA), at different 
cutoff values ($k$). As can be seen, 
the QA model achieves higher accuracy when passages are ranked according to their utility. This finding suggests that Passage 
Utility scores indeed reflect which passages are useful for the target 
QA model.

%%\textcolor{purple}{
\paragraph{Passage Utility struggles with multi-hop questions.}
Our approach estimates uncertainty by predicting individual passage utility and selecting the maximum utility score from a set of retrieved passages. A potential limitation of this method emerges  in multi-hop questions  that require evidence from multiple passages \citep{yang-etal-2018-hotpotqa,pmlr-v174-pal22a-medmcqa}. 
Specifically, the QA model is unable to answer correctly when prompted with any individual passage (resulting in uniformly low utility scores), yet succeeds when given the complete set of passages (in which case the maximum utility score over individual passages fails to reflect this outcome). To quantify this limitation, we evaluate our approach
on the widely used HotPotQA dataset \citep{yang-etal-2018-hotpotqa}, using the splits as provided by \citet{trivedi-etal-2023-interleaving}, and the same retrieval 
settings as defined above (Section~\ref{sec:experimental}).
Table~\ref{tab:main:unc:test:hopo} shows AUROC values for all uncertainty estimation methods. 
In Appendix~\ref{sec:app:mhqa:synth:analysis}, we provide an analysis on a small set of multi-hop questions making use of synthetic HotPotQA data.
%}

%\textcolor{purple}{
Across models,  our approach is on par with p(true) and sequence entropy, and better than
perplexity (Table~\ref{tab:main:unc:test:hopo}). Manual inspection of~100 examples from the development set with \textsc{Gemma2-9B} as the target QA model, reveals two major trends. Firstly,  the QA model frequently (48 cases) manages to correctly answer multi-hop questions using only one of the required 'hop' passages, often not needing the entire set.
This observation aligns with recent findings in \cite{joren2025sufficientcontextnewlens}.
Secondly, in numerous instances (29 cases), the retrieved passages did not contain any useful evidence, leading the model to answer incorrectly even when prompted with the full set. This underscores the inherent difficulty of effective retrieval for complex questions.
% TBC: The other examples (23) consist of one passage that leads to
% the correct answer and the others dont. I loked at 5 of the 23, and
% the one passage leads to the answer but the other passages are
% irrelevant. I will look at more, specially because in these 23
% the QA model answers when prompted with the 5 passages are half
% correct and half incorrect.
Many studies \citep{jeong2024adaptiverag,trivedi-etal-2023-interleaving,Lin_Huang_Zhang_Zhou_Chen_2025} tackle this challenge through sophisticated QA pipelines. Notably, certain approaches decompose complex questions into sub-questions that can be answered independently. In such multi-hop pipelines, our approach could be naturally applied at the sub-question level.
%}

\begin{figure}[t]
\begin{center}
  \begin{minipage}{\textwidth}
  \begin{minipage}[b]{0.48\textwidth}
    \centering
        \include{PassageRerankGraph}
       \vspace{-0.9cm}
    \captionof{figure}{Average RAG accuracy  with
      \mbox{\textsc{Gemma2-9B}} across the five QA development sets. 
      Points on the x-axis correspond to different context 
      sizes, when taking the top-$k$,
      passages according to query relevance (gray) and 
      self-assessment via perplexity (PPL; blue), and 
      Passage Utility  (PU; red).}
    \label{fig:unc:rerank}
  \end{minipage}
  \hspace{1cm}
  \begin{minipage}[b]{0.4\textwidth}
 % \vspace{-0.2em}
      \centering
{\footnotesize
  \begin{tabular}{@{}l@{\hspace{3pt}}c@{\hspace{3pt}}c@{\hspace{3pt}}c@{\hspace{3pt}}c@{\hspace{3pt}}c@{}}\\
  \toprule 
         &  G9B & G27B & L8B & M7B \\
  \midrule \ \\[-1.9ex]
PPL                 &  0.61 & 0.59 & 0.73 & 0.70 \\
p(true)             &  \hspace{0.5em}0.75$^*$ & \textbf{0.78} & \textbf{0.81} & \textbf{0.78} \\
Regular Entropy     &  0.64 & 0.60 & \hspace{0.5em}0.77$^*$ & 0.71 \\
Semantic Entropy    &  0.70 & \hspace{0.5em}0.65$^*$ & 0.74 & 0.73 \\
Passage Utility      &  \textbf{0.78} & 0.77 & \hspace{0.5em}0.77$^*$ & \hspace{0.5em}0.74$^*$  \\
\bottomrule
\end{tabular}
\vfill
}
\captionof{table}{AUROC values for QA models based on \textsc{Gemma2-9B/27B},
  \textsc{Llama-3.1-8B}, and \mbox{\textsc{Mistral-7B-v0.3}} on HotPotQA test set. 
  Best values per model are highlighted in \textbf{bold}; we also mark with$ ^*$ next best values which are significantly different using the paired De Long test (\mbox{$p < 0.05$}). }
\label{tab:main:unc:test:hopo}  
    \end{minipage}
  \end{minipage}
  \end{center}
\end{figure}

\section{Discussion}
\label{sec:discussion}

\paragraph{Key Properties and Usability Scenarios}

Each uncertainty quantification approach comes with its own advantages and limitations. This entails that the choice of a specific method depends on criteria like available resources, desired latency, and the necessary level of control and trustworthiness for the QA system.
For example, in high-stakes applications, a method that favours higher rates of false positives (thereby allowing human intervention) and reduces the chance of overconfident false negatives would be better, even if it requires training data and regular updates.

%\textcolor{purple}{
In Table~\ref{tab:limitations}, we summarise existing features (and limitations) of various uncertainty estimation approaches as we compare them to our work. The first column reports latency, as previously discussed in Figure~\ref{fig:unc:tokens:size}. 
Retrieval augmented QA models can err as a result of being exposed to erroneous sources such as misleading passages \citep{xie2024adaptive} or inaccurate training data \citep{vu-etal-2024-freshllms}. 
Information-theoretic methods are not equipped with an explicit mechanism to deal with these cases \citep{farquhar2024semantic-hallu,soudani-etal-2025-uncertainty}.
Moreover, these methods are also known to suffer from over-confidence \citep{simhi2025trustmeimwrong,soudani-etal-2025-uncertainty,sung-etal-2025-grace}.
While p(true) may be able to detect these challenging cases to a certain extent,  Passage Utility can be \emph{specifically} trained to recognise them.
In terms of supervision, both p(true) and Passage Utility require task-specific training examples, and the performance of both approaches deteriorates on out-of-distribution examples (Table~\ref{tab:main:unc:ood:test}). 
%}

\begin{table}[t]
    \small
    \centering
    \begin{tabular}{@{~}l@{~}|@{~}c@{~}c@{~}c@{~}c@{~}c@{~}}
    \hline
         & \multicolumn{1}{p{1.8cm}}{Low latency} 
         & \multicolumn{1}{p{2.5cm}}{No Training Data} 
         & \multicolumn{1}{p{2.2cm}}{No Fine-tuning} 
         & \multicolumn{1}{p{2.2cm}}{Recognises Erroneous Evidence} 
         & \multicolumn{1}{p{2.2cm}}{Mitigates overconfidence} \\
         \hline
    PPL     & \cmark & \cmark & \cmark & \\
    p(true)     &  & & \cmark & \cmark & \cmark \\
    Semantic Entropy & & \cmark & \cmark & \\
    Passage Utility & \cmark & & & \cmark & \cmark \\
    \hline
    \end{tabular}
    \caption{Categorization of the uncertainty estimation approaches studied in Section~\ref{sec:results} according to different properties (table headers).}
    \label{tab:limitations}
\end{table}

\paragraph{Training Data Requirements}

Our approach requires question-answer pairs to curate a dataset with retrieved passages and utility scores for training. However, general or task specific  training datasets could be generated semi-automatically \cite{li-zhang-2024-planning,wei2024measuringshortformfactualitylarge}. Moreover, in experiments we show that in some QA tasks such as WebQ or PopQA this training data can be relatively small, i.e., in the region of~2.5k or~10k respectively.

\paragraph{Fine-Tuning Requirements}

Our approach, by design, requires fine-tuning to adapt to new QA tasks or models, as its core aim is to model the accuracy behavior of the target QA model. To enhance versatility across QA tasks, a unified training set encompassing diverse QA tasks could be compiled to train a single passage utility predictor. More practically, advanced training schemes promoting generalization, such as meta-learning with a varied set of QA tasks and QA model examples, could be employed to develop a single passage utility predictor.

In addition, while the utility predictor may necessitate recalibration for distinct QA models or tasks, its performance on out-of-distribution scenarios (Appendix~\ref{sec:app:generalisation}) establishes it as a robust warm-up base model. This allows for subsequent fine-tuning with a reduced number of target examples \citep{kamath-etal-2020-selective,zhang-etal-2021-knowing}.
Finally, as the utility predictor relies on a small model, the cost of fine-tuning in terms of resources and time is low.

\section{Conclusions}

In this work we address uncertainty estimation in the
context of retrieval augmented QA with a method that   relies on individual passage utilities. 
Key in our approach is the definition of utility in terms of the
behaviour of the QA model and whether it is able to provide a  correct answer given a retrieved passage. 
We train a small neural model on passage utility judgements elicited from the QA model's responses and use  utility 
predictions to estimate answer uncertainty. Experimental results show that our 
uncertainty estimator is competitive or better than existing strong 
methods while being light-weight.  
Future work could extend this approach to 
long-form generation tasks \citep{stelmakh-etal-2022-asqa,gao-etal-2023-enabling,min-etal-2023-factscore,zhang-etal-2024-luq} 
where evaluating whether answer correctness is more challenging
\citep{zhang-etal-2024-fine} and to multi-modal QA scenarios
\citep{borszukovszki-etal-2025-know}.

\section*{Acknowledgments} 
We thank the action editor and anonymous reviewers for their constructive feedback. We want to thank Ivan Titov and Gustavo Giménez for useful discussions; and Alex Yu for sharing the synthetic multi-hop QA scripts. We are also grateful to Harry Sfyrakis and the UK Research and Innovation via the Innovate UK award (10092998) to Algomo Limited. This work was supported  by  the UK Engineering and Physical Sciences Research Council (grant EP/W002876/1). We also  gratefully acknowledge the support of the Edinburgh International Data Facility (EIDF) and the Data-Driven Innovation Programme at the University of Edinburgh.

\bibliography{ragu_preprint}
\bibliographystyle{tmlr}

\appendix

\section{Ablation of the Passage Utility Predictor Training Objective}
\label{sec:app:objective:abl}

Table~\ref{tab:main:ragqa-ablation:dev} shows AUROC results on answer
uncertainty prediction with Passage Utility estimators trained with
different variants of the objective in Equation~(\ref{eq:combined}).
The first row shows the full objective (see training details in
Section~\ref{sec:app:implem:det}), the second row shows a variant
where the ranking objective uses only entailment utility annotations ($e$),
and in the third row the objective is solely based on accuracy
prediction ($\mathcal{L}_{BCE}$). As can be seen, there is a drop in
performance when the pairwise ranking loss is not used (i.e.,~last
line of Table~\ref{tab:main:ragqa-ablation:dev}); this component of
the objective provides a smoother signal on passage utility, which is
empirically beneficial. However, when the pairwise ranking loss is
only based on entailment, performance drops by several points, 
highlighting the importance of combining both to model QA answering behaviour.

%\textcolor{purple}{
Table~\ref{tab:main:ragqa-ablation:dev:gem9:webq} reports various ablation studies (with \textsc{Gemma2-9B} on the WebQ development set) with different instantiations of the Passage Utility score $\upsilon_\mathcal{M}$ and our training objective.  We present different ways to combine accuracy~($a$) and entailment~($e$) scores to induce utility rankings with the pairwise ranking loss $\mathcal{L}_{rank}, (\cdot)$ (Equation~\ref{eq:rank}). These include  average~$(a+e)/2$ as in Equation~\ref{eq:utility:score}, addition $(a+e)$, addition but inverting the entailment score $((e|1-e)+a)$ when the passage yields an inaccurate answer ($a=0$), or when one of the two is given a zero weight, i.e., $(a)$ or $(e)$ alone. We asses these ranking variants when using the ranking objective alone (rightmost block) as well as when combined with the binary cross-entropy objective $\mathcal{L}_{BCE}$ (Equation~\ref{eq:combined}) (middle block); we also report performance when only the $\mathcal{L}_{BCE}$ objective is used (leftmost block). All model variants are trained with rank-based model selection and $\lambda=0.25$ following training details in Section~\ref{sec:app:implem:det}.
%}

%\textcolor{purple}{
When looking at the pairwise ranking objective alone  (rightmost block), entailment dominates the ranking and the utility scores learnt; $(e)$ pairwise ranking as well as $(a+e)$ variants yield utility scores that have similar discriminative power. The $(a)$ variant exhibits the worst performance, given that many pairs are discarded due to having the same utility. AUROC values improve when the pairwise ranking is combined with the cross-entropy objective (middle block). In this case, in addition to enforcing the pairwise ranking with $\mathcal{L}_{rank}, (\cdot)$, the utility scores are regularised. In other words, the utility scores of a pair of passages with the same accuracy (i.e., only ordered by entailment) will end up closer (differ less) than the utility scores of a pair of passages with different accuracies (differ more). This is reminiscent of the Bradley-Terry model \citep{bradley-terry-model}:  in the first case, one passage is better than the other with a small probability, while in the second case the probability of one being better than the other is higher. The exception here is the $\mathcal{L}_{rank}, (e) + \lambda \, \mathcal{L}_{BCE}$ variant where the pairwise ranking by entailment may often contradict the binary cross-entropy signal.
%}

\begin{table}[t]
    \small
    \centering
    \begin{tabular}{l*{3}{c}}
         \toprule
        \multicolumn{1}{c}{Objective Terms}  & \textsc{G9B} & \textsc{L8B} & \textsc{M7B}\\ 
        \midrule
        $\mathcal{L}_{rank}, (e + a)/2 + \lambda \, \mathcal{L}_{BCE} $ & 0.80 & 0.81 & 0.80 \\
        $\mathcal{L}_{rank}, (e) + \lambda \, \mathcal{L}_{BCE} $  & 0.73 & 0.73 & 0.71 \\
        $\mathcal{L}_{BCE}$  & 0.78 & 0.78 & 0.80\\
        \bottomrule
    \end{tabular}
    %\vspace{-0.2cm}
        \caption{Answer uncertainty estimation with Passage Utility
      predictors trained with different variants of the
      objective in Equation~(\ref{eq:combined}). AUROC values for \textsc{Gemma2-9B (G9B)}, \textsc{Llama3.1-8B (L8B)}, and \textsc{Mistral-7B-v0.3 (M7B)} are averaged over development sets.}
    \label{tab:main:ragqa-ablation:dev}
\end{table}

\begin{table}[t]
    \small
    \centering
    \begin{tabular}{l|cccc|ccccc}
         \toprule
        $\mathcal{L}_{BCE}$ & \multicolumn{4}{c|}{$\mathcal{L}_{rank} + \lambda \,  \mathcal{L}_{BCE}$} & \multicolumn{5}{c}{$\mathcal{L}_{rank}$} \\          
        & $(e + a)/2$ 
        & $(e + a)$ 
        & $((e|1-e) + a)$ 
        & $(e)$ 
        & $(e + a)/2$ 
        & $(e + a)$ 
        & $((e|1-e) + a)$ 
        & $(e)$ 
        & $(a)$  \\
        \midrule
        0.71 & 0.75 & 0.74 & 0.72 & 0.63 & 0.65 & 0.63 & 0.62 & 0.61 & 0.53 \\
        \bottomrule
    \end{tabular}
    %\vspace{-0.2cm}
        \caption{Answer uncertainty estimation with Passage Utility
      predictors trained with different variants of the
      objective in Equation~(\ref{eq:utility:score}). AUROC values are for \textsc{Gemma2-9B} and the WebQ development set.}
    \label{tab:main:ragqa-ablation:dev:gem9:webq}
\end{table}

\section{Test Time Cost of Uncertainty Estimation Methods}
\label{sec:app:app:cost}

Table~\ref{tab:app:cost} shows the cost of estimating uncertainty for
question~$x$, measured by the number of inference calls
required. Simple information theoretic methods (e.g., PPL) require a
single call to the target QA model with the retrieval augmented QA
prompt (i.e.,~$|R|$ retrieved passages and question $x$). However,
approaches that estimate uncertainty based on diversity (e.g., Regular
Entropy, Semantic Entropy, and p(true)) require generating
$N$~answers, i.e., $N$~inference calls with the retrieval augmented QA
prompt. In addition, Semantic Entropy requires the computation of
answer clusters (i.e., grouping answers with the same meaning), so
additional calls to an entailment model are required to compare the
set of sampled answers. p(true) requires one additional LLM call to
elicit a True/False answer but with a very long prompt including
in-context examples and the assessment question with the $|R|$
retrieved passages, sampled and most likely answers, and question~$x$
(see Table~\ref{tab:ptrue:prompt}). In contrast, our approach requires
$|R|$~utility predictions with a BERT-sized model. 
For instance, in our experimental setup with $N=10$ samples and retrieval augmented QA with $|R|=5$, the Semantic Entropy approach would require 11 forward passes with the QA model prompted with 5 passages (one the greedy candidate and 10 random samples) plus 45 calls to the entailment model. Our approach requires 5 forward passes with the BERT-based model.

\begin{table}[t]
    \centering
    \small
    \begin{tabular}{lp{4cm}}
    \toprule
    \multicolumn{1}{c}{Methods}  & \multicolumn{1}{c}{Inference Calls at Test Time} \\ 
    \hline
PPL & $1 \, \text{G}$ \\ 
p(true) & $(N + 1) \, \text{G} + 1 \, \text{E}$ \\
Regular Entropy   & $(N + 1) \, \text{G} $ \\
Semantic Entropy   & $(N + 1) \, \text{G} + {N\choose 2} \, \text{E}$ \\
Passage Utility & $|R| \, \text{BERT-F}$   \\
    \bottomrule
    \end{tabular}
    \caption{Number and type of inference calls required to estimate
      answer uncertainty for question~$x$ and set of retrieved
      passages~$R$. G means inference is performed with a retrieval
      augmented QA model, i.e., an LLM forward pass with the prompt
      including the set of~$|R|$ retrieved passages and question~$x$
      to generate a candidate answer~$y$. E is inference with an
      evaluation model, e.g.,~a forward pass to ask an LLM for
      correctness in p(true) or a forward pass with an entailment
      model in Semantic Entropy. Bert-F is an inference call to
      predict passage utility for passages~$p$ in $R$ and
      question~$x$.}
    \label{tab:app:cost}
\end{table}

\section{Generality of the Uncertainty Aggregation Strategy}
\label{sec:app:unc:aggregation}

To validate the intuition behind our uncertainty estimation step (Section~\ref{sec:ragunc}), we compare the behaviour of the QA model  when prompted with individual passages in $R$ versus when prompted with $R$ (i.e., the top-$|R|$ passages). 
In particular, we want to inspect the proportion of cases where our stratege of taking the maximum uitility score among the passages in $R$ does not agree with the entire set accuracy. In other words, we are interested in cases  where the QA model is accurate with at least one individual passage (\textsc{IndAcc}$\,\uparrow$) but answers incorrectly when promtped with $R$  and vice versa.
In this study,  we consider two individual passage accuracy variants  (related to the combined definition of Passage Utility Section~\ref{sec:utility:ranker}). One is based on accuracy~$a$ (see Section~\ref{sec:experimental}) and is either~0 or~1. The other one is also based on accuracy but smoothed by  entailment~$e$ (computed by an off-the-shelf entailment model; see Section~\ref{sec:experimental}) and downgrades cases where $a=1$ into $a=0$ if $e<0.5$. The latter occurs in cases where the QA model produces an answer that is accurate but not entailed by the passage.

We analyse the behaviour of  four QA models (LLMs) across five datasets in our experimental setup.
Table~\ref{tab:acc:conflicts} shows  the proportion of model disagreements  across development sets. We can see that such disagreements amount to a relatively small number in both settings, i.e., when at least one individual passage in $R$ yields a correct answer (\textsc{IndAcc}$\,\uparrow$) but the QA model prompted with $R$ yields an incorrect answer (\textsc{RAcc}$\,\downarrow$) and vice versa (\textsc{RAcc}$\,\uparrow$ \textsc{IndAcc}$\,\downarrow$). The results in Table~\ref{tab:acc:conflicts} confirm that 
our aggregation approach based on individual passages is fairly general. It approximates answer uncertainty when prompting with~$|R|$ passages while avoiding the  complexity of estimating uncertainty over all possible combinations of input passages in terms of number and order.

\begin{table}[t]
\centering
{\small
  \begin{tabular}{@{}c@{\hspace{0.2cm}}c@{\hspace{0.2cm}}c@{\hspace{6pt}}c@{\hspace{6pt}}
  c@{\hspace{6pt}}c@{\hspace{6pt}}c@{\hspace{6pt}}c@{}}
 
\toprule
  & &  G9B & G27B & L8B  & M7B  \\
\midrule

\multirow{2}{*}{\textsc{RAcc}$\,\downarrow$\textsc{IndAcc}$\uparrow$} & $a$
 &    0.12 & 0.13 & 0.15 & 0.15 \\ %avg_riic_prop
& $a$ + $e$
 &  0.04 & 0.04 & 0.05 & 0.06 \\[1ex] %avg_riic_smooth_prop

\textsc{RAcc}$\,\uparrow$ \textsc{IndAcc}$\,\downarrow$ & $a$ &  0.01 & 0.01 & 0.01 & 0.01 \\ %avg_rcii_prop

\bottomrule
  \end{tabular}  
 }
% \vspace{-0.7em}
 \caption{Average proportion of cases (five development sets) where at least one individual passage in $R$ leads a correct answer (\textsc{IndAcc}$\,\uparrow$) but the QA model prompted with $R$ yields an incorrect one (\textsc{RAcc}$\,\downarrow$) (and vice versa \textsc{RAcc}$\,\uparrow$ \textsc{IndAcc}$\,\downarrow$).}\label{tab:acc:conflicts}
\end{table}

Our study is related to the issue of understanding LLM sensitivity to external evidence \citep{xie2024adaptive,liu-etal-2024-lost}, i.e., how the type of evidence (supportive, contradictory, irrelevant, or misleading), the amount, and order of presentation affect LLM predictions and interact with parametric knowledge. 
The Passage Utility predictor is trained to predict the error of a target QA model (LLM)  on answering questions, independent of the type of passage or any memorized knowledge. Given a question-passage pair, if the LLM relies on its memorized knowledge  rather than adapting to the passage and still produces the correct answer, or conversely, adapts to the passage but produces an incorrect answer, then the Passage Utility predictor should reflect this outcome by predicting a correct or incorrect answer accordingly \cite{xie2024adaptive}.

Note that the Passage Utility predictor is meant to be synchronized with the target QA model and judgments of what is (or not) a correct answer. If an answer to a question changes, the target QA model answer correctness on this question may also change, and the Passage Utility predictor should also reflect this (i.e., it should be adapted with new examples). 
Multiple passage interactions studied in datasets with synthetic evidence \citep{longpre-etal-2021-entity,xie2024adaptive} are observed to a lesser extent in our experiments with six datasets and external retrievers. This has also been recently pointed out in \citet{cub2025}. Our approach could be combined with additional features to capture more complex interactions \citep{dong-etal-2018-confidence}.
Investigating and understanding the relation between QA model uncertainty and (improving) context utilization is an interesting topic on its own right  \citep{xie2024adaptive,longpre-etal-2021-entity,cub2025} but out of scope for  this paper. 

\section{Out-of-distribution Generalization of Uncertainty Estimation}
\label{sec:app:generalisation}

We assess the generalization ability of our Passage Utility estimator both in terms of new QA tasks and QA models. 
We train a unique Passage Utility predictor for the \textsc{Gemma2-9B} model.
Following previous work on question answering and out-of-distribution (o.o.d) scenarios \citep{kamath-etal-2020-selective,zhang-etal-2021-knowing}, we train it on the SQuAD dataset and then use it to predict zero-shot (i.e., without further fine-tuning) passage utilities on all other datasets (test set). As p(true) relies on~20 in context training examples, we also evaluate its ability to generalise in out-of-distribution settings. 

Table~\ref{tab:main:unc:ood:test} ~(QA Task block) shows AUROC for answer uncertainty estimation in o.o.d scenarios. As an upper bound, the i.i.d block of the table shows AUROC values in the in-distribution scenario for Passage Utility and p(true). 
We compare  o.o.d performance w.r.t. PPL and Semantic Entropy which
do not rely on training examples. Although Passage Utility performance decreases in o.o.d settings, it remains competitive in four out of five datasets. In these four cases, it is always statistically significantly different from the PPL method and comparable to p(true) and Semantic Entropy. 
Interestingly, p(true)'s performance also drops in all o.o.d test sets showing that relying on a fixed number of in-context learning examples is neither a robust nor scalable adaptation method.

%\textcolor{purple}{
To understand the observed performance drop, we conducted a comparative analysis of passage utility predictions for 50~question-passage pairs. We examined predictions from a utility predictor trained on WebQ data (i.i.d) versus one trained on SQuAD (o.o.d).
We sampled 10 WebQ test questions: 5 for which the predicted utility decreased from high under i.i.d. conditions to low under o.o.d. conditions (changing from true negative to false positive), and 5 for which it increased in the opposite direction (changing from true positive to false negative).  In both scenarios, the o.o.d predictions were predominantly influenced by token overlap and similarity. We hypothesize that when faced with o.o.d questions (e.g., in terms of type, length, or topic), the prediction mechanism 
defaults to predictions based on a general notion of question-passage similarity, disregarding whether the QA model can answer the question with the given passages. For passages with low (high) similarity to the question, it will predict low (high) utility scores. This behaviour may be a consequence of under-training, suggesting that the Passage Utility predictor cannot predict the  accuracy  of the target QA model on these o.o.d questions and their corresponding retrieved passages.
%}

%\textcolor{purple}{
To empirically validate these observations, we calculate Spearman's $\rho$ correlation coefficients between the predicted Passage Utility scores and the corresponding retriever relevance scores (as computed by Contriever-MSMARCO; \citealt{izacard2022unsupervised}) for each passage, examining correlations separately for both i.i.d and o.o.d predictions. % on the sampled question-passage pairs. 
We use the Retriever Score because it measures the semantic relation between questions and retrieved passages. In agreement with our manual inspection, there is a positive correlation between Passage Utility and Retriever Score in the o.o.d setting (0.536 with $p$-value < 0.01), suggesting that Passage Utility falls back to textual similarity. In contrast, the i.i.d setting exhibits a negative and weaker correlation (-0.318 with $p$-value < 0.05), indicating that training successfully aligns Passage Utility predictions with the target QA model's \emph{actual} performance rather than superficial retriever similarity scores.
%}

%\textcolor{purple}{
We also evaluate generalisation to a new QA model, by training the Passage Utility predictor on utility labels observed for \textsc{Gemma2-9B} and use its predictions to estimate uncertainty for the bigger \textsc{Gemma2-27B} model. We also evaluate p(true) in this o.o.d setting, i.e., we create p(true) prompts as usual with \textsc{Gemma2-27B} generated answers (greedy and sampling), but ask \textsc{Gemma2-9B}  to judge the probability of the most likely answer being true.
%}
%\textcolor{purple}{
Table~\ref{tab:main:unc:ood:test} ~(QA Model block) shows AUROC results; for reference, we include in-distribution values (i.i.d block) and compare the o.o.d results with comparison methods. Passage Utility outperforms all other methods across the board, and while still competitive, p(true) exhibits a higher decrease in performance. These preliminary results suggest that in the context of retrieval augmented QA, models behave alike (also suggested by the distribution of observed correct/incorrect individual passage utilities in Table~\ref{tab:inst:contrast:dist}). 
This highlights practical benefits of our approach, such as training a base Passage Utility predictor using data generated by a less expensive model or developing a more general predictor applicable across multiple QA models.
%}

\begin{table*}[t]
    \small
    \centering
    \begin{tabular}{ccl*{5}{c}}
         \toprule
&& \textsc{Gemma2-9B} & NQ & TQA & WebQ & PopQA & RefuNQ  \\ 
        \cmidrule{2-8}
\parbox[t]{4mm}{\multirow{7}{*}{\rotatebox[origin=c]{90}{\hspace{0.4cm}QA Task}}} & \parbox[t]{2mm}{\multirow{2}{*}{\rotatebox[origin=c]{90}{i.i.d}}} & p(true) & 0.73 & 0.75 & 0.67 & 0.81 & --- \\
&& Passage Utility  & 0.76 & 0.85 & 0.69 & 0.86 & --- \\ 
\cmidrule{2-8}
& \parbox[t]{2mm}{\multirow{2}{*}{\rotatebox[origin=c]{90}{o.o.d\hspace{0.5cm}}}} & PPL & 0.64 & \hspace{0.5em}0.68$^*$ & 0.52 & \hspace{0.5em}0.59$^*$ & 0.51 \\
&& Semantic Entropy & \textbf{0.70} & 0.73 & \hspace{0.5em}0.58$^*$ & \textbf{0.73} & \hspace{0.5em}0.59$^*$ \\
&& p(true) (SQuAD) & 0.67 & 0.63 & \textbf{0.63} & 0.72 & 0.62 \\     
&& Passage Utility (SQuAD) & 0.65 & \textbf{0.79} & 0.60 & 0.72 & \textbf{0.79} \\

         \midrule
&& \textsc{Gemma2-27B} & NQ & TQA & WebQ & PopQA & RefuNQ  \\ 
        \cmidrule{2-8}
\parbox[t]{4mm}{\multirow{7}{*}{\rotatebox[origin=c]{90}{\hspace{0.4cm}QA Model}}} &\parbox[t]{2mm}{\multirow{2}{*}{\rotatebox[origin=c]{90}{i.i.d}}} &         p(true) & 0.77 & 0.83 & 0.67 & 0.79 & 0.60 \\
&&         Passage Utility & 0.73 & 0.82 & 0.69 & 0.87 & 0.80 \\
        \cmidrule{2-8}
& \parbox[t]{2mm}{\multirow{2}{*}{\rotatebox[origin=c]{90}{o.o.d\hspace{0.5cm}}}} &         PPL & \hspace{0.5em}0.64$^*$  & 0.50 & 0.53 & 0.53 & 0.51 \\
&&         Semantic Entropy & 0.68 & 0.62 & 0.59 & 0.66 & 0.58 \\
&&         p(true) (\textsc{Gemma2-9B}) & 0.73 & \hspace{0.5em}0.71$^*$ & \hspace{0.5em}0.64$^*$ & \hspace{0.5em}0.75$^*$ & \hspace{0.5em}0.59$^*$ \\
&&         Passage Utility (\textsc{Gemma2-9B}) & \hspace{0.5em}\textbf{0.75} & \textbf{0.80} & \textbf{0.68} & \textbf{0.87} & \textbf{0.79} \\
\bottomrule
    \end{tabular}
    \caption{Out-of-domain performance of Passage Utility predictor
      for \textsc{Gemma2-9B} both in terms of QA task and QA model. 
      i.i.d blocks report AUROC values from our main in-distribution experiments (Table~\ref{tab:main:unc:test}); o.o.d blocks contain the o.o.d comparison.
      Best values are highlighted in
      \textbf{bold}; we also mark with~$ ^*$ next best values which are significantly different using the paired De Long test ($p < 0.05$).
      For the QA task, Passage Utility and p(true) are supervised with SQuAD data and evaluated on NQ, TQA, WebQ, PopQA, and RefuNQ test data. For the QA model, Passage Utility predictors are trained on \textsc{Gemma2-9B} and used to estimate uncertainty for \textsc{Gemma2-27B}; p(true) assessment is provided by \textsc{Gemma2-9B}. }
    \label{tab:main:unc:ood:test}
\end{table*}

%\textcolor{purple}{
Finally, we train a unique Passage Utility predictor for \emph{all} QA tasks and assess its generalisation capabilities. To this end, we train a predictor for the \textsc{Gemma2-9B} model on a random sample drawn from the five training sets of size equivalent to the training sets of the individual QA task predictors. 
Specifically, we took 10k from NQ, TQA, and SQuAD, 3k from PopQA, all WebQ, totalling 35,474 instances for training and 500 instances from each for validation; the number of pairwise instances in the final curated training set is 354,379.
For p(true), we mixed 4 randomly sampled examples from each dataset as in-context training examples. 
We follow the same training procedure as described in Section~\ref{sec:app:implem:det} (with combined model selection and $\lambda=1$). We show AUROC results in Table~\ref{tab:main:unc:unique:test}. Across the six test sets, the unique Passage Utility predictor trained on a mix of QA tasks (bottom block) achieves similar performance to the individual predictors trained on per-task datasets (top block). The unique predictor keeps comparable or better performance than p(true) and outperforms PPL and Semantic Entropy. This preliminary study suggests that is feasible to train a more general predictor for various QA tasks.
%}

\begin{table*}[t]
    \small
    \centering
    \begin{tabular}{cl*{6}{c}}
         \toprule
        & & NQ & TQA & WebQ & SQuAD & PopQA & RefuNQ  \\ 
        \midrule
\parbox[t]{2mm}{\multirow{2}{*}{\rotatebox[origin=c]{90}{Task}}} & p(true) & 0.73 & 0.75 & 0.67 & 0.68 & 0.81 & 0.62 \\
& Passage Utility & 0.76 & 0.85 & 0.69 & 0.80 & 0.86 & 0.79 \\
\midrule
\parbox[t]{2mm}{\multirow{4}{*}{\rotatebox[origin=c]{90}{Unique}}} & 
  PPL & \hspace{0.5em}0.64$^*$ & 0.68 & 0.52 & 0.59 & 0.59 & 0.51 \\
& Semantic Entropy & 0.70 & \hspace{0.5em}0.73$^*$ & \hspace{0.5em}0.58$^*$ & \hspace{0.5em}0.63$^*$ & 0.73 & 0.59 \\
& p(true)         & \textbf{0.71} & 0.70 & \textbf{0.68} & \hspace{0.5em}0.63$^*$ & \hspace{0.5em}0.75$^*$  & \hspace{0.5em}0.66$^*$ \\
& Passage Utility  & 0.69 & \textbf{0.83} & \textbf{0.68} & \textbf{0.79} & \textbf{0.85} & \textbf{0.79} \\
\bottomrule
    \end{tabular}
    \caption{Generalisation of a Unique Passage Utility predictor
      for \textsc{Gemma2-9B} trained on a mix of QA tasks and evaluated on the six test sets (lower block). Best values are highlighted in \textbf{bold}; we also mark with~$ ^*$ next best values which are significantly different using the paired De Long test ($p < 0.05$).
      The upper block reports AUROC values for the Passage Utility from predictors trained for \textsc{Gemma2-9B} on indiviual QA tasks' training datasets. We also compare p(true) when in-context learning examples are all from the same QA task (upper block) versus from a mix of tasks (lower block).}
    \label{tab:main:unc:unique:test}
\end{table*}

\section{Synthetic Qualitative Analysis of Passage Utility in Multi-hop QA}
\label{sec:app:mhqa:synth:analysis}

%\textcolor{purple}{
In multi-hop QA Passage Utility may fail at cases where the QA model cannot answer when prompted with any individual 'hop' passage, but answers correctly when prompted with the full set.
To analyse this limitation, we carry out a qualitative analysis. 
As discussed in Section~\ref{sec:results}, there are two scenarios in retrieval augmented multi-hop QA (given existing datasets and current LLMs). Either models can answer with one passage (thus no multi-hop) or retrieval completely fails. Thus, to be able to pinpoint actual multi-hop QA cases and expose the limitations of our approach, we make use of synthetic HotPotQA data \citep{yang-etal-2018-hotpotqa}. 
%}

%\textcolor{purple}{
First, we use the sets of gold passages provided in the HotPotQA dataset to categorise question types as follows. We prompt the QA model with individual gold passages and with combinations thereof, measuring accuracy. Out of the 500 test questions, 475 are answerable with only one gold 'hop' passage, 13 require multiple passages, and 12 are unanswerable. Secondly,  we focus on the 13 requiring multi-hop questions and insert gold 'hop' passages within the retrieved set for those where retrieval completely fails. 
Then, we run QA and uncertainty estimation.
In 6 of 13 cases, our approach fails to predict that the QA model’s answer will be correct. It is important to note, however, that these 13 cases represent only 2.6\% of the total (13 out of 500).
%}

\section{Experimental Details}

\subsection{Datasets and Splits}
\label{sec:app:datasets}

In our experiments, we use six QA tasks which we describe below. Table~\ref{tab:datasets:stat} shows dataset statistics and example question-answers pairs. 

\paragraph{Natural Questions} (NQ;
\citealt{kwiatkowski-etal-2019-natural}) is a QA dataset compiled from
real user questions submitted to the Google search engine. As
part of the dataset curation process, annotators judge the quality of
questions and associate them with a short answer that can be extracted
from a related Wikipedia page.  
%These questions are impresice and ambiguous, there is answer variability, experiments in the original paper on this.

\paragraph{TriviaQA} (TQA; \citealt{joshi-etal-2017-triviaqa})  is a  question answering dataset designed for training and evaluating machine learning models on open-domain question answering tasks. The dataset was created by gathering questions from trivia websites, along with their corresponding answers, to provide a broad range of factual questions.
% questions are supposed to be more complex, requiering complex reasoning over several sentences from the evidence and lexical variation of the answer w.r.t the evidence paragraph.

\paragraph{WebQuestions} (WebQ; \citealt{berant-etal-2013-semantic})
was mined off questions generated with the Google Suggest API. The
answers to the questions are defined as Freebase entities (i.e., their
string label) and were elicited by Amazon Mechanical Turk (AMT)
annotators.
%in some cases answers are roughly accurate as AMT workers are obliged to select entities only from Freebase. Also list of entities might be incomplete.

\paragraph{SQuAD} \citep{rajpurkar-etal-2016-squad} contains questions
formulated by AMT annotators based on a given Wikipedia paragraph,
with the answer being a short span in that paragraph. Annotators were
encouraged to use paraphrasing when writing the question. The answer
types not only cover named entities but also other categories such as
noun- and verb-phrases.
% variety of answer types (noun- and verb- phrases). questions and paragraphs' lexical & syntactic divergence, multi-sentence reasoning, ambiguos questions (small percentage).

\paragraph{PopQA} \cite{mallen-etal-2023-popQA} is an open-domain QA
dataset, focusing on popular culture topics, such as movies, TV shows,
music, and sports. It  contains question-answer pairs derived from (subject,
relation, object) triples in Wikidata . Triples were translated into
natural language and the object entity was taken to be the gold
answer. The collection process focused on gathering questions about
subject entities of varying popularity. 

\paragraph{RefuNQ} \cite{refunq} is derived from NQ and consists of answerable and unanswerable questions. Unanswerable questions are created by replacing entities in the original NQ question by non-existing concepts.
\begin{table*}[t]
    \centering
    \small
    \begin{tabular}{@{}lrrrp{5cm}p{3.5cm}@{}}
    \toprule
    Dataset & \multicolumn{1}{c}{Train} & \multicolumn{1}{c}{Dev} & \multicolumn{1}{c}{Test} & \multicolumn{1}{c}{Example Question} & \multicolumn{1}{c}{Example Answer} \\
    \hline
NQ &  79,168 & 8,757 & 3,610 & Who plays Letty in Bring it on all or nothing? & Francia Raisa \\
TQA &78,785 & 8,837 & 11,313 & Who was the first artistic director of the National Theatre in London? & Lord Laurence Olivier \\
WebQ & 2,474 & 361 & 2,032 & What party was Andrew Jackson? & Democratic-Republican Party \\
SQuAD & 78,713 & 8,886 & 10,570 & What is the Grotto at Notre Dame? & A Marian place of prayer and reflection \\
PopQA & 10,000 & 1,267 & 3,000 & Who was the director of Champion? & Rabi Kinagi \\
RefuNQ & \multicolumn{1}{c}{---} & \multicolumn{1}{c}{---} & 2,173 & Who does the voice over in the Requirtion? & \multicolumn{1}{c}{---} \\
\bottomrule
    \end{tabular}
    \caption{Dataset statistics, number of instances per
      Train/Development(Dev)/Test sets, and example question-answer
      pairs (all taken from the Dev set except for RefuNQ).} 
    \label{tab:datasets:stat}
\end{table*}

We follow previous work \citep{lee-etal-2019-latent} and use only the
question and gold answers, i.e., the open versions of NQ, TQA, and
SQuAD. We use the unfiltered TQA dataset. We follow the train/dev/test
splits as used in previous work \cite{lee-etal-2019-latent} and randomly split PopQA. RefuNQ only provides a test set so our experiments on this dataset 
are zero-shot from a Passage Utility predictor trained on SQuAD.
We follow \citet{farquhar2024semantic-hallu} and use 400 test examples 
randomly sampled from the original larger test datasets for evaluation of uncertainty quantification.

\subsection{Implementation Details}
\label{sec:app:implem:det}

\paragraph{QA Models}

For all question answering tasks, we use the off-the-shelf
Contriever-MSMARCO  tool \citep{izacard2022unsupervised}  to retrieve sets of
passages~$R$ for question~$x$ from Wikipedia and the official
Wikipedia embeddings based (2018 snapshot) as our document
knowledge-base. For PopQA, we follow the work of
\citet{mallen-etal-2023-popQA} who also use the full 2018 English
Wikipedia dump.

The QA prompt used for all models (embedded in the corresponding chat templates) is shown in Table~\ref{tab:ragu_prompts}.
For inference, we set the maximum number of generated tokens to 50 for both the greedy (most likely answer) as well as temperature scaled (sampled candidates) decoding. We use vLLM for inference \citep{vllm}. For all models, inference was run on a single A100-80GB GPU.

%\textcolor{purple}{
\paragraph{Curated Passage Utility Dataset}
We train our passage utility predictor on a dataset~$D_{\mathcal{M}}$ curated from benchmark $D$, e.g., WebQ, consisting of question and gold answer pairs $(x, y)$. For each question we retrieve the top-$k$ passages. Then, we pair question~$x$ and retrieved passages~$p$ with utility
scores~$\upsilon_{\mathcal{M}}$ which we obtain after
running the QA model ~$\mathcal{M}$ on inputs ${(x, p)}$ and computing
the generated answer accuracy and entailment scores (Section~\ref{sec:utility:ranker}), i.e., we create tuples ${(x, p, \upsilon_{\mathcal{M}})}$. From the set of $k$ tuples for question~$x$, we derive ${k\choose 2}$ instances for our pairwise ranking loss.
%}

%\textcolor{purple}{
In experiments, we use $k=5$ retrieved passages per question.
Table~\ref{tab:curated:data} reports the size (number of training instances) of the curated datasets for each QA task and model. From each question and set of top-$5$ retrieved passages, we derive 10 pairwise ranking instances, discarding those that have equal utilities (e.g., from the WebQ training split with 2,474 question-answer pairs, we curate 24,720 instances with \textsc{Mistral-7B-v0.3}).
As our top-$5$ passages are obtained via a real retrieval module, i.e., not synthetically assembled, there are questions for which all passages in the top-$5$ set lead to a correct (incorrect) answer. In these cases, the pairwise ranking is dominated by the entailment score (i.e., accuracy is the same). Table~\ref{tab:inst:contrast:dist} shows the distribution of questions with all retrieved passages leading to the same accuracy (Correct/Incorrect) or mixed (Mixed) accuracies in the curated dataset for each QA task and model. 
%}

\begin{table}[t]
    \centering
    \small
    \begin{tabular}{lcllll}
         \toprule
       \multicolumn{1}{c}{Models} & \multicolumn{1}{c}{NQ} & \multicolumn{1}{c}{TQA} & \multicolumn{1}{c}{WebQ} & \multicolumn{1}{c}{SQuAD} & \multicolumn{1}{c}{PopQA}  \\ 
        \hline \\[-1.2ex]
        
    \textsc{Gemma2-9B} & 395,438 & 393,475 & 24,721 & 393,285 & 99,770 \\     
    
    \textsc{Gemma2-27B} & 395,426 & 393,477 & 24,723 & 393,293 & 99,778 \\   
        
    \textsc{Llama-3.1-8B} & 790,862 & 393,465 & 24,713 & 393,288 & 99,787 \\
        
    \textsc{Mistral-7B-v0.3} & 395,397 & 393,474 & 24,720 & 393,271 & 99,772 \\[0.4ex]

\bottomrule
    \end{tabular}
    \caption{Number of pairwise training instances in the curated datasets to train the Passage Utility predictor with the combined pairwise ranking and binary cross-entropy losses (Section~\ref{sec:results}). }
    \label{tab:curated:data}
\end{table}

\begin{table}[t]
    \centering
    \small
    \begin{tabular}{lccc}
         \toprule
        & \multicolumn{1}{c}{Incorrect} & \multicolumn{1}{c}{Mixed} & \multicolumn{1}{c}{Correct} \\ 
        \hline \\[-1.2ex]
        
    \textsc{Gemma2-9B} &  21\% & 52\% & 27\% \\      
        
    \textsc{Gemma2-27B} & 20\% & 53\% & 27\% \\   
        
    \textsc{Llama-3.1-8B} &  22\% & 55\% & 23\% \\
        
    \textsc{Mistral-7B-v0.3} & 21\% & 56\% & 23\% \\[0.4ex]

\bottomrule
    \end{tabular}
    \caption{Number of training instances in curated datasets to train the passage utility predictor with the combined pairwise ranking and binary cross-entropy losses (Section~\ref{sec:results}). }
    \label{tab:inst:contrast:dist}
\end{table}

\paragraph{Passage Utility Predictor Training Details}

We train a different predictor for each target QA model and QA
task. Given the large number of predictors required in  our
experiments, we initially tested the hyper-parameters used in
\citet{fang-etal-2024-efficiently} on the NQ dataset and choose a set
thereof for all predictor instances. We train each predictor for 3
epochs, with a batch size of 32 examples, learning rate equal to
2$^{e-5}$, and weight decay 0.001 (with the exception of  
\textsc{Llama-3.1-8B} and WebQ where we used 0.01).
For each predictor we performed search on values for $\lambda$, i.e., the
contribution of the $\mathcal{L}_{BCE}$ loss
(Equation~\ref{eq:combined}), and different criteria for model
selection, i.e., the best at pairwise ranking or at both pairwise
ranking and accuracy prediction
(combined).

Table~\ref{tab:model:select} shows the configuration for each
predictor. Table cells show selection criteria (R for ranking and C
for combined) and the value for~$\lambda$. 
At inference time we predict a single Passage Utility score given by the selected best checkpoint.
For all predictor instances (except for all WebQ and PopQA predictors and
the predictor for \textsc{Llama-3.1-8B} and NQ),  we use half of the
available training data to speed up experiments. Training and inference was run on a single A100-40GB GPU; training ranges from 2 to 12 hours depending on the dataset.

\begin{table}[t]
    \centering
    \small
    \begin{tabular}{lclllll}
         \toprule
       \multicolumn{1}{c}{Models} & \multicolumn{1}{c}{NQ} & \multicolumn{1}{c}{TQA} & \multicolumn{1}{c}{WebQ} & \multicolumn{1}{c}{SQuAD} & \multicolumn{1}{c}{PopQA} & \multicolumn{1}{c}{HotPotQA} \\ 
        \hline \\[-1.2ex]
        
        \textsc{Gemma2-9B} & C, 0.25 & C, 0.25 & R, 0.25 & C, 1 & C, 1 & C, 0.25 \\      
        
        \textsc{Gemma2-27B} & C, 0.25 & C, 1 & C, 1 & C, 0.25 & R, 0.25 & C, 1 \\   
        
        \textsc{Llama-3.1-8B} & C, 0.25 & C, 0.25 & C, 0.25 & C, 1 & C, 1 & C, 0.25 \\
        
        {\textsc{Mistral-7B-v0.3}} & R, 0.25 & C, 1 & C, 0.25 & C, 0.25 & C, 0.25 & C, 1 \\[0.4ex]

\bottomrule
    \end{tabular}
    \caption{This table shows the $\lambda$ value and selection criteria (R for pairwise ranking or C for combined pairwise ranking and accuracy prediction) for each Passage Utility predictor in our experiments.}
    \label{tab:model:select}
\end{table}

\paragraph{Comparison Approaches}

In this section, we describe
additional answer uncertainty estimation methods (for which we present supplementary results in Section~\ref{sec:app:additionalres}).  Maximum Sequence
Probability (MSP) is based on the probability of the most likely
answer and is computed as:
\begin{equation}
\text{MSP}(x, R, \mathcal{M}) = 1 - P(y | x, R; \mathcal{M}).    
\end{equation}

Note that, in contrast to $\text{PPL}(x, R, \mathcal{M})$ reported in the
main section of the paper, this metric is biased by answer length,
i.e., identifying an answer to have low probability (low confidence)
because of its length. Despite the fact that QA models are instructed
to produce short answers, they do not always follow instructions. For
this reason, we consider perplexity a more accurate metric. Indeed,
answer length could indicate that the model
is uncertain about the answer. Thus, we also estimate answer uncertainty 
from the Average Answer Length
(AvgAnsLen) as the average number of words in the sampled
answers. 
As seen in Section~\ref{sec:app:additionalres:unc}, Table~\ref{tab:main:unc:dev}, MSP and AvgAnsLen  perform similarly across the board.

We also report Cluster Assignment (CA) which is a variant of
SE without answer probabilities where the probability of each
generated meaning (i.e., a cluster) is approximated from the number of
answers in the cluster. We found that in general CA estimations are
very close to Semantic Entropy ones.

Another uncertainty estimation approach is the negative mean
Point-wise Mutual Information (PMI;
\citealt{takayama-arase-2019-relevant}) over tokens; i.e.,~it compares
the probability of answer sequence $y$ given a prompt with question~$x$ and passages~$R$ w.r.t the probability given by~$\mathcal{M}$ to
$y$~without any context. Intuitively, the higher the point-wise mutual
information, the more certain the QA model is on generating~$y$ (i.e.,
the answer is related to or depends on~$x$ and~$R$). PMI is computed
as:
\begin{equation}
\begin{aligned}
\text{PMI}(x, R, \mathcal{M}) =    - \, \frac{1}{|y|} \sum_{t=1}^{|y|} \text{log} \dfrac{p(y_t|y_{1..t-1}, x, R; \mathcal{M})}{p(y_t|y_{1..t-1};\mathcal{M})} .   
\end{aligned}
\end{equation}

We also report Retriever Score as a baseline for Passage Utility. Instead of
using the predicted Passage Utility we use the original relevance score 
assigned by the external retriever (i.e., Contriever MS-MARCO).

We use the implementation provided in \citet{farquhar2024semantic-hallu} to compute Regular Entropy, Semantic Entropy, Cluster Assignment, and p(true). Note that we do not include the supervised baseline reported in \citet{farquhar2024semantic-hallu} as the authors show that it underperforms simple information-theoretic metrics and in addition only works for white-box models. Note that while AvgAnsLen and Retriever Score do not strictly provide
scores in the $[0, 1]$ interval, the package that computes AUROC finds 
ranking decision thresholds.\footnote{\url{https://scikit-learn.org/stable/modules/generated/sklearn.metrics.det_curve.html}}

\paragraph{Metrics}
We use the implementation provided in \citet{farquhar2024semantic-hallu} to compute  AUROC, Accuracy at X\% of rejection, and AURAC metrics.

We use Qwen2-72B-Instruct \citep{qwen2} to obtain accuracy judgments
(i.e., $A$ judge, Section~\ref{sec:experimental});
specifically, we use the Activation-aware Weight Quantization
\citep{awq}, version Qwen2-72B-Instruct-AWQ.  We prompt the accuracy
evaluator with the prompt proposed in \citet{sun-etal-2024-head}, as
we found it to perform well on our datasets. The accuracy evaluation (AccLM) prompt is
shown in Table~\ref{tab:acclm}.  In a sample of 840 generated answers
human and LLM-based judgment of correctness agreed 98\% of the time
\citep{sun-etal-2024-head}.

\begin{table}[t]
    \centering
    \footnotesize
\begin{promptbox}[colback=gray!20!white, left=1mm, right=1mm]{Retrieval augmented QA prompt}
Knowledge: \\ \
[1] \textcolor{blue}{passage} \\ \
[2] \textcolor{blue}{passage} \\ \
... \\ \
[$|R|$] \textcolor{blue}{passage} \\ \
 \\ \
Answer the following question with a very short phrase. \\ \
 \\ \
Question: \textcolor{blue}{question} \\
\end{promptbox}
    \caption{Prompt designed as user turn for QA models.}
    \label{tab:ragu_prompts}
\end{table}

\begin{table}[t]
    \centering
    \footnotesize
\begin{promptbox}[colback=gray!20!white, left=1mm, right=1mm]{p(true) prompt}
Question: \textcolor{blue}{question} \\
Brainstormed Answers: \textcolor{blue}{most likely answer} \\
\textcolor{blue}{sampled answer 1} \\
... \\
\textcolor{blue}{sampled answer N} \\
Possible answer: \textcolor{blue}{most likely answer} \\
Is the possible answer: \\
A) True \\
B) False \\
The possible answer is: \textcolor{blue}{correct choice} \\
\\ \
... \\ \
\\ \
Knowledge: \\ \
[1] \textcolor{blue}{passage} \\ \
[2] \textcolor{blue}{passage} \\ \
... \\ \
[$|R|$] \textcolor{blue}{passage} \\ \
\\ 
Question: \textcolor{blue}{question} \\
Brainstormed Answers: \textcolor{blue}{most likely answer} \\
\textcolor{blue}{sampled answer 1} \\
... \\
\textcolor{blue}{sampled answer $N$} \\
Possible answer: \textcolor{blue}{most likely answer} \\
Is the possible answer: \\
A) True \\
B) False \\
The possible answer is: \\
\end{promptbox}
    \caption{Prompt used for  p(true) approach. Items
       in blue are filled  with in-context examples from the
      training set and the current example being evaluated. $N$~represents
      the number of sampled answers. The ``sequence of in-context
      examples'' prefix is a sequence of examples taken from the
      training split with the same question format but with the answer
      to \emph{The possible answer is:} resolved.}
    \label{tab:ptrue:prompt}
\end{table}

\begin{table}[ht]
    \centering
    \footnotesize
\begin{promptbox}[colback=gray!20!white, left=1mm, right=1mm]{Accuracy evaluation  prompt.}
You need to check whether the prediction of a question-answering system to a question is correct. You should make the judgment based on a list of ground truth answers provided to you. Your response should be "correct" if the prediction is correct or "incorrect" if the prediction is wrong.\\
\ \\
Question: Who authored The Taming of the Shrew (published in 2002)?\\
Ground truth: ["William Shakespeare", "Roma Gill"]\\
Prediction: W Shakespeare\\
Correctness: correct\\
\ \\
Question: Who authored The Taming of the Shrew (published in 2002)?\\
Ground truth: ["William Shakespeare", "Roma Gill"]\\
Prediction: Roma Gill and W Shakespeare\\
Correctness: correct\\
\ \\
Question: Who authored The Taming of the Shrew (published in 2002)?\\
Ground truth: ["William Shakespeare", "Roma Gill"]"\\
Prediction: Roma Shakespeare\\
Correctness: incorrect\\
\ \\
Question: What country is Maharashtra Metro Rail Corporation Limited located in?\\
Ground truth: ["India"]\\
Prediction: Maharashtra\\
Correctness: incorrect\\
\ \\
Question: What's the job of Song Kang-ho in Parasite (2019)?\\
Ground truth: ["actor"]\\
Prediction: He plays the role of Kim Ki-taek, the patriarch of the Kim family.\\
Correctness: correct\\
\ \\
Question: Which era did Michael Oakeshott belong to?\\
Ground truth: ["20th-century philosophy"]\\
Prediction: 20th century."\\
Correctness: correct\\
\ \\
Question: Edward Tise (known for Full Metal Jacket (1987)) is in what department?\\
Ground truth: ["sound department"]\\
Prediction: 2nd Infantry Division, United States Army\\
Correctness: incorrect\\
\ \\
Question: What wine region is Finger Lakes AVA a part of?\\
Ground truth: ["New York wine"]\\
Prediction: Finger Lakes AVA\\
Correctness: incorrect\\
\ \\
Question: \textcolor{blue}{question}\\
Ground truth: \textcolor{blue}{gold answer}\\
Prediction: \textcolor{blue}{generated answer}\\
Correctness:\\
\end{promptbox}
    \caption{Prompt used for LLM-based accuracy evaluation.}
    \label{tab:acclm}
\end{table}

\subsection{Prompts}
\label{sec:app:prompts}

The prompt we use for our QA models is shown in Table~\ref{tab:ragu_prompts}. 
Table~\ref{tab:ptrue:prompt} illustrates the prompt used for the p(true) baseline. Table~\ref{tab:acclm} shows the prompt used for the LLM-based accuracy metric.

\begin{table}[ht]
\centering
{\footnotesize
  \begin{tabular}{@{}l@{\hspace{3pt}}|c@{\hspace{2pt}}c@{\hspace{2pt}}c@{\hspace{3pt}}c@{\hspace{3pt}}c@{\hspace{3pt}}c@{\hspace{3pt}}c@{\hspace{3pt}}|c@{\hspace{2pt}}c@{\hspace{2pt}}c@{\hspace{3pt}}
  c@{\hspace{3pt}}c@{\hspace{3pt}}c@{\hspace{3pt}}c@{\hspace{3pt}}c@{\hspace{3pt}}c@{}} 
    \toprule %\ \\[-1ex]
    
\multicolumn{1}{l}{} & \multicolumn{7}{c@{}}{\textsc{Gemma2-9B}} 
  & \multicolumn{7}{c@{}}{\textsc{Gemma2-27B}} \\
         & {\scriptsize NQ} & {\scriptsize TQA} & {\scriptsize WebQ}  
         & {\scriptsize SQuAD} & {\scriptsize PopQA}  & {\scriptsize RefuNQ}
         & {\scriptsize AVG}
         & {\scriptsize NQ} & {\scriptsize TQA} & {\scriptsize WebQ}  
         & {\scriptsize SQuAD} & {\scriptsize PopQA} & {\scriptsize RefuNQ} 
         & {\scriptsize AVG} \\

\midrule %\ \\[-1.9ex]
PPL & 0.64 & 0.62 & 0.53 & 0.56 & 0.45 & 0.50 & 0.55 %G9B
    & 0.61 & 0.61 & 0.53 & 0.57 & 0.52 & 0.54 & 0.56 \\ %G27B
Passage Utility & 0.77 & 0.80 & 0.71 & 0.74 & 0.90 & 0.71 & 0.77  %G9B
               & 0.74 & 0.8 & 0.69 & 0.76 & 0.90 & 0.72 & 0.77 \\%G27B

\bottomrule %\ \\[-1.9ex]
\multicolumn{13}{c}{} \\\toprule
\multicolumn{1}{l}{} &\multicolumn{7}{c@{}}{\textsc{Llama-3.1-8B}} 
 &\multicolumn{7}{c@{}}{\textsc{Mistral-7B-v0.3}} \\
& {\scriptsize NQ} & {\scriptsize TQA} & {\scriptsize WebQ}  
         & {\scriptsize SQuAD} & {\scriptsize PopQA}  & {\scriptsize RefuNQ}
         & {\scriptsize AVG}
         & {\scriptsize NQ} & {\scriptsize TQA} & {\scriptsize WebQ}  
         & {\scriptsize SQuAD} & {\scriptsize PopQA} & {\scriptsize RefuNQ} 
         & {\scriptsize AVG} \\
\midrule
PPL & 0.69 & 0.78 & 0.67 & 0.67 & 0.65 & 0.68 & 0.69  %L8B
   &  0.64 & 0.71 & 0.59 & 0.64 & 0.62 & 0.68 & 0.65 \\ %M7B
Passage Utility & 0.74 & 0.80 & 0.71 & 0.78 & 0.91 & 0.75 & 0.78 %L8B
               & 0.75 & 0.77 & 0.70 & 0.78 & 0.88 & 0.78 & 0.78 \\ %M7B

\bottomrule
\end{tabular}
}
\captionof{table}{AUROC values for the Passage Utility and perplexity baseline on individual passages on the six test sets (NQ, TQA, WebQ, SQuAD, PopQA, and RefuNQ).}
\label{tab:additionalres:pu:indiv}
\end{table}

\section{Additional Results}
\label{sec:app:additionalres}

\subsection{Reference Retrieval Augmented QA Accuracy}
\label{sec:app:additionalres:refacc}

Table~\ref{tab:main:ragqa-perfor} shows retrieval augmented QA performance for the five QA models on the development and test sets of our six  tasks. 
We report accuracy based on token overlap (Acc) as computed in previous work, i.e., whether the gold answer is contained in the generated answer \citep{mallen-etal-2023-popQA,asai2024selfrag,xie2024adaptive} and accuracy using an LLM as a judge (AccLM). 
Note that AccLM is much higher than Acc across the board, which highlights the importance of using a better accuracy metric,  especially when the target QA models are not fine-tuned.

\subsection{Results on Individual Passage Utility Prediction}
\label{sec:app:additionalres:pu:indiv}

%\textcolor{purple}{
Beyond using  Passage Utility to estimate uncertainty in retrieval augmented QA, we evaluate how it performs on its own. Table~\ref{tab:additionalres:pu:indiv} shows AUROC scores when using Passage Utility to predict accuracy for individual passages. 
We evaluate on the same samples from the test sets in Section~\ref{sec:results} and Table~\ref{tab:main:unc:test} but per passage.  We also include a perplexity baseline (PPL). Overall, results in Table~\ref{tab:additionalres:pu:indiv} follow a similar pattern as those in the QA setting with top-$5$ passages (Table~\ref{tab:main:unc:test}). Across the board, Passage Utility demonstrates strong performance in predicting the usefulness of individual passages.  Moreover, these results highlight that the quality of uncertainty estimation strongly depends on the quality of individual Passage Utility predictions.
%}

\subsection{Detailed Uncertainty Estimation Results}
\label{sec:app:additionalres:unc}

Table~\ref{tab:main:unc:test:AURAC} shows AURAC scores on the test sets.
Table~\ref{tab:main:unc:dev} shows the performance of uncertainty quantification approaches on the development set. We report AUROC and AURAC.

\begin{table*}[t]
    \small
    \centering
    \begin{tabular}{@{}lcccccccccccc@{}}
    
         \toprule
       & \multicolumn{2}{c}{NQ} & \multicolumn{2}{c}{TQA} & \multicolumn{2}{c}{WebQ}  &  \multicolumn{2}{c}{SQuAD} &  \multicolumn{2}{c}{PopQA} &  \multicolumn{2}{c}{RefuNQ} \\ 
      Development   & {\scriptsize Acc} & {\scriptsize AccLM} & {\scriptsize Acc} & {\scriptsize AccLM} & {\scriptsize Acc} & {\scriptsize AccLM} & {\scriptsize Acc} & {\scriptsize AccLM} & {\scriptsize Acc} & {\scriptsize AccLM} & {\scriptsize Acc} & {\scriptsize AccLM}\\ 
        \hline \\[-1.2ex]
%        & \multicolumn{12}{c}{Development} \\[0.4ex]
%        \hline
        \textsc{Gemma2-9B} & 0.48 & 0.66 & 0.74 & 0.80 & 0.46 & 0.66 & 0.38 & 0.60 & 0.51 & 0.52 & --- & --- \\      

        \textsc{Gemma2-27B} & 0.48 & 0.66 & 0.75 & 0.81 & 0.49 & 0.67 & 0.38 & 0.60 & 0.52 & 0.52 & --- & --- \\    
        
        \textsc{Llama-3.1-8B} & 0.48 & 0.62 & 0.71 & 0.77 & 0.53 & 0.64 & 0.39 & 0.57 & 0.51 & 0.49 & --- & --- \\
        
        \textsc{Mistral-7B-v0.3} & 0.48 & 0.62 & 0.72 & 0.76 & 0.52 & 0.69 & 0.37 & 0.58 & 0.53 & 0.51 & --- & --- \\[0.4ex]

        \bottomrule
        \multicolumn{12}{c}{} \\ \toprule
      & \multicolumn{2}{c}{NQ} & \multicolumn{2}{c}{TQA} & \multicolumn{2}{c}{WebQ}  &  \multicolumn{2}{c}{SQuAD} &  \multicolumn{2}{c}{PopQA} &  \multicolumn{2}{c}{RefuNQ} \\ 
      Test   & {\scriptsize Acc} & {\scriptsize AccLM} & {\scriptsize Acc} & {\scriptsize AccLM} & {\scriptsize Acc} & {\scriptsize AccLM} & {\scriptsize Acc} & {\scriptsize AccLM} & {\scriptsize Acc} & {\scriptsize AccLM} & {\scriptsize Acc} & {\scriptsize AccLM}\\ 
        \hline \\[-1.2ex]
        \textsc{Gemma2-9B} & 0.49 & 0.65 & 0.74 & 0.80 & 0.40 & 0.66 & 0.43 & 0.60 & 0.50 & 0.52 & 0.26 & 0.40 \\

        \textsc{Gemma2-27B} & 0.48 & 0.65 & 0.76 & 0.81 & 0.41 & 0.66 & 0.42 & 0.61 & 0.51 & 0.53 & 0.26 & 0.39 \\

        \textsc{Llama-3.1-8B} & 0.49 & 0.61 & 0.71 & 0.77 & 0.44 & 0.63 & 0.43 & 0.58 & 0.50 & 0.49 & 0.27 & 0.36 \\

        \textsc{Mistral-7B-v0.3} & 0.49 & 0.62 & 0.72 & 0.77 & 0.47 & 0.66 & 0.41 & 0.58 & 0.51 & 0.50 & 0.26 & 0.35 \\
        
        \bottomrule
\end{tabular}
    \caption{QA model performance (with $|R|=5$) on the development and test sets. We report token- and model-based accuracy (Acc and AccLM). AccLM is computed by Qwen2-72B-Instruct.}
    \label{tab:main:ragqa-perfor}
\end{table*}

\begin{table}[t]
%\begin{minipage}[b]{.53\linewidth}
\centering
{\footnotesize
  \begin{tabular}{@{}l@{\hspace{3pt}}|c@{\hspace{2pt}}c@{\hspace{2pt}}c@{\hspace{3pt}}c@{\hspace{3pt}}c@{\hspace{3pt}}c@{\hspace{3pt}}|c@{\hspace{3pt}}c@{\hspace{2pt}}c@{\hspace{2pt}}c@{\hspace{3pt}}
  c@{\hspace{3pt}}c@{\hspace{3pt}}c@{}}
    \toprule %\ \\[-1ex]
    
\multicolumn{1}{l}{} 
  & \multicolumn{6}{c@{}}{\textsc{Gemma2-9B}} 
  & \multicolumn{6}{c@{}}{\textsc{Gemma2-27B}} \\
         & {\scriptsize NQ} & {\scriptsize TQA} & {\scriptsize WebQ}  
         & {\scriptsize SQuAD} & {\scriptsize PopQA}  & {\scriptsize RefuNQ}
         & {\scriptsize NQ} & {\scriptsize TQA} & {\scriptsize WebQ}  
         & {\scriptsize SQuAD} & {\scriptsize PopQA} & {\scriptsize RefuNQ} \\

\midrule %\ \\[-1.9ex]
PPL & 0.69 & 0.84 & 0.63 & 0.57 & 0.56 & 0.45 %G9B
 & 0.67 & 0.78 & 0.63 & 0.62 & 0.56 & 0.45 \\ %G27B
p(true) & 0.75 & 0.85 & 0.71 & 0.63 & 0.71 & 0.53  %G9B
 & \textbf{0.76} & \textbf{0.89} & \textbf{0.73} & 0.67 & 0.70 & 0.54 \\ %G27B
Regular Entropy & 0.70 & 0.84 & 0.63 & 0.59 & 0.57 & 0.46   %G9B
 & 0.69 & 0.79 & 0.64 & 0.61 & 0.58 & 0.45 \\ %G27B
Semantic Entropy & 0.71 & 0.85 & 0.64 & 0.65 & 0.64 & 0.51 %G9B
 & 0.69 & 0.81 & 0.67 & 0.63 & 0.61 & 0.50 \\%G27B
Passage Utility & \textbf{0.76} & \textbf{0.90} & \textbf{0.72} & \textbf{0.74} & \textbf{0.73} & \textbf{0.64} %G9B
 & 0.72 & 0.88 & \textbf{0.73} & \textbf{0.74} & \textbf{0.73} & \textbf{0.64} \\%G27B

\bottomrule %\ \\[-1.9ex]
\multicolumn{13}{c}{} \\\toprule
\multicolumn{1}{l}{} &\multicolumn{6}{c@{}}{\textsc{Llama-3.1-8B}} 
 &\multicolumn{6}{c@{}}{\textsc{Mistral-7B-v0.3}} \\
& {\scriptsize NQ} & {\scriptsize TQA} & {\scriptsize WebQ}  
         & {\scriptsize SQuAD} & {\scriptsize PopQA}  & {\scriptsize RefuNQ}
         & {\scriptsize NQ} & {\scriptsize TQA} & {\scriptsize WebQ}  
         & {\scriptsize SQuAD} & {\scriptsize PopQA} & {\scriptsize RefuNQ} \\
\midrule
PPL & 0.73 & 0.87 & 0.71 & 0.68 & 0.69 & 0.54  %L8B
 &  0.67 & 0.83 & 0.66 & 0.66 & 0.61 & 0.54 \\ %M7B
p(true) &  \textbf{0.76} & \textbf{0.89} & \textbf{0.75} & 0.70 & \textbf{0.71} & 0.59 %L8B
 &  0.71 & 0.86 & 0.70 & 0.69 & 0.65 & 0.56 \\ %M7B
Regular Entropy & 0.73 & 0.87 & 0.72 & 0.70 & 0.69 & 0.56  %L8B
 & 0.67 & 0.84 & 0.69 & 0.65 & 0.62 & 0.51 \\ %M7B
Semantic Entropy & 0.71 & 0.87 & 0.71 & 0.70 & 0.67 & 0.54  %L8B
 & 0.68 & 0.85 & \textbf{0.71} & 0.69 & 0.66 & 0.51 \\ %M7B
Passage Utility &  0.74 & 0.87 & 0.73 & \textbf{0.73} & \textbf{0.71} & \textbf{0.65} %L8B
&  \textbf{0.72} & \textbf{0.87} & \textbf{0.71} & \textbf{0.75} & \textbf{0.71} & \textbf{0.66} \\ %M7B

\bottomrule
\end{tabular}
}
%\vspace*{-.2cm}
\captionof{table}{AURAC values for QA models based on \textsc{Gemma2-9B/27B},
  \textsc{Llama-3.1-8B}, and \textsc{Mistral-7B-v0.3} on Natural
  Questions (NQ), TriviaQA (TQA), WebQuestions (WebQ), SQuAD, PopQA,
  and RefuNQ test sets.}
\label{tab:main:unc:test:AURAC}
\end{table}

\begin{table*}[ht]
    \footnotesize
    \centering
  \begin{tabular}{@{}l@{\hspace{7pt}}c@{\hspace{6pt}}c@{\hspace{4pt}}c@{\hspace{4pt}}c@{\hspace{4pt}}c@{\hspace{4pt}}|c@{\hspace{6pt}}c@{\hspace{4pt}}c@{\hspace{4pt}}c@{\hspace{4pt}}c@{\hspace{4pt}}c@{}} \\
  \hline \ \\[-1ex]
 &  \multicolumn{5}{c|}{AUROC} & \multicolumn{5}{c}{AURAC} \\
  \textsc{Gemma2-9B}         & NQ & TQA & WebQ  &  SQuAD & PopQA 
         & NQ & TQA & WebQ  &  SQuAD & PopQA \\

  \hline \ \\[-1.9ex]
PPL & 0.61 & 0.52 & 0.58 & 0.66 & 0.56 & 0.67 & 0.78 & 0.67 & 0.65 & 0.52 \\
MSP & 0.64 & 0.60 & 0.64 & 0.71 & 0.61 & 0.69 & 0.80 & 0.69 & 0.67 & 0.56 \\
PMI & 0.53 & 0.46 & 0.52 & 0.50 & 0.48 & 0.64 & 0.75 & 0.64 & 0.57 & 0.50 \\
p(true) & 0.70 & 0.71 & 0.66 & 0.73 & 0.83 & 0.72 & 0.84 & 0.70 & 0.69 & \textbf{0.71} \\
Regular Entropy & 0.64 & 0.54 & 0.60 & 0.70 & 0.58 & 0.69 & 0.78 & 0.68 & 0.67 & 0.54 \\
Cluster Assignment & 0.68 & 0.65 & 0.65 & 0.70 & 0.68 & 0.71 & 0.82 & 0.70 & 0.67 & 0.60 \\
Semantic Entropy & 0.67 & 0.69 & 0.64 & 0.72 & 0.69 & 0.71 & 0.84 & 0.69 & 0.68 & 0.61 \\
AvgAnsLen & 0.61 & 0.64 & 0.65 & 0.63 & 0.68 & 0.68 & 0.83 & 0.71 & 0.65 & 0.61 \\
Retriever Score & 0.53 & 0.62 & 0.53 & 0.67 & 0.64 & 0.65 & 0.82 & 0.63 & 0.67 & 0.61 \\
Passage Utility & \textbf{0.72} & \textbf{0.84} & \textbf{0.75} & \textbf{0.85} & \textbf{0.85} & \textbf{0.75} & \textbf{0.89} & \textbf{0.77} & \textbf{0.77} & \textbf{0.71} \\ \bottomrule
 \multicolumn{11}{c}{} \\  \toprule \ \\[-2ex]
  &  \multicolumn{5}{c|}{AUROC} & \multicolumn{5}{c}{AURAC} \\
  \textsc{Gemma2-27B}         & NQ & TQA & WebQ  &  SQuAD & PopQA 
         & NQ & TQA & WebQ  &  SQuAD & PopQA \\ \hline

PPL & 0.61 & 0.56 & 0.55 & 0.63 & 0.53 & 0.68 & 0.79 & 0.65 & 0.67 & 0.52 \\
MSP & 0.64 & 0.66 & 0.59 & 0.67 & 0.60 & 0.70 & 0.82 & 0.67 & 0.69 & 0.56 \\
PMI & 0.51 & 0.52 & 0.56 & 0.54 & 0.56 & 0.64 & 0.78 & 0.67 & 0.62 & 0.56 \\
p(true) & \textbf{0.76} & 0.73 & 0.69 & 0.69 & 0.79 & \textbf{0.78} & 0.84 & 0.74 & 0.71 & 0.70 \\
Regular Entropy & 0.65 & 0.53 & 0.56 & 0.64 & 0.53 & 0.71 & 0.78 & 0.66 & 0.67 & 0.52 \\
Cluster Assignment & 0.66 & 0.67 & 0.59 & 0.66 & 0.66 & 0.71 & 0.82 & 0.67 & 0.68 & 0.60 \\
Semantic Entropy & 0.64 & 0.67 & 0.59 & 0.68 & 0.66 & 0.69 & 0.82 & 0.68 & 0.69 & 0.60 \\
AvgAnsLen & 0.63 & 0.68 & 0.65 & 0.60 & 0.69 & 0.69 & 0.83 & 0.72 & 0.66 & 0.61 \\
Retriever Score & 0.56 & 0.60 & 0.51 & 0.69 & 0.65 & 0.67 & 0.81 & 0.64 & 0.71 & 0.62 \\
Passage Utility & 0.73 & \textbf{0.75} & \textbf{0.72} & \textbf{0.84} & \textbf{0.87} & 0.75 & \textbf{0.86} & \textbf{0.75} & \textbf{0.78} & \textbf{0.73} \\ \bottomrule
\multicolumn{11}{c}{} \\ \toprule \ \\[-2ex]
&  \multicolumn{5}{c|}{AUROC} & \multicolumn{5}{c}{AURAC} \\ 
  \textsc{Llama-3.1-8B}         & NQ & TQA & WebQ  &  SQuAD & PopQA 
         & NQ & TQA & WebQ  &  SQuAD & PopQA \\\hline 
PPL & 0.75 & 0.78 & 0.68 & 0.75 & 0.81 & 0.76 & 0.85 & 0.71 & 0.71 & 0.68 \\
MSP & 0.77 & 0.80 & 0.71 & 0.76 & 0.85 & 0.76 & 0.85 & 0.72 & 0.72 & 0.70 \\
PMI & 0.55 & 0.52 & 0.48 & 0.54 & 0.58 & 0.64 & 0.73 & 0.60 & 0.61 & 0.53 \\
p(true) & \textbf{0.80} & \textbf{0.86} & 0.72 & \textbf{0.82} & 0.85 & \textbf{0.78} & \textbf{0.87} & 0.74 & \textbf{0.75} & \textbf{0.71} \\
Regular Entropy & 0.77 & 0.80 & 0.69 & 0.76 & 0.83 & 0.76 & 0.85 & 0.71 & 0.72 & 0.69 \\
Cluster Assignment & 0.75 & 0.83 & 0.69 & 0.75 & 0.82 & 0.75 & 0.85 & 0.71 & 0.71 & 0.67 \\
Semantic Entropy & 0.74 & 0.83 & 0.70 & 0.74 & 0.81 & 0.75 & 0.86 & 0.72 & 0.71 & 0.68 \\
AvgAnsLen & 0.73 & 0.73 & 0.69 & 0.69 & 0.84 & 0.73 & 0.82 & 0.71 & 0.67 & 0.69 \\
Retriever Score & 0.58 & 0.63 & 0.54 & 0.68 & 0.66 & 0.65 & 0.79 & 0.62 & 0.66 & 0.60 \\
Passage Utility & 0.78 & 0.85 & \textbf{0.74} & \textbf{0.82} & \textbf{0.86} & 0.75 & \textbf{0.87} & \textbf{0.75} & \textbf{0.75} & \textbf{0.71} \\
\bottomrule
\multicolumn{11}{c}{}  \\ \toprule \ \\[-2ex]
&  \multicolumn{5}{c|}{AUROC} & \multicolumn{5}{c}{AURAC} \\
 \textsc{Mistral-7B-v0.3}        & NQ & TQA & WebQ  &  SQuAD & PopQA 
         & NQ & TQA & WebQ  &  SQuAD & PopQA \\\hline 
PPL & 0.66 & 0.70 & 0.60 & 0.63 & 0.66 & 0.69 & 0.84 & 0.72 & 0.63 & 0.63 \\
MSP & 0.70 & 0.75 & 0.65 & 0.71 & 0.77 & 0.70 & 0.85 & 0.73 & 0.68 & 0.67 \\
PMI & 0.38 & 0.33 & 0.42 & 0.42 & 0.30 & 0.53 & 0.68 & 0.62 & 0.52 & 0.39 \\
p(true) & 0.72 & \textbf{0.82} & 0.71 & 0.75 & 0.74 & 0.71 & \textbf{0.87} & 0.76 & 0.71 & 0.64 \\
Regular Entropy & 0.67 & 0.71 & 0.63 & 0.66 & 0.68 & 0.69 & 0.85 & 0.73 & 0.66 & 0.63 \\
Cluster Assignment & 0.72 & 0.81 & 0.68 & 0.73 & 0.76 & 0.71 & \textbf{0.87} & 0.75 & 0.68 & 0.66 \\
Semantic Entropy & 0.72 & 0.80 & 0.68 & 0.73 & 0.76 & 0.71 & \textbf{0.87} & 0.76 & 0.69 & 0.66 \\
AvgAnsLen & 0.66 & 0.75 & 0.65 & 0.68 & 0.81 & 0.69 & 0.85 & 0.73 & 0.67 & 0.70 \\
Retriever Score & 0.55 & 0.63 & 0.55 & 0.65 & 0.68 & 0.63 & 0.81 & 0.67 & 0.64 & 0.63 \\
Passage Utility & \textbf{0.76} & 0.81 & \textbf{0.75} & \textbf{0.85} & \textbf{0.85} & \textbf{0.74} & \textbf{0.87} & \textbf{0.78} & \textbf{0.75} & \textbf{0.71} \\
\bottomrule

\hline
\end{tabular}
    \caption{Answer uncertainty estimation for QA models \textsc{Gemma2-9/27B}, \textsc{Llama-3.1-8B}, and \textsc{Mistral-7B-v0.3} on NQ, TQA, WebQ, SQuAD, and PopQA development sets. We report AUROC and AURAC. 
    }
    \label{tab:main:unc:dev}
\end{table*}

\section{Examples of False Positives and Negatives}
\label{sec:app:unc:examples}

Tables~\ref{tab:ex:TN:squad:llama318bit}--\ref{tab:ex:FN:tqa:gemma29bit}
illustrate the working of Passage Utility for answer uncertainty
estimation.  As we report AUROC scores, we do not set any
correct/incorrect decision threshold; for the purpose of this
discussion, we assume a decision point at 0.5 and analyze clear success
and failure cases. For each example, we show the question, gold, and
generated answers in the top block. Then, we show three retrieved
passages with their estimated Passage Utility and a final block with
ten sampled answers, their grouping into clusters, and the Cluster
Assignment entropy.

Table~\ref{tab:ex:TN:squad:llama318bit} shows an example for a SQuAD
question and the \textsc{Llama-3.1-8B} QA model. In this case, the QA
model correctly answers and the Passage Utility estimate is high
(i.e., indicating the answer is correct). Table~\ref{tab:ex:TP:squad:llama318bit} illustrates a case
where \textsc{Llama-3.1-8B}'s answer is incorrect and all Passage
Utilities are very low (i.e.,~indicating the answer is incorrect).  The
example from NQ in Table~\ref{tab:ex:FP:nq:gemma29bit} shows a case
where all Passage Utilities are low but the QA model
(\textsc{Gemma2-9B}) answers correctly. The first passage is not
useful, the second does not explicitly mention the answer but still
primes the QA model to  answer correctly, while the third passage 
mentions the answer.

In Table~\ref{tab:ex:FN:tqa:gemma29bit},
Passage Utility scores are high estimating a correct answer for the
TQA test question; however, \textsc{Gemma2-9B} answers with the
incorrect magazine name. Note that none of the passages corresponds to
the National Geographic magazine but have high token overlap with the
question (in particular the first and second passages).

% Example TN by Passage Utility 
\begin{table*}[t]
\tcbset{colback=white,colframe=green!45!black}
\begin{tcolorbox}[arc=4mm,outer arc=1mm, top=1mm, bottom=1mm]
\begin{flushleft}
\footnotesize{Question: How many platforms does Manchester Oxford Road have?\\
Generated Answer: Four through platforms and one terminating bay platform.\\
Gold Answer: 5.}\end{flushleft}
\end{tcolorbox}
\vspace{-.2cm}
\begin{tcolorbox}[colback=white,colframe=red,enhanced,title=\texttt{\scriptsize Originally one, later expanded.}, boxed title style={colback=black}, colbacktitle=white, coltitle=black, left=1mm, top=1mm, right=1mm]
\scriptsize{\texttt{Manchester Oxford Road railway station. Gloucester Street (now Whitworth Street West) to reach their north front. There was a single platform on the north side of the line through to Manchester London Road (now "Manchester Pic
adilly") and a second platform on a west-facing siding. To allow for extra trains in connection with the Manchester Art Treasures Exhibition in 1857, extra platforms and sidings were bui
lt, but afterwards removed. A meeting of MSJAR contract ticket holders in 1863 included in its list of complaints the want of punctuality "especially as at Oxford Road station there is only one platform used for both passengers and milk".
\hfill \textcolor{blue}{\texttt -0.42}}}
\vspace*{-.1cm}
\end{tcolorbox}
\vspace{-.2cm}
\begin{tcolorbox}[colback=white,colframe=red,enhanced,title=\texttt{\scriptsize{Twelve terminating, two through.}},
boxed title style={colback=white}, colbacktitle=white, coltitle=black, left=1mm, top=1mm, right=1mm]
\scriptsize{\texttt{Transport in Manchester. platforms are planned as part of the Northern Hub project however in 2017 the construction was not confirmed. At present there are twelve terminating platforms but just two through platforms. Congestion is common at peak times and trains often incur delays for following trains down the line at Oxford Road. The new proposals will provide up to 700 more trains per day with space for 44 million more people to travel by train each year. Victoria station has undergone a series of upgrades. The station was rated as the worst in the UK following a 2009 study 
and plans for \hfill \textcolor{blue}{\texttt -0.26}}}
\vspace*{-.1cm}
\end{tcolorbox}
\vspace{-.2cm}
\begin{tcolorbox}[colback=white,colframe=green!45!black,enhanced,title=\texttt{\scriptsize{Five platforms.}},
boxed title style={colback=red}, colbacktitle=white, coltitle=black, left=1mm, top=1mm, right=1mm]
  \scriptsize{\texttt{Manchester Oxford Road railway station. Manchester Oxford Road railway station is a railway station in Manchester, England, at the junction of Whitworth Street West and Oxford Street. It opened in 1849 and was rebuilt in 1960. It is the second busiest of the four stations in Manchester city centre. The station serves the southern part of Manchester city centre, the University of Manchester and Manchester Metropolitan University, on the line from westwards towards Warrington, Chester, Llandudno, Liverpool, and Blackpool. Eastbound trains go beyond Piccadilly to , and . The station consists of four through platforms and one terminating bay platform. \hfill \textcolor{blue}{\texttt 1.34}}}
\end{tcolorbox}
\vspace{-.2cm}
\begin{tcolorbox}[colback=white,colframe=red,enhanced, left=1mm, top=1mm, right=1mm]
\scriptsize{\texttt{['Four through platforms and one terminating bay platform.', 'Four through platforms and one terminating bay.', 'Four through platforms and one terminating bay platform.'], ['Four through and one terminating.', 'Four through platforms and one terminating bay.', 'Four through and one bay.'], ['Five platforms.',  'Five platforms.', 'Five platforms.', 'Five platforms.']\hfill \textcolor{blue}{\texttt 1.09}}}
%[0, 0, 0, 1, 1, 2, 1, 2, 2, 2]
\end{tcolorbox}
    \caption{True negative example (from NQ development set): Passage Utility predicts the right answer as well as the  QA model (\textsc{Gemma2-9B}).}
    \label{tab:ex:TN:squad:llama318bit}
\end{table*}

% Example TP by Passage Utility 
\begin{table*}[t]
\tcbset{colback=white,colframe=red}
\begin{tcolorbox}[arc=4mm,outer arc=1mm, top=1mm, bottom=1mm]
\begin{flushleft}
\footnotesize{Question: Which company was targeted by the NAACP for not having fair practices?\\
Generated Answer: Target Corporation.\\
Gold Answer: Lockheed Aircraft Corporation.}\end{flushleft}
\end{tcolorbox}
\vspace{-.2cm}
\begin{tcolorbox}[colback=white,colframe=red,enhanced,title=\texttt{\scriptsize Target Corporation.}, boxed title style={colback=black}, colbacktitle=white, coltitle=black, left=1mm, top=1mm, right=1mm]
\scriptsize{\texttt{Target Corporation. of Colored People has repeatedly given Target failing grades on its annual Economic Reciprocity Initiative report card, a measure of the company's ``commitment to the African-American citizenry''. In 2003 and 2005, the NAACP has rated Target an ``F'' on this report; in 2004, Target was rated a ``D-''. In 2006, when Target was asked why it didn't participate in the survey again, a representative explained, ``Target views diversity as being inclusive of all people from all different backgrounds, not just one group.'' In February 2006, the National Federation of the Blind (NFB) filed a class action [cont.] \hfill \textcolor{blue}{\texttt -2.64}}}
\vspace*{-.1cm}
\end{tcolorbox}
\vspace{-.2cm}
\begin{tcolorbox}[colback=white,colframe=red,enhanced,title=\texttt{\scriptsize{None, the NAACP was involved in the Duke lacrosse case.}},
boxed title style={colback=white}, colbacktitle=white, coltitle=black, left=1mm, top=1mm, right=1mm]
\scriptsize{\texttt{Reactions to the Duke lacrosse case. formed an opinion on the case. North Carolina NAACP Legal Redress Chair, Al McSurely, explained that ``The NAACP stands for fair play for all parties, zealous investigation and deep concern for the survivors of racist/sexist attacks.'' At the same time, some have criticized the NAACP for making statements that portrayed the players as racist despite evidence to the contrary, using the case to promote the group's cause, and implying guilt. McSurely stated that ``[w]ithin five minutes, the men threatened the women with racial and misogynist verbal assaults, [cont.]\hfill \textcolor{blue}{\texttt -2.77}}}
\vspace*{-.1cm}
\end{tcolorbox}
\vspace{-.2cm}
\begin{tcolorbox}[colback=white,colframe=red,enhanced,title=\texttt{\scriptsize{Philadelphia Transit Company (PTC).}},
boxed title style={colback=red}, colbacktitle=white, coltitle=black, left=1mm, top=1mm, right=1mm]
  \scriptsize{\texttt{Philadelphia transit strike of 1944. (PRTEU), Frank Carney, 
proved to be equally reticent and claimed that he was not authorized by the union members to consider a request to allow promotions of black employees. The black PTC employees enlisted the help of the NAACP and started lobbying the federal authorities, particularly the Fair Employment Practices Commission (FEPC), to intervene. The Fair Employment Practices Commission, created by an executive order of the President in 1941, was charged with ensuring non-discrimination employment practices by government contractors. [cont.] \hfill \textcolor{blue}{\texttt -2.60}}}
\end{tcolorbox}
\vspace{-.2cm}
\begin{tcolorbox}[colback=white,colframe=green!45!black,enhanced, left=1mm, top=1mm, right=1mm]
\scriptsize{\texttt{['Target.'], ['Target Corporation.', 'Target Corporation.', 'Target Corporation.', 'Target Corporation.', 'Target Corporation.', 'Target Corporation.', 'Target Corporation.', 'Target Corporation.'], ['Target Corporation and the National Federation of the Blind filed a lawsuit against Michigan.']\hfill \textcolor{blue}{\texttt 0.33}}}
% [0, 1, 1, 1, 1, 1, 1, 1, 2, 1]
\end{tcolorbox}
    \caption{True positive example (from SQuAD development set):
    Passage Utility predicts the answer provided by the  QA model
    (\textsc{Llama-3.1-8B}) which is incorrect. 
    }
    \label{tab:ex:TP:squad:llama318bit}
\end{table*}

% Example FP by Passage Utility 
\begin{table*}[t]
\tcbset{colback=white,colframe=green!45!black}
\begin{tcolorbox}[arc=4mm,outer arc=1mm, top=1mm, bottom=1mm]
\begin{flushleft}
\footnotesize{Question: What do Alfred Newman, John Williams, Johnny Green, Andr Previn, John Barry and Alan Menken have in common?\\
Generated Answer: They are all Oscar-winning film composers.\\
Gold Answer: Winning the Academy award for Best Original Score.}\end{flushleft}
\end{tcolorbox}
\vspace{-.2cm}
\begin{tcolorbox}[colback=white,colframe=red,enhanced,title=\texttt{\scriptsize They all won multiple Academy Awards for film scores.}, boxed title style={colback=black}, colbacktitle=white, coltitle=black, left=1mm, top=1mm, right=1mm]
\scriptsize{\texttt{Alan Menken. his other film musicals were also nominated: Alan Menken Alan Irwin Menken (born July 22, 1
949) is an American musical theatre and film score composer and pianist. Menken is best known for his scores for films produced by Walt Disney Animation Studios. His scores for "The Little Mermaid" (1989
), "Beauty and the Beast" (1991), "Aladdin" (1992), and "Pocahontas" (1995) have each won him two Academy Awards. He also composed the scores for "Little Shop of Horrors" (1986), "Newsies" (1992), "The Hunchback of Notre Dame" (1996), "Hercules" (1997), "Home on the Range" (2004), "Enchanted" (2007), "Tangled" (2010), and "Sausage Party" (2016) \hfill \textcolor{blue}{\texttt -2.22}}}
\vspace*{-.1cm}
\end{tcolorbox}
\vspace{-.2cm}
\begin{tcolorbox}[colback=white,colframe=red,enhanced,title=\texttt{\scriptsize{They are all prolific Oscar winners in music categories.}},
boxed title style={colback=white}, colbacktitle=white, coltitle=black, left=1mm, top=1mm, right=1mm]
\scriptsize{\texttt{Alan Menken. hi
s work on musical theatre works for Broadway and elsewhere. Some of these are based on his Disney films, but other stage hits include "Little Shop of Horrors" (1982), "A Christmas Carol" (1994) and "S
ister Act" (2009). Menken has collaborated with such lyricists as Howard Ashman, Tim Rice, Glenn Slater, Stephen Schwartz and David Zippel. With eight Academy Award wins (four each for Best Score and Best
 Song), Menken is the second most prolific Oscar winner in the music categories after Alfred Newman, who has nine Oscars. He has also won eleven Grammy Awards, a Tony Award and other honors. \hfill \textcolor{blue}{\texttt -2.30}}}
\vspace*{-.1cm}
\end{tcolorbox}
\vspace{-.2cm}
\begin{tcolorbox}[colback=white,colframe=red,enhanced,title=\texttt{\scriptsize{They are all Academy Award-winning composers.}},
boxed title style={colback=red}, colbacktitle=white, coltitle=black, left=1mm, top=1mm, right=1mm]
  \scriptsize{\texttt{Alan Menken. the Beast", with the so
ngs from the 1991 film and new material by Menken and Rice. , Menken is collaborating on writing new songs with Benj Pasek and Justin Paul for a live-action film version of "Aladdin" and is also working 
with Lin-Manuel Miranda on new music for a live-action film adaptation of "The Little Mermaid". Menken will also once again be working with Stephen Schwartz to write new songs for \"Disenchanted\", the s
equel to "Enchanted". With eight Academy Awards (four each for best score and best song), only composer Alfred Newman (nine wins) and Walt Disney (22 wins) have \hfill \textcolor{blue}{\texttt -2.31}}}
\end{tcolorbox}
\vspace{-.2cm}
\begin{tcolorbox}[colback=white,colframe=red,enhanced, left=1mm, top=1mm, right=1mm]
\scriptsize{\texttt{['They are all Academy Award-winning composers.', 'They are all Academy Award-winning composers.', 'They are all Oscar-winning film composers.', 'They are all Oscar-winning film composers.'], ['They are all multiple-time Oscar winners for music.', 'They are prolific Academy Award-winning composers.'], ['They are all multi-Oscar winning film composers.', 'They are all multiple Oscar-winning composers.', 'They are prolific Oscar-winning film composers.'], ['They are all prolific film music composers.'] \hfill \textcolor{blue}{\texttt 1.28}}}
% [0, 0, 1, 2, 3, 4, 3, 4, 1, 0]
\end{tcolorbox}
    \caption{False positive example (from TQA development set): Passage Utility predicts a wrong answer whereas the prediction of the QA model (\textsc{Gemma2-9B}) is correct.
    }
    \label{tab:ex:FP:nq:gemma29bit}
\end{table*}

% Example FN by Passage Utility 
\begin{table*}[t!]
\tcbset{colback=white,colframe=red}
\begin{tcolorbox}[arc=4mm,outer arc=1mm, top=1mm, bottom=1mm]
\begin{flushleft}
\footnotesize{Question: One Step Behind' is a tribute act to which band?\\
Generated Answer: Furniture.\\
Gold Answer: North London Invaders.}\end{flushleft}
\end{tcolorbox}
\vspace{-.2cm}
\begin{tcolorbox}[colback=white,colframe=green!45!black,enhanced,title=\texttt{\scriptsize Def L
eppard.}, boxed title style={colback=green!45!black}, colbacktitle=white, coltitle=black, left=1mm, top=1mm, right=1mm]
\scriptsize{\texttt{Two Steps Behind. Two Steps Behind "Two Steps Behind" is a 1993 song by English hard rock band Def Leppard from their album "Retro Active" and "Last Action Hero" Soundtrack. It reached \#5 on the "Billboard" Hot Mainstream Rock Tracks chart, and \#12 on the \"Billboard\" Hot 100. Though the band did manage to chart a few more songs in the following years, this is generally considered to be the band's last major hit in the US. In the 1993 "Metal Edge" Readers' Choice Awards, the song was voted "Song of the Year" and "Best Song From a Movie Soundtrack.". Def Leppard have \hfill \textcolor{blue}{\texttt 3.29}}}
\vspace*{-.1cm}
\end{tcolorbox}
\vspace{-.2cm}
\begin{tcolorbox}[colback=white,colframe=green!45!black,enhanced,title=\texttt{\scriptsize{Def Leppard.}},
boxed title style={colback=green!45!black}, colbacktitle=white, coltitle=black, left=1mm, top=1mm, right=1mm]
  \scriptsize{\texttt{Two Steps Behind. and a live footage. The video was aired on August 19 93. Filipino-Chinese singer Rachelle Ann Go covered the song for her 2007 album "Obsession". Two Steps Behind "Two Steps Behind" is a 1993 song by English hard rock band Def Leppard from their album "Retro Active" and "Last Action Hero" Soundtrack. It reached \#5 on the "Billboard" Hot Mainstream Rock Tracks chart, and \#12 on the "Billboard" Hot 100. Though the band did manage to chart a few more songs in the following years, this is generally considered to be the band's last major hit in the US. In the \hfill \textcolor{blue}{\texttt 2.42}}}
\vspace*{-.1cm}
\end{tcolorbox}
\vspace{-.2cm}
\begin{tcolorbox}[colback=white,colframe=red,enhanced,title=\texttt{\scriptsize{Split Enz.}},
boxed title style={colback=red}, colbacktitle=white, coltitle=black, left=1mm, top=1mm, right=1mm]
  \scriptsize{\texttt{One Step Ahead (Split Enz song). unavailable to Australasian markets until 2007 when it became available on iTunes). The video clip to "One Step Ahead" has keyboardist Eddie Rayner performing "Marche sur place", the pantomime illusion walk created by Decroux and Barrault (seen in the 1945 French film Child
ren of Paradise) that is the technique Michael Jackson would base his moonwalk on in 1983. One Step Ahead (Split Enz song) "One Step Ahead" is a 1980 song by New Zealand art rock group Split Enz from the
ir studio album "Waiata". The song continued the group's success in their move towards their own version of new wave \hfill \textcolor{blue}{\texttt -2.93}}}
\end{tcolorbox}
\vspace{-.2cm}
\begin{tcolorbox}[colback=white,colframe=green!45!black,enhanced, left=1mm, top=1mm, right=1mm]
\scriptsize{\texttt{['Furniture', 'Furniture', 'Furniture', 'Furniture', 'Furniture', 'Furniture', 'Furniture', 'Furniture', 'Furniture', 'Furniture'] \hfill \textcolor{blue}{\texttt 0}}}
\end{tcolorbox}
    \caption{False negative (from TQA development set): Passage Utility predicts a correct answer, and the answer by the QA model (\textsc{Gemma2-9B}) is wrong.
    }
    \label{tab:ex:FN:tqa:gemma29bit}
\end{table*}

\end{document}

%% file: math_commands.tex
\usepackage{amsmath,amsfonts,bm}

\def\eqref#1{equation~\ref{#1}}
\def\1{\bm{1}}

\DeclareMathAlphabet{\mathsfit}{\encodingdefault}{\sfdefault}{m}{sl}
\SetMathAlphabet{\mathsfit}{bold}{\encodingdefault}{\sfdefault}{bx}{n}

%% file: CostApproachGraph3.tex
% used PGFPlots v1.16
%\documentclass[border=5pt]{standalone}
%\usepackage{pgfplots}
%\usepackage{tikz,xcolor}

\definecolor{babyblueeyes}{rgb}{0.63, 0.79, 0.95}
%\usepgfplotslibrary{colormaps,scatter}
%    \pgfplotsset{
 %       compat=1.3,
%    }
%\begin{document}
\begin{tikzpicture}[scale=0.60]
    \begin{axis}[
        width=4.7in,
        height=3in,
        xlabel= Number of Inference Tokens,
        ylabel=Average AUROC,
          %colorbar,
          xtick={700,9000,13000},
          xticklabels={700, 9k, 13k},
          scaled ticks = false,
          ymin=0.5,
        %   legend pos=north,         % places legend above plot area inside the axis
    legend columns=-1,        % arranges legend entries horizontally in one row
    legend cell align={left}, % optional, aligns legend entries nicely
    legend style={
        at={(0.5,1.05)},      % shifts legend slightly above the axis (adjust 1.05 as needed)
        anchor=south,         % anchors legend box by its bottom center
        draw=none,            % optional, removes box around legend
        /tikz/every even column/.append style={column sep=0.5cm} % spacing between columns
    },
          %enlarge x limits=0.2,
        % (it is simpler to use `xtick distance`)
%        xtick={0,50,100,150,200,250,300,350,400},
        %xtick distance=50,
    ]

    \addplot[
    scatter,
    only marks,
    mark=*,
    point meta=explicit,
    scatter src=explicit symbolic,
    visualization depends on={value \thisrow{model} \as \model},
    scatter/classes={
        g9={mark=*, draw=babyblueeyes, fill=babyblueeyes},
        g27={mark=*, draw=blue, fill=blue}
    },
    visualization depends on={0.5*\thisrow{Val} \as \perpointmarksize},
    scatter/@pre marker code/.append style={
        /tikz/mark size=\perpointmarksize,
    },
    point meta=explicit symbolic,
    nodes near coords*={\Label},
    visualization depends on={value \thisrow{label} \as \Label},
    visualization depends on={value \thisrow{Pos} \as \myPos},
    visualization depends on={\thisrow{addOffset} \as \myOffset},
    visualization depends on={\thisrow{Val} \as \myval},
    every node near coord/.append style={
        font=\scriptsize,
        \myPos=\perpointmarksize pt + \myOffset pt,
    },
] 
       table [y={Length},x={Info},meta=model] {
       Info    Length  Val   label                 anchor  Pos  addOffset model
772     0.56    27    {PPL}           south   right 0        g27
11954   0.72    27    {p(true)}       south   above 0        g27
8868    0.63    27    {SE}            north   right 0        g27
772     0.78     3    {PU}            south   right 0        g27
810     0.58    10    {PPL}            south   right 0        g9
13841   0.70    10    {\hspace{0.4em}p(true)}  south   above 0 g9
9292    0.66    10    {SE}             north   right 0        g9
810     0.79     3    {PU}             south   right 0        g9
};

\addplot[only marks, mark=*, color=babyblueeyes] coordinates {(0,0)};
\addlegendentry{\textsc{Gemma2-9B}}

\addplot[only marks, mark=*, color=red] coordinates {(0,0)};
\addlegendentry{\textsc{Gemma2-27B}}

  \end{axis}
\end{tikzpicture}
%\end{document}

%% file: ModelAcc80Graph.tex
\begin{tikzpicture}
\begin{axis}[
height=4.5cm,
width=7cm,
symbolic x coords={\textsc{G9B}, \textsc{G27B}, \textsc{L8B}, \textsc{M7B}},
xtick=data,
xticklabel style={anchor=near xticklabel},
extra y tick style={grid=major, tick label style={xshift=-1cm}},
ylabel={Average QA Accuracy},
xlabel={Model variants},
ymin=0.6,
y tick label style={font=\scriptsize},
ylabel style={font=\scriptsize},
x tick label style={font=\scriptsize},
xlabel style={font=\scriptsize},
y label style={at={(axis description cs:-0.16,.5)},anchor=south},
legend columns=2,transpose legend,
        %legend cell align = left,
        legend style={
                at={(-0.35,1.0)},
    inner sep=0.8em,
	      anchor=south west ,
              %column sep=1ex,
              draw=none, legend columns=-1,
              font=\scriptsize,
              cells={align=center},
        } 
]
\addplot[pink, mark=square*] table[x=date,y=value] {acc80_ppl.dat};
\addplot[skybleu, mark=triangle*] table[x=date,y=value] {acc80_ptrue.dat};
\addplot[bluishGreen, mark=*] table[x=date,y=value] {acc80_re.dat};
\addplot[brown, mark=diamond*] table[x=date,y=value] {acc80_se.dat};
\addplot[gray, mark=10-pointed star] table[x=date,y=value] {acc80_utilityranker.dat};
\addplot[only marks] table[x=date,y=value] {acc80_acc100.dat};
      \addlegendentry{PPL}
      \addlegendentry{p(true)}
      \addlegendentry{Regular Entropy}
      \addlegendentry{Semantic Entropy}
      \addlegendentry{Passage Utility}
\end{axis}
\end{tikzpicture}

%% file: PassageRerankGraph.tex
\begin{tikzpicture}
\begin{axis}[
height=4.2cm,
width=7cm,
symbolic x coords={\textsc{5}, \textsc{3}, \textsc{1}},
xtick=data,
xticklabel style={anchor=near xticklabel},
extra y tick style={grid=major, tick label style={xshift=-1cm}},
ylabel={Average Accuracy},
xlabel={Number of Retrieved Passages},
ymin=0.43,
ymax=0.7,
y tick label style={font=\scriptsize},
ylabel style={font=\scriptsize},
x tick label style={font=\scriptsize},
xlabel style={font=\scriptsize},
y label style={at={(axis description cs:-0.16,.5)},anchor=south},
        legend cell align = left,
        legend style={
                at={(-0.05,1.03)},
    inner sep=0.8em,
	      anchor=south west ,
              %column sep=1ex,
              draw=none, legend columns=3,
              font=\scriptsize
        } 
]
\addplot[gray, mark=square*] table[x=date,y=value] {acclm_rr_retriever.dat};
\addplot[red!60, mark=10-pointed star] table[x=date,y=value] {acclm_rr_pu.dat};
\addplot[blue!60, mark=10-pointed star] table[x=date,y=value] {acclm_rr_ppl.dat};
\addlegendentry{RAG}
\addlegendentry{RAG+PU}
\addlegendentry{RAG+PPL}
\end{axis}
\end{tikzpicture}

%% file: ragu_preprint.bbl
\begin{thebibliography}{90}
\providecommand{\natexlab}[1]{#1}
\providecommand{\url}[1]{\texttt{#1}}
\expandafter\ifx\csname urlstyle\endcsname\relax
  \providecommand{\doi}[1]{doi: #1}\else
  \providecommand{\doi}{doi: \begingroup \urlstyle{rm}\Url}\fi

\bibitem[AI@Meta(2024)]{llama3modelcard}
AI@Meta.
\newblock Llama 3 model card.
\newblock 2024.
\newblock URL
  \url{https://github.com/meta-llama/llama3/blob/main/MODEL_CARD.md}.

\bibitem[Asai et~al.(2024)Asai, Wu, Wang, Sil, and Hajishirzi]{asai2024selfrag}
Akari Asai, Zeqiu Wu, Yizhong Wang, Avirup Sil, and Hannaneh Hajishirzi.
\newblock Self-{RAG}: Learning to retrieve, generate, and critique through
  self-reflection.
\newblock In \emph{The Twelfth International Conference on Learning
  Representations}, 2024.
\newblock URL \url{https://openreview.net/forum?id=hSyW5go0v8}.

\bibitem[Berant et~al.(2013)Berant, Chou, Frostig, and
  Liang]{berant-etal-2013-semantic}
Jonathan Berant, Andrew Chou, Roy Frostig, and Percy Liang.
\newblock Semantic parsing on {F}reebase from question-answer pairs.
\newblock In David Yarowsky, Timothy Baldwin, Anna Korhonen, Karen Livescu, and
  Steven Bethard (eds.), \emph{Proceedings of the 2013 Conference on Empirical
  Methods in Natural Language Processing}, pp.\  1533--1544, Seattle,
  Washington, USA, October 2013. Association for Computational Linguistics.
\newblock URL \url{https://aclanthology.org/D13-1160}.

\bibitem[Borszukovszki et~al.(2025)Borszukovszki, De~Jong, and
  Valdenegro-Toro]{borszukovszki-etal-2025-know}
Mirko Borszukovszki, Ivo~Pascal De~Jong, and Matias Valdenegro-Toro.
\newblock Know what you do not know: Verbalized uncertainty estimation
  robustness on corrupted images in vision-language models.
\newblock In Trista Cao, Anubrata Das, Tharindu Kumarage, Yixin Wan, Satyapriya
  Krishna, Ninareh Mehrabi, Jwala Dhamala, Anil Ramakrishna, Aram Galystan,
  Anoop Kumar, Rahul Gupta, and Kai-Wei Chang (eds.), \emph{Proceedings of the
  5th Workshop on Trustworthy NLP (TrustNLP 2025)}, pp.\  247--265,
  Albuquerque, New Mexico, May 2025. Association for Computational Linguistics.
\newblock ISBN 979-8-89176-233-6.
\newblock \doi{10.18653/v1/2025.trustnlp-main.16}.
\newblock URL \url{https://aclanthology.org/2025.trustnlp-main.16/}.

\bibitem[Bradley \& Terry(1952)Bradley and Terry]{bradley-terry-model}
Ralph~Allan Bradley and Milton~E. Terry.
\newblock Rank analysis of incomplete block designs: I. the method of paired
  comparisons.
\newblock \emph{Biometrika}, 39\penalty0 (3/4):\penalty0 324--345, 1952.
\newblock ISSN 00063444, 14643510.
\newblock URL \url{http://www.jstor.org/stable/2334029}.

\bibitem[Brown et~al.(2020)Brown, Mann, Ryder, Subbiah, Kaplan, Dhariwal,
  Neelakantan, Shyam, Sastry, Askell, Agarwal, Herbert-Voss, Krueger, Henighan,
  Child, Ramesh, Ziegler, Wu, Winter, Hesse, Chen, Sigler, Litwin, Gray, Chess,
  Clark, Berner, McCandlish, Radford, Sutskever, and Amodei]{fewshot-learners}
Tom Brown, Benjamin Mann, Nick Ryder, Melanie Subbiah, Jared~D Kaplan, Prafulla
  Dhariwal, Arvind Neelakantan, Pranav Shyam, Girish Sastry, Amanda Askell,
  Sandhini Agarwal, Ariel Herbert-Voss, Gretchen Krueger, Tom Henighan, Rewon
  Child, Aditya Ramesh, Daniel Ziegler, Jeffrey Wu, Clemens Winter, Chris
  Hesse, Mark Chen, Eric Sigler, Mateusz Litwin, Scott Gray, Benjamin Chess,
  Jack Clark, Christopher Berner, Sam McCandlish, Alec Radford, Ilya Sutskever,
  and Dario Amodei.
\newblock Language models are few-shot learners.
\newblock In H.~Larochelle, M.~Ranzato, R.~Hadsell, M.F. Balcan, and H.~Lin
  (eds.), \emph{Advances in Neural Information Processing Systems}, volume~33,
  pp.\  1877--1901. Curran Associates, Inc., 2020.
\newblock URL
  \url{https://proceedings.neurips.cc/paper_files/paper/2020/file/1457c0d6bfcb4967418bfb8ac142f64a-Paper.pdf}.

\bibitem[Chen et~al.(2017)Chen, Fisch, Weston, and Bordes]{chen-etal-2017-drQA}
Danqi Chen, Adam Fisch, Jason Weston, and Antoine Bordes.
\newblock Reading {W}ikipedia to answer open-domain questions.
\newblock In Regina Barzilay and Min-Yen Kan (eds.), \emph{Proceedings of the
  55th Annual Meeting of the Association for Computational Linguistics (Volume
  1: Long Papers)}, pp.\  1870--1879, Vancouver, Canada, July 2017. Association
  for Computational Linguistics.
\newblock \doi{10.18653/v1/P17-1171}.
\newblock URL \url{https://aclanthology.org/P17-1171}.

\bibitem[Chen \& Mueller(2024)Chen and Mueller]{chen-mueller-2024-quantifying}
Jiuhai Chen and Jonas Mueller.
\newblock Quantifying uncertainty in answers from any language model and
  enhancing their trustworthiness.
\newblock In Lun-Wei Ku, Andre Martins, and Vivek Srikumar (eds.),
  \emph{Proceedings of the 62nd Annual Meeting of the Association for
  Computational Linguistics (Volume 1: Long Papers)}, pp.\  5186--5200,
  Bangkok, Thailand, August 2024. Association for Computational Linguistics.
\newblock URL \url{https://aclanthology.org/2024.acl-long.283}.

\bibitem[DeLong et~al.(1988)DeLong, DeLong, and Clarke-Pearson]{delong-test}
Elizabeth~R. DeLong, David~M. DeLong, and Daniel~L. Clarke-Pearson.
\newblock Comparing the areas under two or more correlated receiver operating
  characteristic curves: A nonparametric approach.
\newblock \emph{Biometrics}, 44\penalty0 (3):\penalty0 837--845, 1988.
\newblock ISSN 0006341X, 15410420.
\newblock URL \url{http://www.jstor.org/stable/2531595}.

\bibitem[Devlin et~al.(2019)Devlin, Chang, Lee, and
  Toutanova]{devlin-etal-2019-bert}
Jacob Devlin, Ming-Wei Chang, Kenton Lee, and Kristina Toutanova.
\newblock {BERT}: Pre-training of deep bidirectional transformers for language
  understanding.
\newblock In Jill Burstein, Christy Doran, and Thamar Solorio (eds.),
  \emph{Proceedings of the 2019 Conference of the North {A}merican Chapter of
  the Association for Computational Linguistics: Human Language Technologies,
  Volume 1 (Long and Short Papers)}, pp.\  4171--4186, Minneapolis, Minnesota,
  June 2019. Association for Computational Linguistics.
\newblock \doi{10.18653/v1/N19-1423}.
\newblock URL \url{https://aclanthology.org/N19-1423}.

\bibitem[Dong et~al.(2018)Dong, Quirk, and Lapata]{dong-etal-2018-confidence}
Li~Dong, Chris Quirk, and Mirella Lapata.
\newblock Confidence modeling for neural semantic parsing.
\newblock In Iryna Gurevych and Yusuke Miyao (eds.), \emph{Proceedings of the
  56th Annual Meeting of the Association for Computational Linguistics (Volume
  1: Long Papers)}, pp.\  743--753, Melbourne, Australia, July 2018.
  Association for Computational Linguistics.
\newblock \doi{10.18653/v1/P18-1069}.
\newblock URL \url{https://aclanthology.org/P18-1069}.

\bibitem[Fadeeva et~al.(2023)Fadeeva, Vashurin, Tsvigun, Vazhentsev, Petrakov,
  Fedyanin, Vasilev, Goncharova, Panchenko, Panov, Baldwin, and
  Shelmanov]{polygraph}
Ekaterina Fadeeva, Roman Vashurin, Akim Tsvigun, Artem Vazhentsev, Sergey
  Petrakov, Kirill Fedyanin, Daniil Vasilev, Elizaveta Goncharova, Alexander
  Panchenko, Maxim Panov, Timothy Baldwin, and Artem Shelmanov.
\newblock {LM}-polygraph: Uncertainty estimation for language models.
\newblock In Yansong Feng and Els Lefever (eds.), \emph{Proceedings of the 2023
  Conference on Empirical Methods in Natural Language Processing: System
  Demonstrations}, pp.\  446--461, Singapore, December 2023. Association for
  Computational Linguistics.
\newblock \doi{10.18653/v1/2023.emnlp-demo.41}.
\newblock URL \url{https://aclanthology.org/2023.emnlp-demo.41}.

\bibitem[Fan et~al.(2018)Fan, Lewis, and Dauphin]{fan-etal-2018-hierarchical}
Angela Fan, Mike Lewis, and Yann Dauphin.
\newblock Hierarchical neural story generation.
\newblock In Iryna Gurevych and Yusuke Miyao (eds.), \emph{Proceedings of the
  56th Annual Meeting of the Association for Computational Linguistics (Volume
  1: Long Papers)}, pp.\  889--898, Melbourne, Australia, July 2018.
  Association for Computational Linguistics.
\newblock \doi{10.18653/v1/P18-1082}.
\newblock URL \url{https://aclanthology.org/P18-1082}.

\bibitem[Fang et~al.(2024)Fang, Gor, and Simpson]{fang-etal-2024-efficiently}
Haishuo Fang, Jeet Gor, and Edwin Simpson.
\newblock Efficiently acquiring human feedback with {B}ayesian deep learning.
\newblock In Ra{\'u}l V{\'a}zquez, Hande Celikkanat, Dennis Ulmer, J{\"o}rg
  Tiedemann, Swabha Swayamdipta, Wilker Aziz, Barbara Plank, Joris Baan, and
  Marie-Catherine de~Marneffe (eds.), \emph{Proceedings of the 1st Workshop on
  Uncertainty-Aware NLP (UncertaiNLP 2024)}, pp.\  70--80, St Julians, Malta,
  March 2024. Association for Computational Linguistics.
\newblock URL \url{https://aclanthology.org/2024.uncertainlp-1.7}.

\bibitem[Farquhar et~al.(2024)Farquhar, Kossen, Kuhn, and
  Gal]{farquhar2024semantic-hallu}
Sebastian Farquhar, Jannik Kossen, Lorenz Kuhn, and Yarin Gal.
\newblock Detecting hallucinations in large language models using semantic
  entropy.
\newblock \emph{Nature}, 2024.
\newblock URL \url{https://doi.org/10.1038/s41586-024-07421-0}.

\bibitem[Gal \& Ghahramani(2016)Gal and Ghahramani]{mcdropout-gal16}
Yarin Gal and Zoubin Ghahramani.
\newblock Dropout as a bayesian approximation: Representing model uncertainty
  in deep learning.
\newblock In Maria~Florina Balcan and Kilian~Q. Weinberger (eds.),
  \emph{Proceedings of The 33rd International Conference on Machine Learning},
  volume~48 of \emph{Proceedings of Machine Learning Research}, pp.\
  1050--1059, New York, New York, USA, 20--22 Jun 2016. PMLR.
\newblock URL \url{https://proceedings.mlr.press/v48/gal16.html}.

\bibitem[Gao et~al.(2023)Gao, Yen, Yu, and Chen]{gao-etal-2023-enabling}
Tianyu Gao, Howard Yen, Jiatong Yu, and Danqi Chen.
\newblock Enabling large language models to generate text with citations.
\newblock In Houda Bouamor, Juan Pino, and Kalika Bali (eds.),
  \emph{Proceedings of the 2023 Conference on Empirical Methods in Natural
  Language Processing}, pp.\  6465--6488, Singapore, December 2023. Association
  for Computational Linguistics.
\newblock \doi{10.18653/v1/2023.emnlp-main.398}.
\newblock URL \url{https://aclanthology.org/2023.emnlp-main.398/}.

\bibitem[Geva et~al.(2021)Geva, Schuster, Berant, and
  Levy]{geva-etal-2021-transformer}
Mor Geva, Roei Schuster, Jonathan Berant, and Omer Levy.
\newblock Transformer feed-forward layers are key-value memories.
\newblock In Marie-Francine Moens, Xuanjing Huang, Lucia Specia, and Scott
  Wen-tau Yih (eds.), \emph{Proceedings of the 2021 Conference on Empirical
  Methods in Natural Language Processing}, pp.\  5484--5495, Online and Punta
  Cana, Dominican Republic, November 2021. Association for Computational
  Linguistics.
\newblock \doi{10.18653/v1/2021.emnlp-main.446}.
\newblock URL \url{https://aclanthology.org/2021.emnlp-main.446/}.

\bibitem[Guo et~al.(2017)Guo, Pleiss, Sun, and Weinberger]{pmlr-v70-guo17a}
Chuan Guo, Geoff Pleiss, Yu~Sun, and Kilian~Q. Weinberger.
\newblock On calibration of modern neural networks.
\newblock In Doina Precup and Yee~Whye Teh (eds.), \emph{Proceedings of the
  34th International Conference on Machine Learning}, volume~70 of
  \emph{Proceedings of Machine Learning Research}, pp.\  1321--1330. PMLR,
  06--11 Aug 2017.
\newblock URL \url{https://proceedings.mlr.press/v70/guo17a.html}.

\bibitem[Guu et~al.(2020)Guu, Lee, Tung, Pasupat, and Chang]{guu20a-realm}
Kelvin Guu, Kenton Lee, Zora Tung, Panupong Pasupat, and Mingwei Chang.
\newblock Retrieval augmented language model pre-training.
\newblock In Hal~Daumé III and Aarti Singh (eds.), \emph{Proceedings of the
  37th International Conference on Machine Learning}, volume 119 of
  \emph{Proceedings of Machine Learning Research}, pp.\  3929--3938. PMLR,
  13--18 Jul 2020.
\newblock URL \url{https://proceedings.mlr.press/v119/guu20a.html}.

\bibitem[Hagström et~al.(2025)Hagström, Kim, Yu, goo Lee, Johansson, Cho, and
  Augenstein]{cub2025}
Lovisa Hagström, Youna Kim, Haeun Yu, Sang goo Lee, Richard Johansson, Hyunsoo
  Cho, and Isabelle Augenstein.
\newblock Cub: Benchmarking context utilisation techniques for language models,
  2025.
\newblock URL \url{https://arxiv.org/abs/2505.16518}.

\bibitem[Holtzman et~al.(2020)Holtzman, Buys, Du, Forbes, and Choi]{nucleus}
Ari Holtzman, Jan Buys, Li~Du, Maxwell Forbes, and Yejin Choi.
\newblock The curious case of neural text degeneration.
\newblock In \emph{International Conference on Learning Representations}, 2020.
\newblock URL \url{https://openreview.net/forum?id=rygGQyrFvH}.

\bibitem[Hou et~al.(2024)Hou, Liu, Qian, Andreas, Chang, and
  Zhang]{hou2024decomposing}
Bairu Hou, Yujian Liu, Kaizhi Qian, Jacob Andreas, Shiyu Chang, and Yang Zhang.
\newblock Decomposing uncertainty for large language models through input
  clarification ensembling.
\newblock In \emph{Forty-first International Conference on Machine Learning},
  2024.
\newblock URL \url{https://openreview.net/forum?id=byxXa99PtF}.

\bibitem[Izacard \& Grave(2021)Izacard and
  Grave]{izacard-grave-2021-leveraging}
Gautier Izacard and Edouard Grave.
\newblock Leveraging passage retrieval with generative models for open domain
  question answering.
\newblock In Paola Merlo, Jorg Tiedemann, and Reut Tsarfaty (eds.),
  \emph{Proceedings of the 16th Conference of the European Chapter of the
  Association for Computational Linguistics: Main Volume}, pp.\  874--880,
  Online, April 2021. Association for Computational Linguistics.
\newblock \doi{10.18653/v1/2021.eacl-main.74}.
\newblock URL \url{https://aclanthology.org/2021.eacl-main.74}.

\bibitem[Izacard et~al.(2022)Izacard, Caron, Hosseini, Riedel, Bojanowski,
  Joulin, and Grave]{izacard2022unsupervised}
Gautier Izacard, Mathilde Caron, Lucas Hosseini, Sebastian Riedel, Piotr
  Bojanowski, Armand Joulin, and Edouard Grave.
\newblock Unsupervised dense information retrieval with contrastive learning.
\newblock \emph{Transactions on Machine Learning Research}, 2022.
\newblock ISSN 2835-8856.
\newblock URL \url{https://openreview.net/forum?id=jKN1pXi7b0}.

\bibitem[Izacard et~al.(2024)Izacard, Lewis, Lomeli, Hosseini, Petroni, Schick,
  Dwivedi-Yu, Joulin, Riedel, and Grave]{atlas}
Gautier Izacard, Patrick Lewis, Maria Lomeli, Lucas Hosseini, Fabio Petroni,
  Timo Schick, Jane Dwivedi-Yu, Armand Joulin, Sebastian Riedel, and Edouard
  Grave.
\newblock Atlas: few-shot learning with retrieval augmented language models.
\newblock \emph{J. Mach. Learn. Res.}, 24\penalty0 (1), mar 2024.
\newblock ISSN 1532-4435.

\bibitem[Jeong et~al.(2024)Jeong, Baek, Cho, Hwang, and
  Park]{jeong2024adaptiverag}
Soyeong Jeong, Jinheon Baek, Sukmin Cho, Sung~Ju Hwang, and Jong Park.
\newblock Adaptive-rag: Learning to adapt retrieval-augmented large language
  models through question complexity.
\newblock In \emph{NAACL}, 2024.
\newblock URL \url{https://arxiv.org/abs/2403.14403}.

\bibitem[Jiang et~al.(2023)Jiang, Sablayrolles, Mensch, Bamford, Chaplot,
  de~las Casas, Bressand, Lengyel, Lample, Saulnier, Lavaud, Lachaux, Stock,
  Scao, Lavril, Wang, Lacroix, and Sayed]{jiang2023mistral7b}
Albert~Q. Jiang, Alexandre Sablayrolles, Arthur Mensch, Chris Bamford,
  Devendra~Singh Chaplot, Diego de~las Casas, Florian Bressand, Gianna Lengyel,
  Guillaume Lample, Lucile Saulnier, Lélio~Renard Lavaud, Marie-Anne Lachaux,
  Pierre Stock, Teven~Le Scao, Thibaut Lavril, Thomas Wang, Timothée Lacroix,
  and William~El Sayed.
\newblock Mistral 7b, 2023.
\newblock URL \url{https://arxiv.org/abs/2310.06825}.

\bibitem[Jiang et~al.(2021)Jiang, Araki, Ding, and
  Neubig]{10.1162/tacl_a_00407}
Zhengbao Jiang, Jun Araki, Haibo Ding, and Graham Neubig.
\newblock {How Can We Know When Language Models Know? On the Calibration of
  Language Models for Question Answering}.
\newblock \emph{Transactions of the Association for Computational Linguistics},
  9:\penalty0 962--977, 09 2021.
\newblock ISSN 2307-387X.
\newblock \doi{10.1162/tacl_a_00407}.
\newblock URL \url{https://doi.org/10.1162/tacl\_a\_00407}.

\bibitem[Joren et~al.(2025)Joren, Zhang, Ferng, Juan, Taly, and
  Rashtchian]{joren2025sufficientcontextnewlens}
Hailey Joren, Jianyi Zhang, Chun-Sung Ferng, Da-Cheng Juan, Ankur Taly, and
  Cyrus Rashtchian.
\newblock Sufficient context: A new lens on retrieval augmented generation
  systems.
\newblock In \emph{The Thirteenth International Conference on Learning
  Representations}, 2025.
\newblock URL \url{https://openreview.net/pdf?id=Jjr2Odj8DJ}.

\bibitem[Joshi et~al.(2017)Joshi, Choi, Weld, and
  Zettlemoyer]{joshi-etal-2017-triviaqa}
Mandar Joshi, Eunsol Choi, Daniel Weld, and Luke Zettlemoyer.
\newblock {T}rivia{QA}: A large scale distantly supervised challenge dataset
  for reading comprehension.
\newblock In Regina Barzilay and Min-Yen Kan (eds.), \emph{Proceedings of the
  55th Annual Meeting of the Association for Computational Linguistics (Volume
  1: Long Papers)}, pp.\  1601--1611, Vancouver, Canada, July 2017. Association
  for Computational Linguistics.
\newblock \doi{10.18653/v1/P17-1147}.
\newblock URL \url{https://aclanthology.org/P17-1147}.

\bibitem[Kadavath et~al.(2022)Kadavath, Conerly, Askell, Henighan, Drain,
  Perez, Schiefer, Hatfield-Dodds, DasSarma, Tran-Johnson, Johnston, El-Showk,
  Jones, Elhage, Hume, Chen, Bai, Bowman, Fort, Ganguli, Hernandez, Jacobson,
  Kernion, Kravec, Lovitt, Ndousse, Olsson, Ringer, Amodei, Brown, Clark,
  Joseph, Mann, McCandlish, Olah, and Kaplan]{kadavath2022-ptrue}
Saurav Kadavath, Tom Conerly, Amanda Askell, Tom Henighan, Dawn Drain, Ethan
  Perez, Nicholas Schiefer, Zac Hatfield-Dodds, Nova DasSarma, Eli
  Tran-Johnson, Scott Johnston, Sheer El-Showk, Andy Jones, Nelson Elhage,
  Tristan Hume, Anna Chen, Yuntao Bai, Sam Bowman, Stanislav Fort, Deep
  Ganguli, Danny Hernandez, Josh Jacobson, Jackson Kernion, Shauna Kravec,
  Liane Lovitt, Kamal Ndousse, Catherine Olsson, Sam Ringer, Dario Amodei, Tom
  Brown, Jack Clark, Nicholas Joseph, Ben Mann, Sam McCandlish, Chris Olah, and
  Jared Kaplan.
\newblock Language models (mostly) know what they know, 2022.
\newblock URL \url{https://arxiv.org/abs/2207.05221}.

\bibitem[Kamath et~al.(2020)Kamath, Jia, and Liang]{kamath-etal-2020-selective}
Amita Kamath, Robin Jia, and Percy Liang.
\newblock Selective question answering under domain shift.
\newblock In Dan Jurafsky, Joyce Chai, Natalie Schluter, and Joel Tetreault
  (eds.), \emph{Proceedings of the 58th Annual Meeting of the Association for
  Computational Linguistics}, pp.\  5684--5696, Online, July 2020. Association
  for Computational Linguistics.
\newblock \doi{10.18653/v1/2020.acl-main.503}.
\newblock URL \url{https://aclanthology.org/2020.acl-main.503}.

\bibitem[Kasai et~al.(2024)Kasai, Sakaguchi, Takahashi, Le~Bras, Asai, Yu,
  Radev, Smith, Choi, and Inui]{realtimeQA:2024}
Jungo Kasai, Keisuke Sakaguchi, Yoichi Takahashi, Ronan Le~Bras, Akari Asai,
  Xinyan~Velocity Yu, Dragomir Radev, Noah~A. Smith, Yejin Choi, and Kentaro
  Inui.
\newblock Realtime qa: what's the answer right now?
\newblock In \emph{Proceedings of the 37th International Conference on Neural
  Information Processing Systems}, NIPS '23, Red Hook, NY, USA, 2024. Curran
  Associates Inc.

\bibitem[Kuhn et~al.(2023)Kuhn, Gal, and Farquhar]{kuhn2023semantic}
Lorenz Kuhn, Yarin Gal, and Sebastian Farquhar.
\newblock Semantic uncertainty: Linguistic invariances for uncertainty
  estimation in natural language generation.
\newblock In \emph{The Eleventh International Conference on Learning
  Representations}, 2023.
\newblock URL \url{https://openreview.net/forum?id=VD-AYtP0dve}.

\bibitem[Kwiatkowski et~al.(2019)Kwiatkowski, Palomaki, Redfield, Collins,
  Parikh, Alberti, Epstein, Polosukhin, Devlin, Lee, Toutanova, Jones, Kelcey,
  Chang, Dai, Uszkoreit, Le, and Petrov]{kwiatkowski-etal-2019-natural}
Tom Kwiatkowski, Jennimaria Palomaki, Olivia Redfield, Michael Collins, Ankur
  Parikh, Chris Alberti, Danielle Epstein, Illia Polosukhin, Jacob Devlin,
  Kenton Lee, Kristina Toutanova, Llion Jones, Matthew Kelcey, Ming-Wei Chang,
  Andrew~M. Dai, Jakob Uszkoreit, Quoc Le, and Slav Petrov.
\newblock Natural questions: A benchmark for question answering research.
\newblock \emph{Transactions of the Association for Computational Linguistics},
  7:\penalty0 452--466, 2019.
\newblock \doi{10.1162/tacl_a_00276}.
\newblock URL \url{https://aclanthology.org/Q19-1026}.

\bibitem[Kwon et~al.(2023)Kwon, Li, Zhuang, Sheng, Zheng, Yu, Gonzalez, Zhang,
  and Stoica]{vllm}
Woosuk Kwon, Zhuohan Li, Siyuan Zhuang, Ying Sheng, Lianmin Zheng, Cody~Hao Yu,
  Joseph Gonzalez, Hao Zhang, and Ion Stoica.
\newblock Efficient memory management for large language model serving with
  pagedattention.
\newblock In \emph{Proceedings of the 29th Symposium on Operating Systems
  Principles}, SOSP '23, pp.\  611–626, New York, NY, USA, 2023. Association
  for Computing Machinery.
\newblock ISBN 9798400702297.
\newblock \doi{10.1145/3600006.3613165}.
\newblock URL \url{https://doi.org/10.1145/3600006.3613165}.

\bibitem[Lahlou et~al.(2023)Lahlou, Jain, Nekoei, Butoi, Bertin, Rector-Brooks,
  Korablyov, and Bengio]{lahlou2023deup}
Salem Lahlou, Moksh Jain, Hadi Nekoei, Victor~I Butoi, Paul Bertin, Jarrid
  Rector-Brooks, Maksym Korablyov, and Yoshua Bengio.
\newblock {DEUP}: Direct epistemic uncertainty prediction.
\newblock \emph{Transactions on Machine Learning Research}, 2023.
\newblock ISSN 2835-8856.
\newblock URL \url{https://openreview.net/forum?id=eGLdVRvvfQ}.
\newblock Expert Certification.

\bibitem[Lakshminarayanan et~al.(2017)Lakshminarayanan, Pritzel, and
  Blundell]{NIPS2017_deep:ensembles}
Balaji Lakshminarayanan, Alexander Pritzel, and Charles Blundell.
\newblock Simple and scalable predictive uncertainty estimation using deep
  ensembles.
\newblock In I.~Guyon, U.~Von Luxburg, S.~Bengio, H.~Wallach, R.~Fergus,
  S.~Vishwanathan, and R.~Garnett (eds.), \emph{Advances in Neural Information
  Processing Systems}, volume~30. Curran Associates, Inc., 2017.
\newblock URL
  \url{https://proceedings.neurips.cc/paper_files/paper/2017/file/9ef2ed4b7fd2c810847ffa5fa85bce38-Paper.pdf}.

\bibitem[Lan et~al.(2020)Lan, Chen, Goodman, Gimpel, Sharma, and
  Soricut]{Lan2020ALBERT}
Zhenzhong Lan, Mingda Chen, Sebastian Goodman, Kevin Gimpel, Piyush Sharma, and
  Radu Soricut.
\newblock Albert: A lite bert for self-supervised learning of language
  representations.
\newblock In \emph{International Conference on Learning Representations}, 2020.
\newblock URL \url{https://openreview.net/forum?id=H1eA7AEtvS}.

\bibitem[Lee et~al.(2019)Lee, Chang, and Toutanova]{lee-etal-2019-latent}
Kenton Lee, Ming-Wei Chang, and Kristina Toutanova.
\newblock Latent retrieval for weakly supervised open domain question
  answering.
\newblock In Anna Korhonen, David Traum, and Llu{\'\i}s M{\`a}rquez (eds.),
  \emph{Proceedings of the 57th Annual Meeting of the Association for
  Computational Linguistics}, pp.\  6086--6096, Florence, Italy, July 2019.
  Association for Computational Linguistics.
\newblock \doi{10.18653/v1/P19-1612}.
\newblock URL \url{https://aclanthology.org/P19-1612}.

\bibitem[Lewis et~al.(2020)Lewis, Perez, Piktus, Petroni, Karpukhin, Goyal,
  K\"{u}ttler, Lewis, Yih, Rockt\"{a}schel, Riedel, and Kiela]{rag-lewis}
Patrick Lewis, Ethan Perez, Aleksandra Piktus, Fabio Petroni, Vladimir
  Karpukhin, Naman Goyal, Heinrich K\"{u}ttler, Mike Lewis, Wen-tau Yih, Tim
  Rockt\"{a}schel, Sebastian Riedel, and Douwe Kiela.
\newblock Retrieval-augmented generation for knowledge-intensive nlp tasks.
\newblock In \emph{Proceedings of the 34th International Conference on Neural
  Information Processing Systems}, NIPS '20, Red Hook, NY, USA, 2020. Curran
  Associates Inc.
\newblock ISBN 9781713829546.

\bibitem[Li \& Zhang(2024)Li and Zhang]{li-zhang-2024-planning}
Kunze Li and Yu~Zhang.
\newblock Planning first, question second: An {LLM}-guided method for
  controllable question generation.
\newblock In Lun-Wei Ku, Andre Martins, and Vivek Srikumar (eds.),
  \emph{Findings of the Association for Computational Linguistics: ACL 2024},
  pp.\  4715--4729, Bangkok, Thailand, August 2024. Association for
  Computational Linguistics.
\newblock \doi{10.18653/v1/2024.findings-acl.280}.
\newblock URL \url{https://aclanthology.org/2024.findings-acl.280/}.

\bibitem[Lin et~al.(2024)Lin, Tang, Tang, Yang, Chen, Wang, Xiao, Dang, Gan,
  and Han]{awq}
Ji~Lin, Jiaming Tang, Haotian Tang, Shang Yang, Wei-Ming Chen, Wei-Chen Wang,
  Guangxuan Xiao, Xingyu Dang, Chuang Gan, and Song Han.
\newblock Awq: Activation-aware weight quantization for on-device llm
  compression and acceleration.
\newblock In P.~Gibbons, G.~Pekhimenko, and C.~De Sa (eds.), \emph{Proceedings
  of Machine Learning and Systems}, volume~6, pp.\  87--100, 2024.
\newblock URL
  \url{https://proceedings.mlsys.org/paper_files/paper/2024/file/42a452cbafa9dd64e9ba4aa95cc1ef21-Paper-Conference.pdf}.

\bibitem[Lin et~al.(2022)Lin, Hilton, and Evans]{lin2022teaching}
Stephanie Lin, Jacob Hilton, and Owain Evans.
\newblock Teaching models to express their uncertainty in words.
\newblock \emph{Transactions on Machine Learning Research}, 2022.
\newblock ISSN 2835-8856.
\newblock URL \url{https://openreview.net/forum?id=8s8K2UZGTZ}.

\bibitem[Lin et~al.(2025)Lin, Huang, Zhang, Zhou, and
  Chen]{Lin_Huang_Zhang_Zhou_Chen_2025}
Xin Lin, Zhenya Huang, Zhiqiang Zhang, Jun Zhou, and Enhong Chen.
\newblock Explore what llm does not know in complex question answering.
\newblock \emph{Proceedings of the AAAI Conference on Artificial Intelligence},
  39\penalty0 (23):\penalty0 24585--24594, Apr. 2025.
\newblock \doi{10.1609/aaai.v39i23.34638}.
\newblock URL \url{https://ojs.aaai.org/index.php/AAAI/article/view/34638}.

\bibitem[Liu et~al.(2024{\natexlab{a}})Liu, Wang, Yuan, Chen, and Peng]{refunq}
Genglin Liu, Xingyao Wang, Lifan Yuan, Yangyi Chen, and Hao Peng.
\newblock Examining llms' uncertainty expression towards questions outside
  parametric knowledge, 2024{\natexlab{a}}.

\bibitem[Liu et~al.(2024{\natexlab{b}})Liu, Lin, Hewitt, Paranjape, Bevilacqua,
  Petroni, and Liang]{liu-etal-2024-lost}
Nelson~F. Liu, Kevin Lin, John Hewitt, Ashwin Paranjape, Michele Bevilacqua,
  Fabio Petroni, and Percy Liang.
\newblock Lost in the middle: How language models use long contexts.
\newblock \emph{Transactions of the Association for Computational Linguistics},
  12:\penalty0 157--173, 2024{\natexlab{b}}.
\newblock \doi{10.1162/tacl_a_00638}.
\newblock URL \url{https://aclanthology.org/2024.tacl-1.9}.

\bibitem[Longpre et~al.(2021)Longpre, Perisetla, Chen, Ramesh, DuBois, and
  Singh]{longpre-etal-2021-entity}
Shayne Longpre, Kartik Perisetla, Anthony Chen, Nikhil Ramesh, Chris DuBois,
  and Sameer Singh.
\newblock Entity-based knowledge conflicts in question answering.
\newblock In Marie-Francine Moens, Xuanjing Huang, Lucia Specia, and Scott
  Wen-tau Yih (eds.), \emph{Proceedings of the 2021 Conference on Empirical
  Methods in Natural Language Processing}, pp.\  7052--7063, Online and Punta
  Cana, Dominican Republic, November 2021. Association for Computational
  Linguistics.
\newblock \doi{10.18653/v1/2021.emnlp-main.565}.
\newblock URL \url{https://aclanthology.org/2021.emnlp-main.565/}.

\bibitem[Ma et~al.(2024)Ma, Wang, Yang, Wei, and Lin]{rankLlama}
Xueguang Ma, Liang Wang, Nan Yang, Furu Wei, and Jimmy Lin.
\newblock Fine-tuning llama for multi-stage text retrieval.
\newblock In \emph{Proceedings of the 47th International ACM SIGIR Conference
  on Research and Development in Information Retrieval}, SIGIR '24, pp.\
  2421–2425, New York, NY, USA, 2024. Association for Computing Machinery.
\newblock ISBN 9798400704314.
\newblock \doi{10.1145/3626772.3657951}.
\newblock URL \url{https://doi.org/10.1145/3626772.3657951}.

\bibitem[Mallen et~al.(2023)Mallen, Asai, Zhong, Das, Khashabi, and
  Hajishirzi]{mallen-etal-2023-popQA}
Alex Mallen, Akari Asai, Victor Zhong, Rajarshi Das, Daniel Khashabi, and
  Hannaneh Hajishirzi.
\newblock When not to trust language models: Investigating effectiveness of
  parametric and non-parametric memories.
\newblock In Anna Rogers, Jordan Boyd-Graber, and Naoaki Okazaki (eds.),
  \emph{Proceedings of the 61st Annual Meeting of the Association for
  Computational Linguistics (Volume 1: Long Papers)}, pp.\  9802--9822,
  Toronto, Canada, July 2023. Association for Computational Linguistics.
\newblock \doi{10.18653/v1/2023.acl-long.546}.
\newblock URL \url{https://aclanthology.org/2023.acl-long.546}.

\bibitem[Mielke et~al.(2022)Mielke, Szlam, Dinan, and
  Boureau]{mielke-etal-2022-reducing}
Sabrina~J. Mielke, Arthur Szlam, Emily Dinan, and Y-Lan Boureau.
\newblock Reducing conversational agents{'} overconfidence through linguistic
  calibration.
\newblock \emph{Transactions of the Association for Computational Linguistics},
  10:\penalty0 857--872, 2022.
\newblock \doi{10.1162/tacl_a_00494}.
\newblock URL \url{https://aclanthology.org/2022.tacl-1.50}.

\bibitem[Min et~al.(2023)Min, Krishna, Lyu, Lewis, Yih, Koh, Iyyer,
  Zettlemoyer, and Hajishirzi]{min-etal-2023-factscore}
Sewon Min, Kalpesh Krishna, Xinxi Lyu, Mike Lewis, Wen-tau Yih, Pang Koh, Mohit
  Iyyer, Luke Zettlemoyer, and Hannaneh Hajishirzi.
\newblock {FA}ct{S}core: Fine-grained atomic evaluation of factual precision in
  long form text generation.
\newblock In Houda Bouamor, Juan Pino, and Kalika Bali (eds.),
  \emph{Proceedings of the 2023 Conference on Empirical Methods in Natural
  Language Processing}, pp.\  12076--12100, Singapore, December 2023.
  Association for Computational Linguistics.
\newblock \doi{10.18653/v1/2023.emnlp-main.741}.
\newblock URL \url{https://aclanthology.org/2023.emnlp-main.741/}.

\bibitem[Nogueira et~al.(2019)Nogueira, Yang, Cho, and
  Lin]{Nogueira2019MultiStageDR}
Rodrigo Nogueira, Wei Yang, Kyunghyun Cho, and Jimmy~J. Lin.
\newblock Multi-stage document ranking with bert.
\newblock \emph{ArXiv}, abs/1910.14424, 2019.
\newblock URL \url{https://api.semanticscholar.org/CorpusID:207758365}.

\bibitem[Ouyang et~al.(2024)Ouyang, Wu, Jiang, Almeida, Wainwright, Mishkin,
  Zhang, Agarwal, Slama, Ray, Schulman, Hilton, Kelton, Miller, Simens, Askell,
  Welinder, Christiano, Leike, and Lowe]{instructGPT}
Long Ouyang, Jeff Wu, Xu~Jiang, Diogo Almeida, Carroll~L. Wainwright, Pamela
  Mishkin, Chong Zhang, Sandhini Agarwal, Katarina Slama, Alex Ray, John
  Schulman, Jacob Hilton, Fraser Kelton, Luke Miller, Maddie Simens, Amanda
  Askell, Peter Welinder, Paul Christiano, Jan Leike, and Ryan Lowe.
\newblock Training language models to follow instructions with human feedback.
\newblock In \emph{Proceedings of the 36th International Conference on Neural
  Information Processing Systems}, NIPS '22, Red Hook, NY, USA, 2024. Curran
  Associates Inc.
\newblock ISBN 9781713871088.

\bibitem[Pal et~al.(2022)Pal, Umapathi, and
  Sankarasubbu]{pmlr-v174-pal22a-medmcqa}
Ankit Pal, Logesh~Kumar Umapathi, and Malaikannan Sankarasubbu.
\newblock Medmcqa: A large-scale multi-subject multi-choice dataset for medical
  domain question answering.
\newblock In Gerardo Flores, George~H Chen, Tom Pollard, Joyce~C Ho, and
  Tristan Naumann (eds.), \emph{Proceedings of the Conference on Health,
  Inference, and Learning}, volume 174 of \emph{Proceedings of Machine Learning
  Research}, pp.\  248--260. PMLR, 07--08 Apr 2022.
\newblock URL \url{https://proceedings.mlr.press/v174/pal22a.html}.

\bibitem[Park et~al.(2024)Park, Choi, Kim, and Lee]{park-etal-2024-enhancing}
SeongIl Park, Seungwoo Choi, Nahyun Kim, and Jay-Yoon Lee.
\newblock Enhancing robustness of retrieval-augmented language models with
  in-context learning.
\newblock In Wenhao Yu, Weijia Shi, Michihiro Yasunaga, Meng Jiang, Chenguang
  Zhu, Hannaneh Hajishirzi, Luke Zettlemoyer, and Zhihan Zhang (eds.),
  \emph{Proceedings of the 3rd Workshop on Knowledge Augmented Methods for
  NLP}, pp.\  93--102, Bangkok, Thailand, August 2024. Association for
  Computational Linguistics.
\newblock \doi{10.18653/v1/2024.knowledgenlp-1.7}.
\newblock URL \url{https://aclanthology.org/2024.knowledgenlp-1.7}.

\bibitem[Rajpurkar et~al.(2016)Rajpurkar, Zhang, Lopyrev, and
  Liang]{rajpurkar-etal-2016-squad}
Pranav Rajpurkar, Jian Zhang, Konstantin Lopyrev, and Percy Liang.
\newblock {SQ}u{AD}: 100,000+ questions for machine comprehension of text.
\newblock In Jian Su, Kevin Duh, and Xavier Carreras (eds.), \emph{Proceedings
  of the 2016 Conference on Empirical Methods in Natural Language Processing},
  pp.\  2383--2392, Austin, Texas, November 2016. Association for Computational
  Linguistics.
\newblock \doi{10.18653/v1/D16-1264}.
\newblock URL \url{https://aclanthology.org/D16-1264}.

\bibitem[Riviere et~al.(2024)Riviere, Pathak, Sessa, Hardin, Bhupatiraju,
  Hussenot, Mesnard, Shahriari, Ram'e, Ferret, Liu, Tafti, Friesen, Casbon,
  Ramos, Kumar, Lan, Jerome, Tsitsulin, Vieillard, Stańczyk, Girgin, Momchev,
  Hoffman, Thakoor, Grill, Neyshabur, Walton, Severyn, Parrish, Ahmad,
  Hutchison, Abdagic, Carl, Shen, Brock, Coenen, Laforge, Paterson, Bastian,
  Piot, Wu, Royal, Chen, Kumar, Perry, Welty, Choquette-Choo, Sinopalnikov,
  Weinberger, Vijaykumar, Rogozi'nska, Herbison, Bandy, Wang, Noland, Moreira,
  Senter, Eltyshev, Visin, Rasskin, Wei, Cameron, Martins, Hashemi,
  Klimczak-Pluci'nska, Batra, Dhand, Nardini, Mein, Zhou, Svensson, Stanway,
  Chan, Zhou, Carrasqueira, Iljazi, Becker, Fernandez, van Amersfoort, Gordon,
  Lipschultz, Newlan, Ji, Mohamed, Badola, Black, Millican, McDonell, Nguyen,
  Sodhia, Greene, Sjoesund, Usui, Sifre, Heuermann, Lago, McNealus, Soares,
  Kilpatrick, Dixon, Martins, Reid, Singh, Iverson, Gorner, Velloso, Wirth,
  Davidow, Miller, Rahtz, Watson, Risdal, Kazemi, Moynihan, Zhang, Kahng, Park,
  Rahman, Khatwani, Dao, Bardoliwalla, Devanathan, Dumai, Chauhan, Wahltinez,
  Botarda, Barnes, Barham, Michel, Jin, Georgiev, Culliton, Kuppala, Comanescu,
  Merhej, Jana, Rokni, Agarwal, Mullins, Saadat, Carthy, Perrin, Arnold,
  Krause, Dai, Garg, Sheth, Ronstrom, Chan, Jordan, Yu, Eccles, Hennigan,
  Kocisk{\'y}, Doshi, Jain, Yadav, Meshram, Dharmadhikari, Barkley, Wei, Ye,
  Han, Kwon, Xu, Shen, Gong, Wei, Cotruta, Kirk, Rao, Giang, Peran, Warkentin,
  Collins, Barral, Ghahramani, Hadsell, Sculley, Banks, Dragan, Petrov,
  Vinyals, Dean, Hassabis, Kavukcuoglu, Farabet, Buchatskaya, Borgeaud, Fiedel,
  Joulin, Kenealy, Dadashi, and Andreev]{Riviere2024Gemma2I}
Gemma Team~Morgane Riviere, Shreya Pathak, Pier~Giuseppe Sessa, Cassidy Hardin,
  Surya Bhupatiraju, L'eonard Hussenot, Thomas Mesnard, Bobak Shahriari,
  Alexandre Ram'e, Johan Ferret, Peter Liu, Pouya~Dehghani Tafti, Abe Friesen,
  Michelle Casbon, Sabela Ramos, Ravin Kumar, Charline~Le Lan, Sammy Jerome,
  Anton Tsitsulin, Nino Vieillard, Piotr Stańczyk, Sertan Girgin, Nikola
  Momchev, Matt Hoffman, Shantanu Thakoor, Jean-Bastien Grill, Behnam
  Neyshabur, Alanna Walton, Aliaksei Severyn, Alicia Parrish, Aliya Ahmad,
  Allen Hutchison, Alvin Abdagic, Amanda Carl, Amy Shen, Andy Brock, Andy
  Coenen, Anthony Laforge, Antonia Paterson, Ben Bastian, Bilal Piot, Boxi Wu,
  Brandon Royal, Charlie Chen, Chintu Kumar, Chris Perry, Christoper~A. Welty,
  Christopher~A. Choquette-Choo, Danila Sinopalnikov, David Weinberger, Dimple
  Vijaykumar, Dominika Rogozi'nska, D.~Herbison, Elisa Bandy, Emma Wang, Eric
  Noland, Erica Moreira, Evan Senter, Evgenii Eltyshev, Francesco Visin,
  Gabriel Rasskin, Gary Wei, Glenn Cameron, Gus Martins, Hadi Hashemi, Hanna
  Klimczak-Pluci'nska, Harleen Batra, Harsh Dhand, Ivan Nardini, Jacinda Mein,
  Jack Zhou, James Svensson, Jeff Stanway, Jetha Chan, Jin Zhou, Joana
  Carrasqueira, Joana Iljazi, Jocelyn Becker, Joe Fernandez, Joost~R. van
  Amersfoort, Josh Gordon, Josh Lipschultz, Joshua Newlan, Junsong Ji, Kareem
  Mohamed, Kartikeya Badola, Kat Black, Katie Millican, Keelin McDonell, Kelvin
  Nguyen, Kiranbir Sodhia, Kish Greene, Lars~Lowe Sjoesund, Lauren Usui,
  L.~Sifre, L.~Heuermann, Leticia Lago, Lilly McNealus, Livio~Baldini Soares,
  Logan Kilpatrick, Lucas Dixon, Luciano Martins, Machel Reid, Manvinder Singh,
  Mark Iverson, Martin Gorner, Mat Velloso, Mateo Wirth, Matt Davidow, Matt
  Miller, Matthew Rahtz, Matthew Watson, Meg Risdal, Mehran Kazemi, Michael
  Moynihan, Ming Zhang, Minsuk Kahng, Minwoo Park, Mofi Rahman, Mohit Khatwani,
  Natalie Dao, Nenshad Bardoliwalla, Nesh Devanathan, Neta Dumai, Nilay
  Chauhan, Oscar Wahltinez, Pankil Botarda, Parker Barnes, Paul Barham, Paul
  Michel, Pengchong Jin, Petko Georgiev, Phil Culliton, Pradeep Kuppala, Ramona
  Comanescu, Ramona Merhej, Reena Jana, Reza Rokni, Rishabh Agarwal, Ryan
  Mullins, Samaneh Saadat, S.~Mc Carthy, Sarah Perrin, S'ebastien Arnold,
  Sebastian Krause, Shengyang Dai, Shruti Garg, Shruti Sheth, Sue Ronstrom,
  Susan Chan, Timothy Jordan, Ting Yu, Tom Eccles, Tom Hennigan, Tom{\'a}s
  Kocisk{\'y}, Tulsee Doshi, Vihan Jain, Vikas Yadav, Vilobh Meshram, Vishal
  Dharmadhikari, Warren Barkley, Wei Wei, Wenming Ye, Woohyun Han, Woosuk Kwon,
  Xiang Xu, Zhe Shen, Zhitao Gong, Zichuan Wei, Victor Cotruta, Phoebe Kirk,
  Anand Rao, Minh Giang, Ludovic Peran, Tris~Brian Warkentin, Eli Collins,
  Joelle Barral, Zoubin Ghahramani, Raia Hadsell, D.~Sculley, Jeanine Banks,
  Anca Dragan, Slav Petrov, Oriol Vinyals, Jeffrey Dean, Demis Hassabis, Koray
  Kavukcuoglu, Cl'ement Farabet, Elena Buchatskaya, Sebastian Borgeaud, Noah
  Fiedel, Armand Joulin, Kathleen Kenealy, Robert Dadashi, and Alek Andreev.
\newblock Gemma 2: Improving open language models at a practical size.
\newblock \emph{ArXiv}, abs/2408.00118, 2024.
\newblock URL \url{https://api.semanticscholar.org/CorpusID:270843326}.

\bibitem[Rodriguez \& Boyd-Graber(2021)Rodriguez and
  Boyd-Graber]{rodriguez-boyd-graber-2021-evaluation}
Pedro Rodriguez and Jordan Boyd-Graber.
\newblock Evaluation paradigms in question answering.
\newblock In Marie-Francine Moens, Xuanjing Huang, Lucia Specia, and Scott
  Wen-tau Yih (eds.), \emph{Proceedings of the 2021 Conference on Empirical
  Methods in Natural Language Processing}, pp.\  9630--9642, Online and Punta
  Cana, Dominican Republic, November 2021. Association for Computational
  Linguistics.
\newblock \doi{10.18653/v1/2021.emnlp-main.758}.
\newblock URL \url{https://aclanthology.org/2021.emnlp-main.758/}.

\bibitem[Schuster et~al.(2021)Schuster, Fisch, and
  Barzilay]{schuster-etal-2021-get}
Tal Schuster, Adam Fisch, and Regina Barzilay.
\newblock Get your vitamin {C}! robust fact verification with contrastive
  evidence.
\newblock In Kristina Toutanova, Anna Rumshisky, Luke Zettlemoyer, Dilek
  Hakkani-Tur, Iz~Beltagy, Steven Bethard, Ryan Cotterell, Tanmoy Chakraborty,
  and Yichao Zhou (eds.), \emph{Proceedings of the 2021 Conference of the North
  American Chapter of the Association for Computational Linguistics: Human
  Language Technologies}, pp.\  624--643, Online, June 2021. Association for
  Computational Linguistics.
\newblock \doi{10.18653/v1/2021.naacl-main.52}.
\newblock URL \url{https://aclanthology.org/2021.naacl-main.52}.

\bibitem[Sciavolino et~al.(2021)Sciavolino, Zhong, Lee, and
  Chen]{sciavolino-etal-2021-simple}
Christopher Sciavolino, Zexuan Zhong, Jinhyuk Lee, and Danqi Chen.
\newblock Simple entity-centric questions challenge dense retrievers.
\newblock In Marie-Francine Moens, Xuanjing Huang, Lucia Specia, and Scott
  Wen-tau Yih (eds.), \emph{Proceedings of the 2021 Conference on Empirical
  Methods in Natural Language Processing}, pp.\  6138--6148, Online and Punta
  Cana, Dominican Republic, November 2021. Association for Computational
  Linguistics.
\newblock \doi{10.18653/v1/2021.emnlp-main.496}.
\newblock URL \url{https://aclanthology.org/2021.emnlp-main.496}.

\bibitem[Sculley(2010)]{combined:sculley}
D.~Sculley.
\newblock Combined regression and ranking.
\newblock In \emph{Proceedings of the 16th ACM SIGKDD International Conference
  on Knowledge Discovery and Data Mining}, KDD '10, pp.\  979–988, New York,
  NY, USA, 2010. Association for Computing Machinery.
\newblock ISBN 9781450300551.
\newblock \doi{10.1145/1835804.1835928}.
\newblock URL \url{https://doi.org/10.1145/1835804.1835928}.

\bibitem[Shorinwa et~al.(2025)Shorinwa, Mei, Lidard, Ren, and
  Majumdar]{10.1145/3744238}
Ola Shorinwa, Zhiting Mei, Justin Lidard, Allen~Z. Ren, and Anirudha Majumdar.
\newblock A survey on uncertainty quantification of large language models:
  Taxonomy, open research challenges, and future directions.
\newblock \emph{ACM Comput. Surv.}, June 2025.
\newblock ISSN 0360-0300.
\newblock \doi{10.1145/3744238}.
\newblock URL \url{https://doi.org/10.1145/3744238}.
\newblock Just Accepted.

\bibitem[Shuster et~al.(2021)Shuster, Poff, Chen, Kiela, and
  Weston]{shuster-etal-2021-retrieval-augmentation}
Kurt Shuster, Spencer Poff, Moya Chen, Douwe Kiela, and Jason Weston.
\newblock Retrieval augmentation reduces hallucination in conversation.
\newblock In Marie-Francine Moens, Xuanjing Huang, Lucia Specia, and Scott
  Wen-tau Yih (eds.), \emph{Findings of the Association for Computational
  Linguistics: EMNLP 2021}, pp.\  3784--3803, Punta Cana, Dominican Republic,
  November 2021. Association for Computational Linguistics.
\newblock \doi{10.18653/v1/2021.findings-emnlp.320}.
\newblock URL \url{https://aclanthology.org/2021.findings-emnlp.320/}.

\bibitem[Simhi et~al.(2025)Simhi, Itzhak, Barez, Stanovsky, and
  Belinkov]{simhi2025trustmeimwrong}
Adi Simhi, Itay Itzhak, Fazl Barez, Gabriel Stanovsky, and Yonatan Belinkov.
\newblock Trust me, i'm wrong: High-certainty hallucinations in llms, 2025.
\newblock URL \url{https://arxiv.org/abs/2502.12964}.

\bibitem[Simpson et~al.(2020)Simpson, Gao, and Gurevych]{text:ranking:bo}
Edwin Simpson, Yang Gao, and Iryna Gurevych.
\newblock {Interactive Text Ranking with Bayesian Optimization: A Case Study on
  Community QA and Summarization}.
\newblock \emph{Transactions of the Association for Computational Linguistics},
  8:\penalty0 759--775, 12 2020.
\newblock ISSN 2307-387X.
\newblock \doi{10.1162/tacl_a_00344}.
\newblock URL \url{https://doi.org/10.1162/tacl\_a\_00344}.

\bibitem[Soudani et~al.(2025)Soudani, Kanoulas, and
  Hasibi]{soudani-etal-2025-uncertainty}
Heydar Soudani, Evangelos Kanoulas, and Faegheh Hasibi.
\newblock Why uncertainty estimation methods fall short in {RAG}: An axiomatic
  analysis.
\newblock In Wanxiang Che, Joyce Nabende, Ekaterina Shutova, and Mohammad~Taher
  Pilehvar (eds.), \emph{Findings of the Association for Computational
  Linguistics: ACL 2025}, pp.\  16596--16616, Vienna, Austria, July 2025.
  Association for Computational Linguistics.
\newblock ISBN 979-8-89176-256-5.
\newblock \doi{10.18653/v1/2025.findings-acl.852}.
\newblock URL \url{https://aclanthology.org/2025.findings-acl.852/}.

\bibitem[Stelmakh et~al.(2022)Stelmakh, Luan, Dhingra, and
  Chang]{stelmakh-etal-2022-asqa}
Ivan Stelmakh, Yi~Luan, Bhuwan Dhingra, and Ming-Wei Chang.
\newblock {ASQA}: Factoid questions meet long-form answers.
\newblock In Yoav Goldberg, Zornitsa Kozareva, and Yue Zhang (eds.),
  \emph{Proceedings of the 2022 Conference on Empirical Methods in Natural
  Language Processing}, pp.\  8273--8288, Abu Dhabi, United Arab Emirates,
  December 2022. Association for Computational Linguistics.
\newblock \doi{10.18653/v1/2022.emnlp-main.566}.
\newblock URL \url{https://aclanthology.org/2022.emnlp-main.566/}.

\bibitem[Sun et~al.(2024)Sun, Xu, Zha, Liu, and Dong]{sun-etal-2024-head}
Kai Sun, Yifan Xu, Hanwen Zha, Yue Liu, and Xin~Luna Dong.
\newblock Head-to-tail: How knowledgeable are large language models ({LLM}s)?
  {A}.{K}.{A}. will {LLM}s replace knowledge graphs?
\newblock In Kevin Duh, Helena Gomez, and Steven Bethard (eds.),
  \emph{Proceedings of the 2024 Conference of the North American Chapter of the
  Association for Computational Linguistics: Human Language Technologies
  (Volume 1: Long Papers)}, pp.\  311--325, Mexico City, Mexico, June 2024.
  Association for Computational Linguistics.
\newblock \doi{10.18653/v1/2024.naacl-long.18}.
\newblock URL \url{https://aclanthology.org/2024.naacl-long.18}.

\bibitem[Sun et~al.(2025)Sun, Zang, Zheng, Xu, Zhang, Yu, Song, and
  Li]{sun2025redeep}
ZhongXiang Sun, Xiaoxue Zang, Kai Zheng, Jun Xu, Xiao Zhang, Weijie Yu, Yang
  Song, and Han Li.
\newblock {ReDe}{EP}: Detecting hallucination in retrieval-augmented generation
  via mechanistic interpretability.
\newblock In \emph{The Thirteenth International Conference on Learning
  Representations}, 2025.
\newblock URL \url{https://openreview.net/forum?id=ztzZDzgfrh}.

\bibitem[Sung et~al.(2025)Sung, Fleisig, Hou, Upadhyay, and
  Boyd-Graber]{sung-etal-2025-grace}
Yoo~Yeon Sung, Eve Fleisig, Yu~Hou, Ishan Upadhyay, and Jordan~Lee Boyd-Graber.
\newblock {GRACE}: A granular benchmark for evaluating model calibration
  against human calibration.
\newblock In Wanxiang Che, Joyce Nabende, Ekaterina Shutova, and Mohammad~Taher
  Pilehvar (eds.), \emph{Proceedings of the 63rd Annual Meeting of the
  Association for Computational Linguistics (Volume 1: Long Papers)}, pp.\
  19586--19587, Vienna, Austria, July 2025. Association for Computational
  Linguistics.
\newblock ISBN 979-8-89176-251-0.
\newblock \doi{10.18653/v1/2025.acl-long.962}.
\newblock URL \url{https://aclanthology.org/2025.acl-long.962/}.

\bibitem[Takayama \& Arase(2019)Takayama and
  Arase]{takayama-arase-2019-relevant}
Junya Takayama and Yuki Arase.
\newblock Relevant and informative response generation using pointwise mutual
  information.
\newblock In Yun-Nung Chen, Tania Bedrax-Weiss, Dilek Hakkani-Tur, Anuj Kumar,
  Mike Lewis, Thang-Minh Luong, Pei-Hao Su, and Tsung-Hsien Wen (eds.),
  \emph{Proceedings of the First Workshop on NLP for Conversational AI}, pp.\
  133--138, Florence, Italy, August 2019. Association for Computational
  Linguistics.
\newblock \doi{10.18653/v1/W19-4115}.
\newblock URL \url{https://aclanthology.org/W19-4115}.

\bibitem[Tian et~al.(2023)Tian, Mitchell, Zhou, Sharma, Rafailov, Yao, Finn,
  and Manning]{tian-etal-2023-just}
Katherine Tian, Eric Mitchell, Allan Zhou, Archit Sharma, Rafael Rafailov,
  Huaxiu Yao, Chelsea Finn, and Christopher Manning.
\newblock Just ask for calibration: Strategies for eliciting calibrated
  confidence scores from language models fine-tuned with human feedback.
\newblock In Houda Bouamor, Juan Pino, and Kalika Bali (eds.),
  \emph{Proceedings of the 2023 Conference on Empirical Methods in Natural
  Language Processing}, pp.\  5433--5442, Singapore, December 2023. Association
  for Computational Linguistics.
\newblock \doi{10.18653/v1/2023.emnlp-main.330}.
\newblock URL \url{https://aclanthology.org/2023.emnlp-main.330}.

\bibitem[Trivedi et~al.(2023)Trivedi, Balasubramanian, Khot, and
  Sabharwal]{trivedi-etal-2023-interleaving}
Harsh Trivedi, Niranjan Balasubramanian, Tushar Khot, and Ashish Sabharwal.
\newblock Interleaving retrieval with chain-of-thought reasoning for
  knowledge-intensive multi-step questions.
\newblock In Anna Rogers, Jordan Boyd-Graber, and Naoaki Okazaki (eds.),
  \emph{Proceedings of the 61st Annual Meeting of the Association for
  Computational Linguistics (Volume 1: Long Papers)}, pp.\  10014--10037,
  Toronto, Canada, July 2023. Association for Computational Linguistics.
\newblock \doi{10.18653/v1/2023.acl-long.557}.
\newblock URL \url{https://aclanthology.org/2023.acl-long.557/}.

\bibitem[Van~Amersfoort et~al.(2020)Van~Amersfoort, Smith, Teh, and
  Gal]{pmlr-v119-van-amersfoort20a}
Joost Van~Amersfoort, Lewis Smith, Yee~Whye Teh, and Yarin Gal.
\newblock Uncertainty estimation using a single deep deterministic neural
  network.
\newblock In Hal~Daumé III and Aarti Singh (eds.), \emph{Proceedings of the
  37th International Conference on Machine Learning}, volume 119 of
  \emph{Proceedings of Machine Learning Research}, pp.\  9690--9700. PMLR,
  13--18 Jul 2020.
\newblock URL \url{https://proceedings.mlr.press/v119/van-amersfoort20a.html}.

\bibitem[Vu et~al.(2024)Vu, Iyyer, Wang, Constant, Wei, Wei, Tar, Sung, Zhou,
  Le, and Luong]{vu-etal-2024-freshllms}
Tu~Vu, Mohit Iyyer, Xuezhi Wang, Noah Constant, Jerry Wei, Jason Wei, Chris
  Tar, Yun-Hsuan Sung, Denny Zhou, Quoc Le, and Thang Luong.
\newblock {F}resh{LLM}s: Refreshing large language models with search engine
  augmentation.
\newblock In Lun-Wei Ku, Andre Martins, and Vivek Srikumar (eds.),
  \emph{Findings of the Association for Computational Linguistics: ACL 2024},
  pp.\  13697--13720, Bangkok, Thailand, August 2024. Association for
  Computational Linguistics.
\newblock \doi{10.18653/v1/2024.findings-acl.813}.
\newblock URL \url{https://aclanthology.org/2024.findings-acl.813/}.

\bibitem[Wang et~al.(2024)Wang, Ren, Li, Zhao, Liu, and Wen]{wang2024rear}
Yuhao Wang, Ruiyang Ren, Junyi Li, Wayne~Xin Zhao, Jing Liu, and Ji-Rong Wen.
\newblock Rear: A relevance-aware retrieval-augmented framework for open-domain
  question answering.
\newblock \emph{arXiv preprint arXiv:2402.17497}, 2024.
\newblock URL \url{https://arxiv.org/abs/2402.17497}.

\bibitem[Wei et~al.(2024)Wei, Karina, Chung, Jiao, Papay, Glaese, Schulman, and
  Fedus]{wei2024measuringshortformfactualitylarge}
Jason Wei, Nguyen Karina, Hyung~Won Chung, Yunxin~Joy Jiao, Spencer Papay,
  Amelia Glaese, John Schulman, and William Fedus.
\newblock Measuring short-form factuality in large language models, 2024.
\newblock URL \url{https://arxiv.org/abs/2411.04368}.

\bibitem[Williams et~al.(2018)Williams, Nangia, and
  Bowman]{williams-etal-2018-broad}
Adina Williams, Nikita Nangia, and Samuel Bowman.
\newblock A broad-coverage challenge corpus for sentence understanding through
  inference.
\newblock In Marilyn Walker, Heng Ji, and Amanda Stent (eds.),
  \emph{Proceedings of the 2018 Conference of the North {A}merican Chapter of
  the Association for Computational Linguistics: Human Language Technologies,
  Volume 1 (Long Papers)}, pp.\  1112--1122, New Orleans, Louisiana, June 2018.
  Association for Computational Linguistics.
\newblock \doi{10.18653/v1/N18-1101}.
\newblock URL \url{https://aclanthology.org/N18-1101}.

\bibitem[Xie et~al.(2024)Xie, Zhang, Chen, Lou, and Su]{xie2024adaptive}
Jian Xie, Kai Zhang, Jiangjie Chen, Renze Lou, and Yu~Su.
\newblock Adaptive chameleon or stubborn sloth: Revealing the behavior of large
  language models in knowledge conflicts.
\newblock In \emph{The Twelfth International Conference on Learning
  Representations}, 2024.
\newblock URL \url{https://openreview.net/forum?id=auKAUJZMO6}.

\bibitem[Xu et~al.(2024)Xu, Shi, and Choi]{xu2024recomp}
Fangyuan Xu, Weijia Shi, and Eunsol Choi.
\newblock {RECOMP}: Improving retrieval-augmented {LM}s with context
  compression and selective augmentation.
\newblock In \emph{The Twelfth International Conference on Learning
  Representations}, 2024.
\newblock URL \url{https://openreview.net/forum?id=mlJLVigNHp}.

\bibitem[Yang et~al.(2024)Yang, Yang, Hui, Zheng, Yu, Zhou, Li, Li, Liu, Huang,
  Dong, Wei, Lin, Tang, Wang, Yang, Tu, Zhang, Ma, Xu, Zhou, Bai, He, Lin,
  Dang, Lu, Chen, Yang, Li, Xue, Ni, Zhang, Wang, Peng, Men, Gao, Lin, Wang,
  Bai, Tan, Zhu, Li, Liu, Ge, Deng, Zhou, Ren, Zhang, Wei, Ren, Fan, Yao,
  Zhang, Wan, Chu, Liu, Cui, Zhang, and Fan]{qwen2}
An~Yang, Baosong Yang, Binyuan Hui, Bo~Zheng, Bowen Yu, Chang Zhou, Chengpeng
  Li, Chengyuan Li, Dayiheng Liu, Fei Huang, Guanting Dong, Haoran Wei, Huan
  Lin, Jialong Tang, Jialin Wang, Jian Yang, Jianhong Tu, Jianwei Zhang,
  Jianxin Ma, Jin Xu, Jingren Zhou, Jinze Bai, Jinzheng He, Junyang Lin, Kai
  Dang, Keming Lu, Keqin Chen, Kexin Yang, Mei Li, Mingfeng Xue, Na~Ni, Pei
  Zhang, Peng Wang, Ru~Peng, Rui Men, Ruize Gao, Runji Lin, Shijie Wang, Shuai
  Bai, Sinan Tan, Tianhang Zhu, Tianhao Li, Tianyu Liu, Wenbin Ge, Xiaodong
  Deng, Xiaohuan Zhou, Xingzhang Ren, Xinyu Zhang, Xipin Wei, Xuancheng Ren,
  Yang Fan, Yang Yao, Yichang Zhang, Yu~Wan, Yunfei Chu, Yuqiong Liu, Zeyu Cui,
  Zhenru Zhang, and Zhihao Fan.
\newblock Qwen2 technical report.
\newblock \emph{arXiv preprint arXiv:2407.10671}, 2024.

\bibitem[Yang et~al.(2018)Yang, Qi, Zhang, Bengio, Cohen, Salakhutdinov, and
  Manning]{yang-etal-2018-hotpotqa}
Zhilin Yang, Peng Qi, Saizheng Zhang, Yoshua Bengio, William Cohen, Ruslan
  Salakhutdinov, and Christopher~D. Manning.
\newblock {H}otpot{QA}: A dataset for diverse, explainable multi-hop question
  answering.
\newblock In Ellen Riloff, David Chiang, Julia Hockenmaier, and Jun{'}ichi
  Tsujii (eds.), \emph{Proceedings of the 2018 Conference on Empirical Methods
  in Natural Language Processing}, pp.\  2369--2380, Brussels, Belgium,
  October-November 2018. Association for Computational Linguistics.
\newblock \doi{10.18653/v1/D18-1259}.
\newblock URL \url{https://aclanthology.org/D18-1259/}.

\bibitem[Yao et~al.(2024)Yao, Qi, Pan, Cao, Hu, Liu, Hou, and Li]{seakr}
Zijun Yao, Weijian Qi, Liangming Pan, Shulin Cao, Linmei Hu, Weichuan Liu, Lei
  Hou, and Juanzi Li.
\newblock Seakr: Self-aware knowledge retrieval for adaptive retrieval
  augmented generation, 2024.
\newblock URL \url{https://arxiv.org/abs/2406.19215}.

\bibitem[Yoran et~al.(2024)Yoran, Wolfson, Ram, and Berant]{yoran2024making}
Ori Yoran, Tomer Wolfson, Ori Ram, and Jonathan Berant.
\newblock Making retrieval-augmented language models robust to irrelevant
  context.
\newblock In \emph{The Twelfth International Conference on Learning
  Representations}, 2024.
\newblock URL \url{https://openreview.net/forum?id=ZS4m74kZpH}.

\bibitem[Yu et~al.(2023)Yu, Zhang, Pan, Ma, Wang, and Yu]{yu2023CoN}
Wenhao Yu, Hongming Zhang, Xiaoman Pan, Kaixin Ma, Hongwei Wang, and Dong Yu.
\newblock Chain-of-note: Enhancing robustness in retrieval-augmented language
  models, 2023.
\newblock URL \url{https://arxiv.org/abs/2311.09210}.

\bibitem[Zhang et~al.(2024{\natexlab{a}})Zhang, Liu, Basaldella, and
  Collier]{zhang-etal-2024-luq}
Caiqi Zhang, Fangyu Liu, Marco Basaldella, and Nigel Collier.
\newblock {LUQ}: Long-text uncertainty quantification for {LLM}s.
\newblock In Yaser Al-Onaizan, Mohit Bansal, and Yun-Nung Chen (eds.),
  \emph{Proceedings of the 2024 Conference on Empirical Methods in Natural
  Language Processing}, pp.\  5244--5262, Miami, Florida, USA, November
  2024{\natexlab{a}}. Association for Computational Linguistics.
\newblock \doi{10.18653/v1/2024.emnlp-main.299}.
\newblock URL \url{https://aclanthology.org/2024.emnlp-main.299/}.

\bibitem[Zhang et~al.(2024{\natexlab{b}})Zhang, Xu, and
  Perez-Beltrachini]{zhang-etal-2024-fine}
Huajian Zhang, Yumo Xu, and Laura Perez-Beltrachini.
\newblock Fine-grained natural language inference based faithfulness evaluation
  for diverse summarisation tasks.
\newblock In Yvette Graham and Matthew Purver (eds.), \emph{Proceedings of the
  18th Conference of the European Chapter of the Association for Computational
  Linguistics (Volume 1: Long Papers)}, pp.\  1701--1722, St. Julian{'}s,
  Malta, March 2024{\natexlab{b}}. Association for Computational Linguistics.
\newblock URL \url{https://aclanthology.org/2024.eacl-long.102}.

\bibitem[Zhang et~al.(2021)Zhang, Gong, and Choi]{zhang-etal-2021-knowing}
Shujian Zhang, Chengyue Gong, and Eunsol Choi.
\newblock Knowing more about questions can help: Improving calibration in
  question answering.
\newblock In Chengqing Zong, Fei Xia, Wenjie Li, and Roberto Navigli (eds.),
  \emph{Findings of the Association for Computational Linguistics: ACL-IJCNLP
  2021}, pp.\  1958--1970, Online, August 2021. Association for Computational
  Linguistics.
\newblock \doi{10.18653/v1/2021.findings-acl.172}.
\newblock URL \url{https://aclanthology.org/2021.findings-acl.172}.

\end{thebibliography}
